\documentclass{article}
\PassOptionsToPackage{dvipsnames}{xcolor}
\usepackage[utf8]{inputenc}
\usepackage{iclr2026_conference}
\usepackage{newtxtext}
\usepackage{bbm}
\usepackage{algorithm}
\usepackage{algpseudocode}
\usepackage{enumitem}
\setlist{nosep}
\usepackage{graphicx}
\usepackage{fontawesome}
\usepackage{tabularx}
\usepackage{booktabs}
\usepackage{subcaption}
\usepackage{xcolor}
\usepackage{colortbl}
\usepackage{hyperref}
\hypersetup{colorlinks=true, linkcolor=blue, citecolor=blue, urlcolor=blue}
\usepackage{verbatim}
\usepackage{float}
\usepackage{flafter}
\usepackage{placeins}
\usepackage{needspace}
\usepackage{amsmath}
\usepackage{cleveref}
\usepackage{amssymb}
\usepackage{wrapfig}
\usepackage{makecell}
\usepackage{multirow}
\usepackage{fancyvrb}
\usepackage[normalem]{ulem}
\usepackage{url}
\usepackage{tcolorbox}
\usepackage{diagbox}
\usepackage{CJKutf8}
\usepackage{dashrule}
\tcbuselibrary{most}
\usepackage{arydshln}

\usepackage{makecell}
\usepackage{array}
\usepackage{ragged2e}
\tcbuselibrary{breakable} 
\usepackage{fvextra}  


\newcolumntype{Y}{>{\RaggedRight\arraybackslash}X}
\newcolumntype{Z}{>{\RaggedRight\arraybackslash}p{\dimexpr 5\linewidth/11-2\tabcolsep}}
\usepackage{threeparttable} 

\renewenvironment{abstract}{%
  \vskip 0.5em
  \begin{tcolorbox}[
      enhanced, breakable,
      colback=huaweilight,
      colframe=huaweired,
      boxrule=1pt,
      arc=2pt,
      coltitle=white,
      colbacktitle=huaweired,
      fonttitle=\bfseries\small,
      title={Abstract},
      attach boxed title to top left={xshift=12pt, yshift=-2.5mm},
      boxed title style={
        enhanced, colback=huaweired, colframe=huaweireddark,
        arc=0pt, outer arc=0pt, boxsep=1pt,
        left=7pt, right=7pt, top=2pt, bottom=2pt
      },
      left=12pt, right=12pt, top=8pt, bottom=8pt,
    ]\fontsize{9.5}{12.5}\selectfont
}{\end{tcolorbox}\par\vskip 1ex}

\newcommand{\abstractprojectlinks}{%
  \par\vspace{0.55em}%
  {\color{huaweired!55}\hrule height 0.45pt}%
  \vspace{0.5em}%
  {\footnotesize
    \noindent\textcolor{black}{\faGithub}\hspace{0.35em}%
    \textbf{GitHub: }%
    \url{https://huawei-noah.github.io/noah-research/ScienceFlow/website/}\par
  }%
}

\makeatletter
\renewcommand{\maketitle}{%
  \begin{center}
    \begin{tcolorbox}[
        enhanced,
        colback=white,
        frame empty,
        arc=0pt,
        width=\linewidth,
        halign=center,
        top=12pt, bottom=12pt, left=8pt, right=8pt,
        overlay={%
          \draw[huaweired, line width=1.6pt]
            ([yshift=-2pt]frame.north west) -- ([yshift=-2pt]frame.north east);
          \draw[huaweired, line width=1.6pt]
            ([yshift=2pt]frame.south west) -- ([yshift=2pt]frame.south east);
        }
      ]
      {\LARGE\bfseries\color{black}\@title\par}
    \end{tcolorbox}
    \vspace{0.8em}
    {\large\@author\par}
  \end{center}
  \vspace{0.5ex}
}
\makeatother

\newtcolorbox{mybox}[2][]
  {colback = black!5!white, colframe = black!75!black, fonttitle = \bfseries,
    colbacktitle = black!100!black, enhanced, before upper={\fontsize{8}{11}\obeyspaces\obeylines\selectfont}, fontupper=\selectfont,
    attach boxed title to top left={yshift=-2.2mm,xshift=4mm},
    title=#2,#1}

\newtcolorbox{promptbox}[3][]{
colback=black!5!white,
arc=5pt, 
boxrule=0.5pt,
fonttitle=\bfseries,
title=#3, 
before upper={\fontsize{8}{11}\obeyspaces\obeylines\selectfont}, 
fontupper=\selectfont,
colframe=#2,
}

\newtcolorbox{caseStudyBox}[2][]{
    breakable, 
    title=#2,  
    colback=gray!5!white, 
    colframe=gray!60!black, 
    fonttitle=\bfseries, 
    coltitle=black,
    attach boxed title to top left={yshift=-2mm, xshift=4mm}, 
    boxed title style={
        colback=white,
        colframe=gray!60!black,
    },
    #1 
}

\definecolor{lightgreen}{RGB}{145, 204, 117}
\definecolor{lightblue}{RGB}{173, 216, 230}
\definecolor{techblue}{RGB}{70, 130, 180}
\definecolor{academicred}{RGB}{178, 34, 34}
\definecolor{coolgray}{RGB}{128, 128, 128}

\definecolor{huaweired}{RGB}{199, 0, 11}
\definecolor{huaweireddark}{RGB}{150, 0, 10}
\definecolor{huaweilight}{RGB}{250, 240, 240}

\newlist{questions}{enumerate}{2}
\setlist[questions,1]{label=RQ\arabic*.,ref=RQ\arabic*}
\setlist[questions,2]{label=(\alph*),ref=\thequestionsi(\alph*)}

\title{\textcolor{huaweired}{ScienceFlow}: A Long-horizon Agent for\\
\mbox{ML Research}, Scientific Discovery and Beyond}

\author{
    ScienceFlow Team \\
    \normalfont Noah's Ark Lab, Huawei
}
\iclrfinalcopy
\begin{document}

\maketitle
\lhead{\textsf{Technical Report}}
\rhead{\href{https://www.noahlab.com.hk/news/212}{%
    \raisebox{-0.18\height}{\includegraphics[height=18pt]{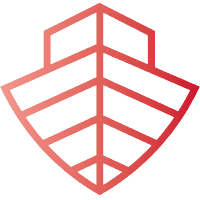}}}}
\fancypagestyle{scienceflowfirstpage}{%
    \fancyhf{}%
    \fancyhead[L]{\textsf{Technical Report}}%
    \fancyhead[R]{\href{https://www.noahlab.com.hk/news/212}{%
        \raisebox{-0.18\height}{\includegraphics[height=18pt]{template/noah_logo.png}}}}%
    \fancyfoot[L]{%
        \parbox[b]{0.48\textwidth}{\raggedright%
            \rule{0.75\linewidth}{0.4pt}\\[0.25em]
            \scriptsize Full author list: \hyperref[sec:contributions-acknowledgments]{\textcolor{huaweired}{Contributions and Acknowledgments}}.}}%
    \fancyfoot[C]{\thepage}%
}
\thispagestyle{scienceflowfirstpage}
\vspace{-1em}
\begin{abstract}
Enabling LLM agents to sustain productive, stable, and goal-aligned research over extended horizons is a central challenge for autonomous machine learning and scientific discovery, as progress hinges on continuously managing evolving state, exploration decisions, and computational resources.
Pioneering autoresearch agents, despite great success, still lack mechanisms for continuity, recovery from dead ends, and value‑driven compute allocation, which inherently undermines overall search efficiency, wastes computational resources, and lowers the chance of ultimate success. 
To bridge this gap, we introduce \textbf{ScienceFlow}, an end-to-end autoresearch agent framework that organizes long-horizon research work into research segments grounded in executable workspaces. It represents research progress as recoverable executable states, enabling efficient exploration, revision, and execution. Transitions between research segments are governed by Executable-State Transition through Re-Anchoring (\textbf{ESTRA}), which selects either the live state or an archived state as the next anchor and determines whether to continue or redirect the research trajectory.
An evidence-aware execution controller allocates resources to physical jobs based on resource availability, remaining budget, and validated progress.
We evaluate ScienceFlow on tasks spanning machine learning, scientific modeling, and mathematical optimization. 
Results on diverse long-horizon benchmarks demonstrate its ability to sustain effective research processes, highlighted by a SOTA \textbf{70.22\%} Any-Medal score on the full MLE-bench within a 24-hour budget, and outperforming prior reported results by 4.92 percentage points. 
The efficacy of ScienceFlow further demonstrates that efficient state management, adaptive exploration, and objective‑aligned execution are critical for scaling autonomous research beyond short‑horizon interactions.
\abstractprojectlinks
\end{abstract}

\vspace{-1em}
\begin{figure}[H]
    \centering
    \begin{subfigure}[b]{0.685\textwidth}
        \vspace{0pt}
        \centering
        \includegraphics[width=\linewidth]{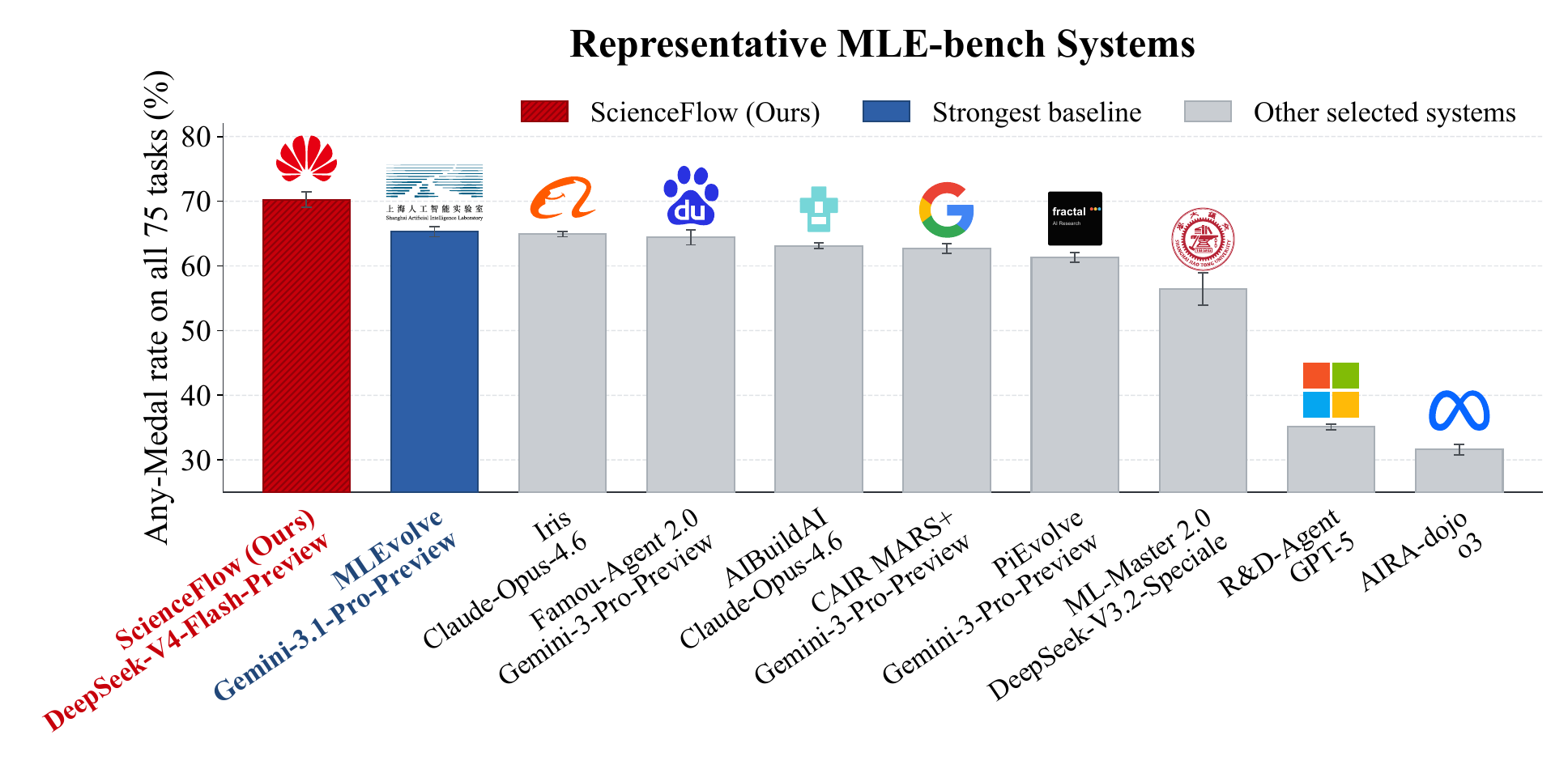}
        \vspace{-2em}
        \caption{Full MLE-bench Any-Medal rate.}
        \label{fig:mlebench-top10}
    \end{subfigure}\hfill
    \begin{subfigure}[b]{0.3\textwidth}
        \vspace{0pt}
        \centering
        \raisebox{7mm}{\includegraphics[width=\linewidth]{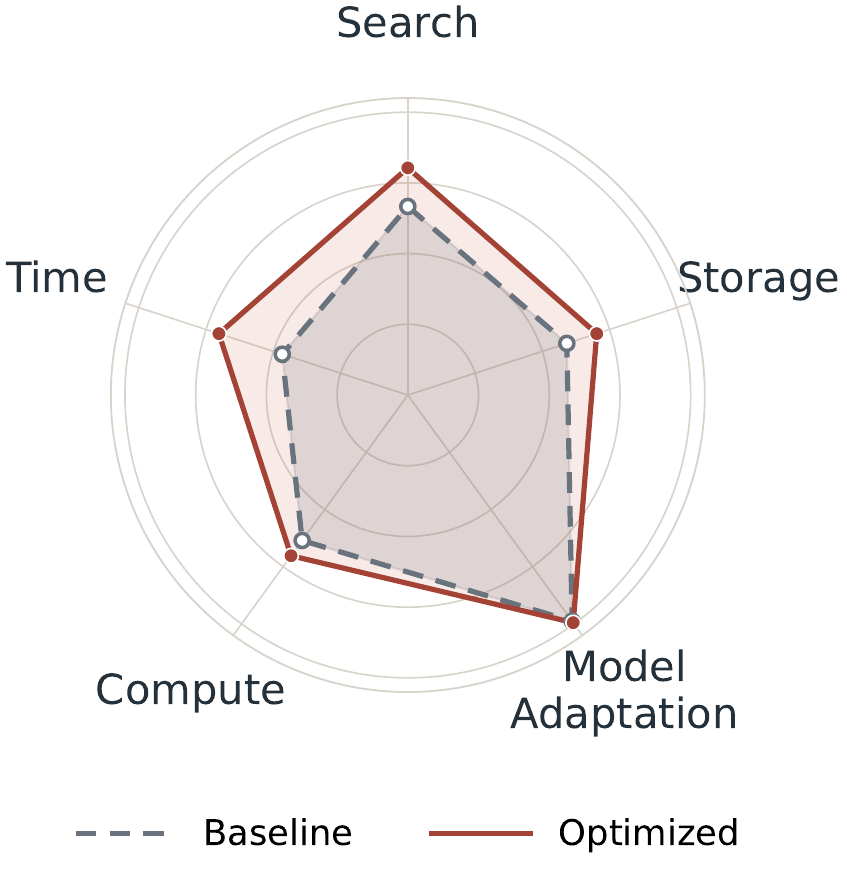}}
        \vspace{-2em}
        \caption{Agent capability profile.}
        \label{fig:homepage-system-profile}
    \end{subfigure}
    \caption{Performance overview. 
    (a) End-to-end performance on the full MLE-bench, where ScienceFlow achieves $70.22\pm1.18\%$ Any-Medal.  
    (b) Capability analysis across Search, Storage, Compute, Model Adaptation, and Time dimensions under full and constrained-reference settings. Each dimension is evaluated independently.
    }
    \label{fig:homepage-summary}
\end{figure}

\clearpage
\setcounter{tocdepth}{2}
\tableofcontents
\clearpage




\section{Introduction}
\label{sec:introduction}


AI autoresearch agents are rapidly evolving beyond isolated reasoning and tool use toward increasingly complete, long-horizon research workflows.
Recent systems can formulate hypotheses, modify executable artifacts, run experiments, interpret intermediate evidence, and revise subsequent decisions across machine learning and scientific discovery~\citep{aiscientistv1,aiscientistv2,tie2026aiscientists,wei2025agenticscience}. AlphaEvolve~\citep{novikov2025alphaevolve}, for example, demonstrates that iterative code modification guided by evaluator feedback can produce meaningful improvements in algorithms and scientific constructions. 
As research agents scale to longer, highly automated workflows, the bottleneck shifts from executing discrete research steps to sustaining productive, stable, and goal-aligned progress over time. This demands that agents preserve state across iterations, synthesize intermediate evidence, navigate competing directions, recover from dead ends, and translate available time and compute into tangible improvements.


Achieving sustained progress is complicated by the expanding nature of the research state itself. 
Over time, the agent's context grows beyond a simple interaction log into a rich executable workspace containing source code, datasets, cached artifacts, model checkpoints, solver states, evaluation outputs, and other artifacts.
While textual summaries can compress past interactions, they cannot faithfully reconstruct the executable state required to reliably resume, reuse, or revisit earlier work---a limitation that has motivated structured memory and persistent workspace designs~\citep{zhu2026mlmaster2,chen2026longhorizon,qian2026autosci}.
Moreover, promising directions often consume significant time and compute before revealing their flaws; apparent progress may be an artifact of noisy evaluation signals, implementation choices, or strategies that have already exhausted their local potential~\citep{toledo2025ai}. Decisions to continue, branch, or revert to an earlier state thus directly govern where additional resources are deployed. Consequently, effective long-horizon research demands that state persistence, exploration choices, and execution control evolve together as evidence accumulates.

Recent systems have tackled isolated facets of this problem through disparate representations of the research process. 
Structured and persistent memories preserve project knowledge and intermediate artifacts across iterations~\citep{qian2026autosci,chen2026longhorizon}, while trajectory-centric methods support branching, recovery, or iterative refinement of research paths~\citep{chen2026mage,zhang2026pivot}. 
Resource-aware systems operate at yet another layer, using runtime signals to place, monitor, and reschedule heterogeneous workloads~\citep{wang2026marscoscheduling,lu2026agenticscheduling}. 
While individually useful, these representations leave research state, trajectory decisions, and physical execution only loosely connected. 
Recovery, exploration, evaluation, and resource allocation each often operate over different views of the research process, making it difficult to maintain a consistent notion of progress as research evolves. 
This fragmentation remains a fundamental obstacle to sustaining coherent, adaptive, and efficient research over long horizons.

To this end, we introduce \textbf{ScienceFlow}, a workspace-grounded autonomous research system that supports one or more homogeneous research workers.
ScienceFlow organizes research around recoverable executable workspaces, which preserve the concrete state of each research trajectory and serve as stable boundaries for continuation, branching, and recovery. 
During forward research, task-specific result signals produce validated workspace checkpoints, while text-only responses or context-capacity limits define research-segment boundaries that invoke ESTRA.
At each research-segment boundary, a research worker uses ESTRA to continue or redirect from the live workspace, or to restore an archived checkpoint and proceed from that state.
Multiple homogeneous workers can operate on independent trajectories in parallel, exchanging only compact progress summaries at research-segment boundaries to limit interference while sharing useful evidence.
Beneath this research layer, an evidence-aware execution controller manages the physical jobs associated with each trajectory, using both resource conditions and validated progress to admit, monitor, pause, or terminate execution. 
Scientific decisions remain with the research workers, while the controller provides resource-aware execution and system-level safeguards.

The broader question is whether a common set of long-horizon research abstractions can transfer across substantially different executable tasks. 
We evaluate this question across machine learning, scientific modeling, and mathematical optimization. 
ScienceFlow achieves strong performance across all three domains, including a $70.22 \pm 1.18\%$ Any-Medal rate on the full 75-task MLE-bench~\citep{chan2025mle} under a 24-hour budget, 4.92 percentage points above the strongest reported baseline, while also attaining the best group-balanced score on SciModelingBench~\citep{scimodelingbench2026} and reaching/improving published frontiers on multiple mathematical optimization problems.
Our contributions are summarized as follows:
\begin{itemize}[leftmargin=*]
    \item We identify that effective long-horizon research depends not simply on extending interaction time or compute, but on jointly maintaining executable state, adapting research trajectories, and aligning execution with validated progress.
    \item ScienceFlow realizes this principle through recoverable executable workspaces, ESTRA-governed transitions between research segments, and evidence-aware execution control, enabling research to continue, redirect, or recover without losing useful progress.
    \item ScienceFlow demonstrates strong and consistent performance across a diverse range of tasks, including machine learning, scientific modeling, and mathematical optimization, with substantial gains on MLE-bench and SciModelingBench and state-of-the-art results across multiple optimization problems.
\end{itemize}

\section{Method}
\label{sec:method}

\subsection{Preliminaries}
\label{subsec:task-formulation}
We formalize executable research tasks whose outcome quality can be quantified by a goal-dependent utility.
A task is defined as $\mathcal{T}=(\mathcal{G},U_{\mathcal{G}},V_{\mathcal{G}},B,W_0)$, where $\mathcal{G}$ specifies the research goal and success criteria, $U_{\mathcal{G}}(\cdot)$ denotes the final task utility, $V_{\mathcal{G}}(\cdot)$ is the validation signal observable during research, $B$ specifies the available resource budget, and $W_0$ is the initial executable workspace.
Starting from $W_0$, the agent explores a growing collection of recoverable executable states. 
Let \(\mathcal{A}_T\) denote the archive of checkpointed executable states created over a research horizon \(T\).
Each executable state \(s_v \in \mathcal{A}_T\) represents a concrete checkpoint of the evolving research process, including the executable workspace and its associated artifacts, memory, validation evidence, and resource records. 
The state archive therefore provides a set of recoverable states from which the agent can continue, revisit, or redirect its research trajectory.
Since $U_{\mathcal{G}}$ is often unavailable during execution, the agent uses $V_{\mathcal{G}}$ to guide search and seeks the best validated state within the available budget:
\begin{equation}
\hat{s}_T=\arg \max _{s_v \in \mathcal{A}_T} V_{\mathcal{G}}\left(s_v\right) \quad \text { s.t. } \quad C_T=\sum_{t=1}^{T}\rho_t\preceq B,
\end{equation}
where \(\rho_t\) denotes the resource-consumption vector at step \(t\), and \(C_T\) is the cumulative resource expenditure. The budget vector \(B=(B_1,\ldots,B_K)\) specifies \(K\) component-wise limits, such as wall-clock time, compute, and storage. For any \(t\le T\), let \(C_t=\sum_{\tau=1}^{t}\rho_\tau\) and \(B_t=B-C_t\) denote the cumulative expenditure and remaining resource budget, respectively. The selected state is ultimately assessed by $U_{\mathcal{G}}$.

\subsection{System Overview}
\begin{figure}[t]
    \centering
    \includegraphics[width=\linewidth]{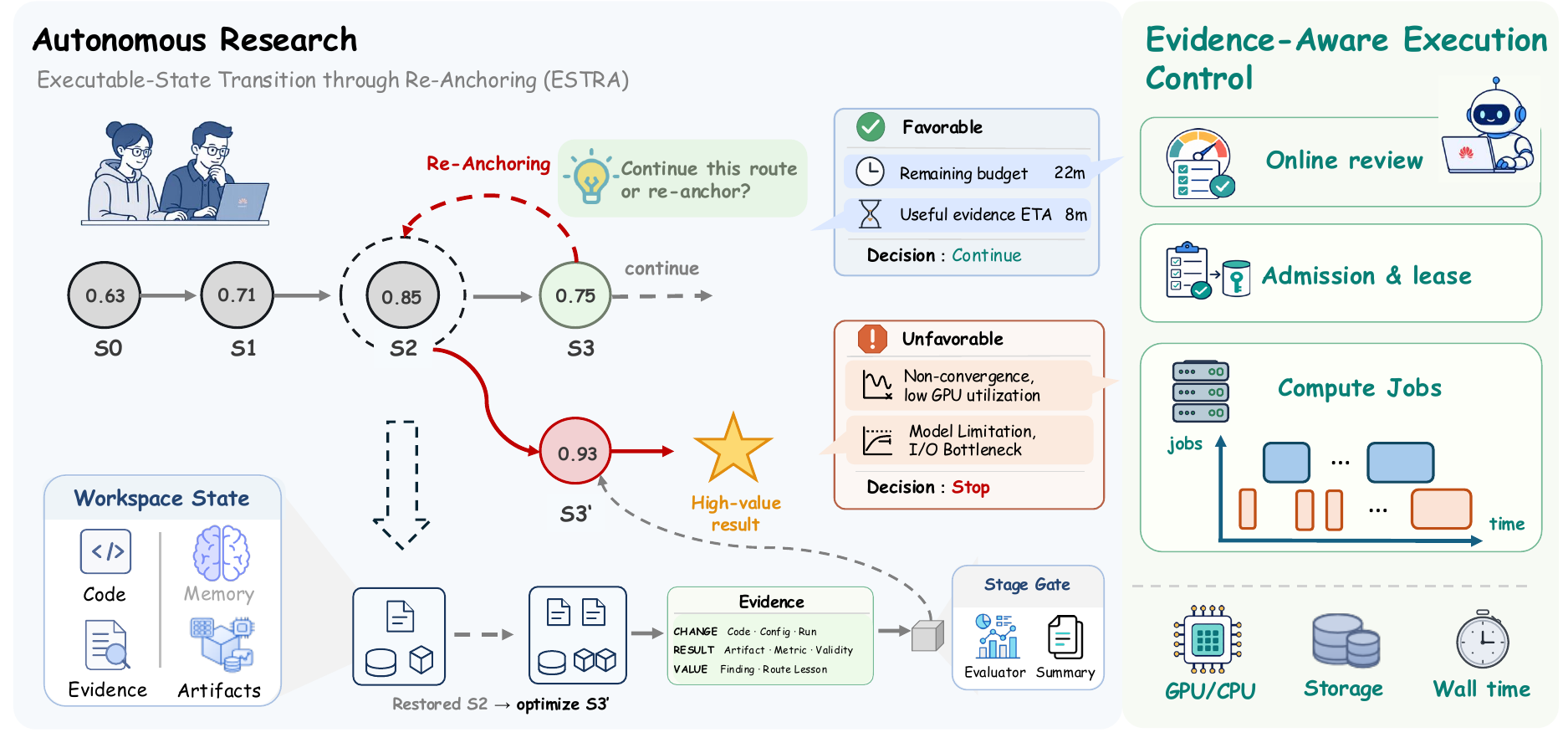}
    \caption{ScienceFlow system architecture. 
    Research workers operate over recoverable executable states and adapt long-horizon trajectories through ESTRA-governed transitions between research segments, while an evidence-aware execution controller coordinates physical resource allocation and runtime execution.
    }
    \label{fig:scienceflow-architecture}
\end{figure}

\paragraph{Long-Horizon Research Process.} ScienceFlow extends conventional tool‑using inference loops to long‑horizon executable research by organizing the process around recoverable executable states. Each state binds an executable workspace with compact memory, validation evidence, and resource records. These states persist across research iterations, providing a common foundation for resuming execution, adapting trajectories, and coordinating physical resources as new evidence accumulates.
During the long‑horizon search, ScienceFlow uses a stage gate to determine when to checkpoint each research state, including all artifacts of the current workspace and execution results—into the state archive.
This stage gate is triggered automatically by a task‑specific result signal. In addition, \emph{Executable-State Transition through Re-Anchoring} (\textbf{ESTRA}) governs transitions between research segments by selecting the next anchor state and an extend-or-redirect direction.
More concretely, ESTRA is triggered by either a text‑only response from the worker or an approaching context‑capacity limit. At each trigger, the research worker uses ESTRA to select an execution anchor and an extend-or-redirect direction based on the available states, accumulated validation evidence, and remaining resource budget. If an archived anchor is selected, ScienceFlow restores the corresponding state before the next research segment begins.
Together, these mechanisms connect recoverable executable states, adaptive trajectory transitions, and physical execution within a single long-horizon research process.
Figure~\ref{fig:scienceflow-architecture} provides an overview of the ScienceFlow architecture and its long-horizon research process.



\paragraph{System and Worker Model.}
ScienceFlow supports one or more homogeneous research workers, each operating within an isolated executable workspace and maintaining its own state archive. 
Multiple workers explore independent research trajectories and exchange only compact validated progress at research-segment boundaries.
They share a physical resource pool governed by a common execution controller, which manages resource availability and runtime execution without participating in scientific route selection.

\subsection{Executable Research Process}
Long‑horizon research is organized as a sequence of research segments. Within each segment, the worker follows its local tool-use policy to inspect, modify, and execute the active workspace. Whenever ScienceFlow detects a task-specific result signal, the stage gate evaluates the result, records the resulting evidence, collects a compact progress summary, and checkpoints the corresponding executable state in the archive. Multiple such checkpoint events may occur within one segment. A text-only response or an approaching context-capacity limit closes the current segment and invokes ESTRA to initialize the next segment from a selected research state and direction. Each design is described in detail in the following subsections.

\subsubsection{Research State Representation}
We begin by defining the research state $s_v$, which captures a specific milestone in a long-running research process. Each state is indexed by a unique archive identifier \(v\), which denotes its corresponding entry in the archive $\mathcal A$ and makes the state retrievable. In ScienceFlow, $s_v$ maintains four distinct types of information:
 \begin{itemize}[leftmargin=*]
    \item \textbf{Workspace snapshot} \(W_v\): contains all executable artifacts available at the time of archival—including source code, data-processing scripts, model checkpoints, cached features, validation outputs, submissions, environment metadata, and references to large artifacts.
    \item \textbf{Structured research memory} \(m_v\): maintains a compact long-horizon agent memory that persists across all research segments, which is detailed in~\cref{sec:memory}.
    \item \textbf{Validation evidence} \(e_v\): stores the validation results associated with the current research state, defined as $e_v = V_{\mathcal{G}}(s_v)$.
    \item \textbf{Resource records} \(\ell_v\): resource records captured from the current resource ledger at the time of checkpointing.
\end{itemize}
To make this research state $s_v$ actionable, it is represented as a tuple of these four components:
\begin{equation}
s_v=\left(W_v, m_v, e_v, \ell_v\right),
\end{equation}
which can be inserted into or retrieved from the archive $\mathcal A$ as needed.

\subsubsection{Forward Research}
Each research segment executes a forward research process that may produce multiple checkpointed research states. During this process, the worker follows a local reasoning–action–observation policy to interact with the active workspace. For a given segment $n$, we define:
\begin{itemize}[leftmargin=*]
    \item  \(P_n\): the fixed segment-level base context, recovered from the cross-segment structured memory $m_v$ from the last research segment.
    \item  \(W_{n,j}\): the active workspace at round $j$, where \(W_{n,0}\) is cloned from the workspace snapshot $W_v$ of the previous research segment.
    \item \(h_{n,j}\): the segment-local interaction history (agent context) accumulated up to round $j$. It captures the reasoning content, tool actions, and observations across rounds and is initialized as $h_{n,0}=\varnothing$.
\end{itemize}
At each round $j$ the policy generates a reasoning output \(r_{n,j}\) and tool action $\alpha_{n,j}$:
\begin{equation}
    (r_{n,j},\alpha_{n,j})\sim\pi_{\mathrm{LLM}}\left(\cdot\mid P_n,h_{n,j}\right).
\end{equation}
If \(\alpha_{n,j}\neq\varnothing\), ScienceFlow executes the action, producing an observation \(o_{n,j}\) and an updated workspace:
\begin{equation}
\left(o_{n, j}, W_{n, j+1}\right)=\operatorname{Exec}\left(\alpha_{n, j} ; W_{n, j}\right) .
\end{equation}
If \(\alpha_{n,j}=\varnothing\), the step is text‑only; the workspace remains unchanged, \((W_{n,j+1}=W_{n,j}\)) and no observation is produced (\(o_{n,j}=\varnothing\)). 
After each round, the local history is updated by appending the reasoning, action, and observation:
\begin{equation}
h_{n, j+1}=h_{n, j} \oplus\left(r_{n, j}, \alpha_{n, j}, o_{n, j}\right),
\end{equation}
where $\oplus$ denotes ordered concatenation. 

Independently of the local interaction policy, ScienceFlow monitors the updated workspace for a task-specific result signal specified by the task adapter. Upon detection, the stage gate collects the result, invokes the evaluator to obtain validation evidence, and requests a compact progress summary from the worker. It then snapshots the workspace as $W_v=\operatorname{Snapshot}(W_{n,j+1})$, records the associated memory $m_v$, validation evidence $e_v$, and resource records $\ell_v$, and inserts the resulting state $s_v=(W_v,m_v,e_v,\ell_v)$ into the archive $\mathcal A_n$. Temporary stage-gate exchanges are excluded from the segment-local history $h_{n,j+1}$, and the worker resumes forward research within the same segment.

Separately, the current research segment closes and transfers control to ESTRA when either of the following conditions for a research-segment boundary is met:
\begin{enumerate}[leftmargin=*]
    \item The policy produces a text‑only response \(\alpha_{n,j}=\varnothing\); or
    \item The accumulated context exceeds the configured capacity threshold $\left|P_n\oplus h_{n,j+1}\right|\ge L_{\mathrm{ctx}}$
\end{enumerate}
If neither condition holds, the forward loop proceeds to round $j+1$. Otherwise, the live workspace at the trigger defines the current state $s_n^{\mathrm{cur}}$, and ScienceFlow initiates the route-level deliberation described next. When a result signal and the context-capacity condition coincide, ScienceFlow completes stage-gate processing first, making the newly checkpointed state available to the subsequent ESTRA decision.

\subsubsection{Re-Anchoring and Exact Restoration}
\label{subsubsec:reanchoring}
At each research-segment boundary, ScienceFlow invokes ESTRA for route-level deliberation by the research worker. The worker policy \(\pi_{\mathrm{LLM}}\) produces an anchor-and-direction decision \((a_n,d_n)\). This decision operates on two independent axes: \(a_n\) selects the execution anchor, either the current trigger state \(s_n^{\mathrm{cur}}\) or a previously archived state \(s_v\in\mathcal A_n\), while \(d_n\) determines whether to extend the current trajectory or redirect onto a new branch from that anchor. The worker bases this decision on the compact memory view, available anchor candidates with their validation evidence, and the current resource envelope. Formally:
\begin{equation}
a_n \in \{s_n^{\mathrm{cur}}\}\cup\mathcal{A}_n,\qquad
d_n \in \{\textsc{extend},\textsc{redirect}\}.
\end{equation}
Combining these two choices yields four possible outcomes: extending or redirecting from the current state, and extending or redirecting from an archived state. After the ESTRA decision, the next research segment $n+1$ is initialized by defining the segment-level base context $P_{n+1}$ and workspace $W_{n+1,0}$ as:
\begin{equation}
\left(P_{n+1},W_{n+1,0}\right)=\operatorname{ESTRA}\left(a_n,d_n;\mathcal A_n\right).
\end{equation}
If \(a_n\) selects the current executable state, ESTRA retains the current workspace \(W_{n,j+1}\) as \(W_{n+1,0}\). If \(a_n\) selects an archived executable state \(s_v\in\mathcal A_n\), ESTRA restores the full state, i.e., workspace \(W_v\), memory \(m_v\), validation evidence \(e_v\), and resource records \(\ell_v\), with resource accounting remaining cumulative across restoration, including resources consumed after archiving. The anchor thus determines the workspace from which research resumes, while \(d_n\) determines whether research extends the anchor's existing route or redirects from it. The resulting \(P_{n+1}\) carries the anchor's research summary, relevant evidence, and selected direction, while \(W_{n+1,0}\) provides the corresponding executable artifacts. ESTRA preserves the full archive while cleanly setting both the starting state and research direction for the next segment.


\subsubsection{Memory and Context Assembly}
\label{sec:memory}
ScienceFlow maintains a persistent memory $m_v$ that carries the entire research trajectory across segments. This memory is updated through two core operations: \textsc{Add} appends new progress summaries, and \textsc{Fold} compresses the memory when it grows too large. To support anchor selection during ESTRA, \textsc{Fold} retrieves historical records, and \textsc{Assemble} constructs the prompt that guides the agent's decision. Together, these operations ensure that the agent always has a compact yet complete view of its research history.

\paragraph{\textsc{Add}: accumulating progress.} At each stage-gate event, the worker summarizes the evaluated result, relevant artifacts, and current progress into a compact result card \(q_v\). This card is added to the persistent memory via:
\begin{equation}
m_v=\operatorname{Add}(m_v, q_v).
\end{equation}
This update preserves the new progress information in $m_v$ without resetting the segment-local history. ESTRA later uses the accumulated memory to assemble the context for the next research segment.

\paragraph{\textsc{Fold}: compressing memory.} When persistent memory $m_v$ exceeds its allocated context space  \(B_{\mathrm{mem}}\), \textsc{Fold} automatically compresses it:
\begin{equation}
(\widetilde m_v,\iota_v)=\operatorname{Fold}(m_v;B_{\mathrm{mem}}).
\end{equation}

where $\widetilde m_v$ is a compact view that keeps recent, best-validated, and anchor-relevant records in full while summarizing older ones, and $\iota_v$ is an index that maps each folded summary to its original records in the archive. The complete memory $m_v$ remains preserved in the archive.

ScienceFlow also applies Fold whenever an ESTRA transition closes a research segment, independently of $B_{\mathrm{mem}}$. If the selected anchor is the current state (\(a_n=s_n^{\mathrm{cur}}\)), \textsc{Fold} summarizes the research segment just ended. If the anchor is an archived state (\(a_n=s_v\in\mathcal A_n\)) it summarizes the exploration performed since last time that state was restored. This summary is carried forward as historical evidence, while the raw result cards remain addressable via \textsc{unfold}.

\paragraph{\textsc{Unfold}: retrieving history for anchor selection.} To select an anchor during ESTRA, the agent must understand the full trajectory. \textsc{Unfold} retrieves complete records associated with a requested identifier ($z_v$ a folded summary or an indexed result card):
\begin{equation}
u_v=\operatorname{Unfold}(m_v,\iota_v,z_v).
\end{equation}
The retrieved records $u_v$ temporarily augment the compact memory view for the current deliberation, while $m_v$ remains the persistent source.




\paragraph{\textsc{Assemble}: constructing the anchor-selection context.} With the memory view prepared, ScienceFlow constructs an anchor-specific context $P_{\mathrm{anchor}}(a_n)$ that presents the agent with the information needed to choose the next anchor and direction. For a selected anchor  \(a_n\), let \(W(a_n)\), \(m(a_n)\), \(e(a_n)\), and \(\ell(a_n)\) denote its retrieved state components, and let \(\widetilde m(a_n)\) denote its budgeted memory view (or $m(a_n)$ if folding is unnecessary). Let \(u(a_n)\) denote any records retrieved via Unfold. The \textsc{Assemble} operation combines these:
\begin{equation}
P_{\mathrm{anchor}}(a_n)
=
\operatorname{Assemble}\!\left(
\widetilde m(a_n)\oplus u(a_n),
e(a_n),
\ell(a_n),
\operatorname{View}(W(a_n))
\right).
\end{equation}
Where \(\operatorname{View}(W(a_n))\) provides a compact description of the available workspace artifacts and execution state.
The full base context for the next segment is then: 
\begin{equation}
P_{n+1}=\underbrace{P_{\text {worker }} \oplus P_{\text {runtime }} \oplus P_{\text {tools }} \oplus P_{\text {task }}}_{P_{\text {stable }}} \oplus P_{\text {anchor }}\left(a_n\right) \oplus P_{\text {dir }}\left(d_n\right),
\end{equation}
where $P_{\text {stable }}$ contains the worker policy, runtime instructions, tool descriptions, and task specification, and $P_{\operatorname{dir}}\left(d_n\right)$ encodes whether the next segment extends or redirects from the selected anchor. The next segment begins with $h_{n+1,0}=\varnothing$ and accumulates new interactions through the forward loop.
In summary, \textsc{Add} and \textsc{Fold} maintain a complete yet compact persistent memory across the research trajectory; \textsc{Unfold} retrieves historical records when the agent needs to re-anchor; and \textsc{Assemble} constructs the context that guides anchor selection and initializes each new segment. This design preserves the full research record while supplying each segment with a context-bounded, anchor-specific memory view.

\subsection{Evidence-Aware Execution Control} 
\label{subsec:execution-control}
A key distinction that separates ScienceFlow from existing autonomous research agents is the explicit separation of research-route selection from physical execution control. In this design, the research worker proposes executable jobs, while a dedicated evidence-aware execution controller determines whether, where, and for how long each job runs. This controller evaluates both pending and running jobs using current resource availability, remaining budget, validated research progress, and the recoverable state of execution artifacts. Accordingly, the control process operates in two stages: admission and allocation before a job is launched, and online review and replanning during execution.

\subsubsection{Execution Admission and Allocation}
\label{subsubsec:execution-admission}

For each proposed executable job \(b\), ScienceFlow constructs a resource request describing its command, requested device class and count, estimated runtime, memory requirements, and any research-value hints supplied by the research worker. ScienceFlow complements these hints with estimates inferred from the command and execution context to establish the job's admission priority. This priority reflects expected research value, lineage diversity, proximity to a deliverable, worker starvation, duplication, and runtime risk.

The admission-and-lease module then evaluates the request against the live resource state \(R_t\), which includes available GPU/CPU capacity, storage constraints, queue state, GPU pressure, and active leases, while also accounting for the remaining budget \(B_t\). If the preflight check yields a confident result, it is applied directly; otherwise, the request is passed to an isolated admission LLM for review. This LLM combines the resource request, research-value information, and preflight result to produce the admission decision:

\begin{equation}
\delta_b^{\mathrm{adm}}=\operatorname{Admit}(b;R_t,B_t)\in\{
\mathtt{RUN\_NOW},
\mathtt{OBSERVE\_THEN\_RUN},
\mathtt{PENDING},
\mathtt{REPLAN}\}.
\end{equation}

Each decision corresponds to a specific action:
\begin{itemize}[leftmargin=*]
    \item \textbf{\textsc{Run\_now}}: the job is launched immediately after ScienceFlow atomically acquires the required device lease.
    \item \textbf{\textsc{Observe\_then\_run}}: the job is launched with a short early-review window when its resource behavior remains uncertain.
    \item \textbf{\textsc{Pending}}: the job is retained in the priority-aware queue without a device lease.
    \item \textbf{\textsc{Replan}}:  the blocking evidence is returned to the research worker for plan revision.
\end{itemize}

Device leases coordinate exclusive or controlled shared access to compute resources and are released when the corresponding job finishes, is stopped, or no longer requires the assigned device.

\subsubsection{Online Review and Replan}
\label{subsubsec:execution-review}
Admission establishes the initial execution state of a compute job, after which ScienceFlow reviews its progress at successive observation boundaries. 
At each boundary, the controller collects timestamped evidence from process liveness, logs, metric history, artifact updates, resource utilization, checkpoint availability, and the remaining budget. The controller treats process liveness, log growth, and sustained resource utilization as evidence that a job remains active. 
Metric improvement, validated results, or newly recoverable artifacts provide stronger evidence that the job is making meaningful research progress.

From the accumulated evidence, ScienceFlow estimates the time until the job produces its next useful evidence, such as a comparable metric update or a recoverable artifact. When this useful-evidence ETA falls within the remaining budget and is supported by recent progress, the controller favors continued execution. Persistent non-convergence, low GPU utilization, model limitations, or I/O bottlenecks indicate that a job may fail to produce useful evidence within the remaining budget. When these signals persist across multiple observation windows and the job has produced no recoverable value, they provide strong evidence for stop-and-replan. 

When execution evidence alone leaves the job's prospective research value uncertain, ScienceFlow requests a resource advisory from the research worker. The worker reports its preferred action---\textsc{continue}, \textsc{timebox continue}, \textsc{safe to stop}, \textsc{replan}, or \textsc{unknown}---together with its confidence, a measurable commitment, and the expected next artifact. The controller incorporates this advisory as research-value evidence while retaining authority over lease allocation, execution timeboxing, and process termination. The controller then combines the accumulated execution evidence \(y_{0:t}\), the optional worker advisory \(g_t\), and the remaining budget \(B_t\) to produce a control decision:
\begin{equation}
c_t=\operatorname{Review}(y_{0:t},g_t;B_t)\in\{
\mathtt{CONTINUE},
\mathtt{TIMEBOX},
\mathtt{STOP\_AND\_REPLAN}\}.
\end{equation}
Each decision corresponds to a specific action:
\begin{itemize}[leftmargin=*]
    \item \textbf{\textsc{continue} }: keeps the current job running.
    \item \textbf{\textsc{timebox}}: grants a bounded proof window tied to a measurable progress condition.
    \item \textbf{\textsc{stop\_and\_replan}}: terminates the current job and redirects control to the research worker.
\end{itemize}
Notably, recent validated progress, a new checkpoint, or a feasible phase-completion ETA favors continued execution even when the job progresses slowly. Before applying \textsc{stop and replan}, ScienceFlow preserves recoverable artifacts, releases the assigned leases, and returns the observed execution facts to the research worker. 

Here, \(\mathtt{REPLAN}\) and \(\mathtt{STOP\_AND\_REPLAN}\) denote execution-control handoffs: the controller returns blocking or runtime evidence to the research worker, which determines the revised scientific plan. These outcomes do not themselves select the ESTRA anchor \(a_n\) or direction \(d_n\).

Together, admission and online review form a closed execution-control loop: ScienceFlow applies \(\operatorname{Admit}\) before launch and repeatedly applies \(\operatorname{Review}\) at subsequent observation boundaries until the job completes or control returns to the research worker for replanning. This loop allocates compute according to research value while coordinating GPU/CPU capacity, storage usage, and wall-clock time throughout execution. 

\section{Experiment}
We evaluate ScienceFlow across three classes of executable research tasks: machine learning engineering, mathematical optimization, and scientific modeling, covering substantially different forms of long-horizon research.
All runs are conducted in isolated environments without cross-run state sharing, and held-out evaluation information is never exposed to the research agent. 
Task-specific models, resource budgets, evaluation protocols, tools, and baselines are described in the corresponding subsections. 
Beyond overall task performance, we examine the research dynamics of ScienceFlow, including state evolution, trajectory adaptation, and resource utilization over the course of extended execution.


\subsection{Machine Learning Engineering}\label{sec:exp-mle}
\subsubsection{Setup}\label{subsec:exp-mlebench}

\begin{table}[t]
    \caption{Representative results on the full 75-task MLE-bench.
    Any-Medal rates (\%) are reported as mean $\pm$ SEM when three runs
    are available. Best results in each column are shown in \textbf{bold}.}
    \label{tab:main_results}
    \centering
    \footnotesize
    \renewcommand{\arraystretch}{1.}
    \setlength{\tabcolsep}{1.8pt}
    \begin{tabularx}{\textwidth}{
        @{}
        >{\raggedright\arraybackslash}p{0.2\textwidth}
        l
        *{4}{>{\centering\arraybackslash}X}
        @{}
    }
        \toprule
        \textbf{Agent} &
        \textbf{LLM(s)} & \textbf{Lite} (\%) & \textbf{Medium} (\%) & \textbf{High} (\%) & \textbf{All} (\%) \\
        \midrule
        \addlinespace[2pt]
        \multicolumn{6}{c}{\textit{Other reported systems}} \\
        \addlinespace[2pt]
        \midrule

        Famou-Agent & Gemini-2.5-Pro & 75.76 $\pm$ 1.52 & 57.89 $\pm$ 1.52 & 40.00 $\pm$ 0.00 & 59.56 $\pm$ 0.89 \\
        CAIR MARS+ & Gemini-3-Pro-Preview & 78.79 $\pm$ 1.52 & 60.53 $\pm$ 1.52 & 44.44 $\pm$ 2.22 & 62.67 $\pm$ 0.77 \\
        AIBuildAI & Claude-Opus-4.6 & 77.27 $\pm$ 0.00 & 61.40 $\pm$ 0.88 & 46.67 $\pm$ 0.00 & 63.11 $\pm$ 0.44 \\
        Famou-Agent 2.0 & Gemini-3-Pro-Preview & \textbf{80.30 $\pm$ 1.52} & 64.04 $\pm$ 2.32 & 42.22 $\pm$ 2.22 & 64.44 $\pm$ 1.18 \\
        Iris & Claude-Opus-4.6 & \textbf{80.30 $\pm$ 1.50} & 64.00 $\pm$ 0.90 & 44.40 $\pm$ 2.20 & 64.90 $\pm$ 0.40 \\
        \midrule
        \addlinespace[2pt]
        \multicolumn{6}{c}{\textit{Open-source baselines}} \\
        \addlinespace[2pt]
        \midrule
        ML-Master 2.0 & DeepSeek-V3.2-Speciale & 75.76 $\pm$ 1.51 & 50.88 $\pm$ 3.51 & 42.22 $\pm$ 2.22 & 56.44 $\pm$ 2.47 \\
        MLEvolve & Gemini-3-Pro-Preview & \textbf{80.30 $\pm$ 1.52} & 57.89 $\pm$ 1.52 & 42.22 $\pm$ 2.22 & 61.33 $\pm$ 1.33 \\
        PiEvolve & Gemini-3-Pro-Preview & \textbf{80.30 $\pm$ 1.52} & 58.77 $\pm$ 0.88 & 40.00 $\pm$ 0.00 & 61.33 $\pm$ 0.77 \\
        MLEvolve & Gemini-3.1-Pro-Preview & \textbf{80.30 $\pm$ 1.50} & 64.00 $\pm$ 0.90 & \textbf{46.70 $\pm$ 0.00} & 65.30 $\pm$ 0.80 \\
        \midrule
        \makecell[l]{\textbf{ScienceFlow}\\\textbf{(Ours)}} &
        \makecell[l]{\textbf{DeepSeek-V4-Flash-}\\\textbf{Preview}} &
        \makecell[c]{\textbf{80.30 $\pm$ 1.52}} &
        \makecell[c]{\textbf{74.56 $\pm$ 0.88}} &
        \makecell[c]{44.44 $\pm$ 2.22} &
        \makecell[c]{\textbf{70.22 $\pm$ 1.18}} \\
        \bottomrule
    \end{tabularx}
\end{table}

\paragraph{Benchmark and Metric}
We evaluate ScienceFlow on the full MLE-bench~\citep{chan2025mle}, which contains 75 real-world Kaggle competitions spanning tabular, vision, language, audio, and time-series tasks. Each task requires the agent to construct an executable machine-learning pipeline and produce a valid submission evaluated on a held-out test set constructed from the original competition data. Unlike evaluations restricted to the 22-task Lite split, we report results on the official Lite, Medium, and High tiers and on the aggregate 75-task benchmark. The primary metric is Any-Medal rate: a task run succeeds when its submission reaches the Bronze, Silver, or Gold threshold derived from the original Kaggle leaderboard. ScienceFlow results are reported as mean $\pm$ SEM over three independent runs, while baseline statistics are retained in the form reported by their original sources.

\paragraph{Model and Resource Budget}
For the primary MLE-bench evaluation, ScienceFlow uses DeepSeek-V4-Flash-Preview, with a 24-hour budget and at most 2 GPUs, 16--32 logical CPU cores, and 256\,GB RAM per run. Because hardware and concurrency vary across systems, comparisons use wall-clock time rather than normalized accelerator-hours. Budget deviations are documented in Table~\ref{tab:mlebench-full-leaderboard} in Appendix~\ref{app:full-mlebench}.

\paragraph{Tools and Baselines}
ScienceFlow operates in a sandboxed Python environment, with network access limited to third-party packages and publicly available model checkpoints. Held-out test labels and test-derived feedback are never exposed to the agent; the test set is used only by the final scorer. Baselines are drawn from the official MLE-bench leaderboard~\citep{openai2026mlebenchleaderboard} and publicly reported studies.

\subsubsection{Main Results}
\label{subsec:results-mle}
Table~\ref{tab:main_results} summarizes the main results on MLE-bench.
Across the full 75-task benchmark, ScienceFlow achieves an Any-Medal rate of $70.22 \pm 1.18\%$, exceeding the strongest reported baseline by 4.92 pp. 
The three independent runs obtain medals on 54, 53, and 51 tasks, respectively, demonstrating consistent performance across runs.
The improvement is particularly strong on the Medium tier, where ScienceFlow reaches $74.56 \pm 0.88\%$, 10.52 pp above the best reported baseline. 
It also achieves $80.30 \pm 1.52\%$ on Lite, matching the best reported result, and $44.44 \pm 2.22\%$ on High, within 2.26 pp of the best reported baseline.
Overall, ScienceFlow performs consistently across task complexities, with its largest advantage concentrated on the Medium tier.
The complete leaderboard and protocol annotations are provided in Table~\ref{tab:mlebench-full-leaderboard} in Appendix~\ref{app:full-mlebench}.

\subsubsection{Case Study and Ablations}\label{subsec:analysis-mlebench}
\paragraph{Long-horizon case study}

\begin{figure}[t]
    \centering
    \includegraphics[width=0.8\textwidth]{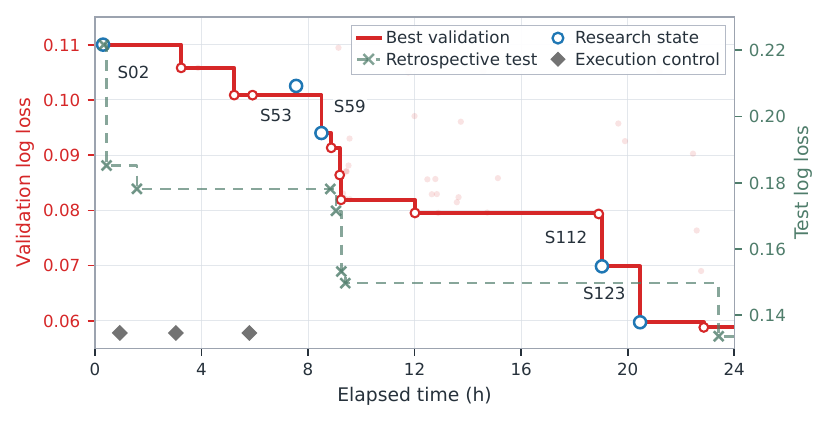}
    \caption{\textbf{A 24-hour ScienceFlow trajectory on the Statoil Iceberg Classifier Challenge}, showing validation and retrospective test progress together with checkpointed workspace states and resource-control events. Retrospective test scores are shown only for analysis and are never exposed to the research agent.
    }
    \label{fig:mle-case-study}
\end{figure}

We use the \emph{Statoil Iceberg Classifier Challenge}, a satellite-SAR binary classification task, as a representative long-horizon case study. The task requires distinguishing icebergs from ships from satellite SAR imagery under the same 24-hour budget used throughout MLE-bench. ScienceFlow earns a medal in all three independent runs. Figure~\ref{fig:mle-case-study} traces one run and illustrates how recoverable workspace states, ESTRA transitions, and evidence-aware execution control interact over the course of research.

The trajectory exhibits both state restoration and selective artifact reuse. After subsequent CNN-only exploration fails to improve the incumbent, ScienceFlow restores \texttt{S59} as an execution anchor and redirects the search toward tree models over saved CNN features. A later unproductive stacking branch similarly returns to \texttt{S112} before exploring a new route. Within these branches, the agent selectively reuses earlier workspace artifacts: \texttt{S59} combines fold checkpoints from \texttt{S02} and \texttt{S53}, \texttt{S112} trains a LightGBM model on CNN logits and angle features derived from \texttt{S53}, and \texttt{S123} combines the wide-64 and wide-96 assets from \texttt{S53}/\texttt{S55}. These operations reduce recomputation while improving validation loss from $0.1100$ at \texttt{S02} to $0.0597$ at \texttt{S123}.

Validation evidence continuously guides these trajectory decisions. The best validated loss decreases from $0.11$ to $0.1025$ and finally $0.0597$, while exploratory trials that fail to improve the incumbent remain preserved in the archive for later comparison or reuse. Retrospective test scores shown in Figure~\ref{fig:mle-case-study} are never exposed to the agent.  Their improvement alongside validation provides additional evidence that the validation signal used during search remains informative for the final submission.

Resource control operates concurrently with trajectory search. Two workers share a single-GPU device pool, and the marked events show three controller decisions: preserving exclusive GPU access after a sharing review, terminating a CPU-active job that reports $0\%$ GPU utilization, and falling back from a blocked heavy execution to lightweight inference, which produces a valid submission in $13.91$ seconds. Research-route selection remains with the worker, while the controller admits and monitors the corresponding physical execution under the available resources.

\begin{figure}[t]
    \centering
    \includegraphics[width=\textwidth]{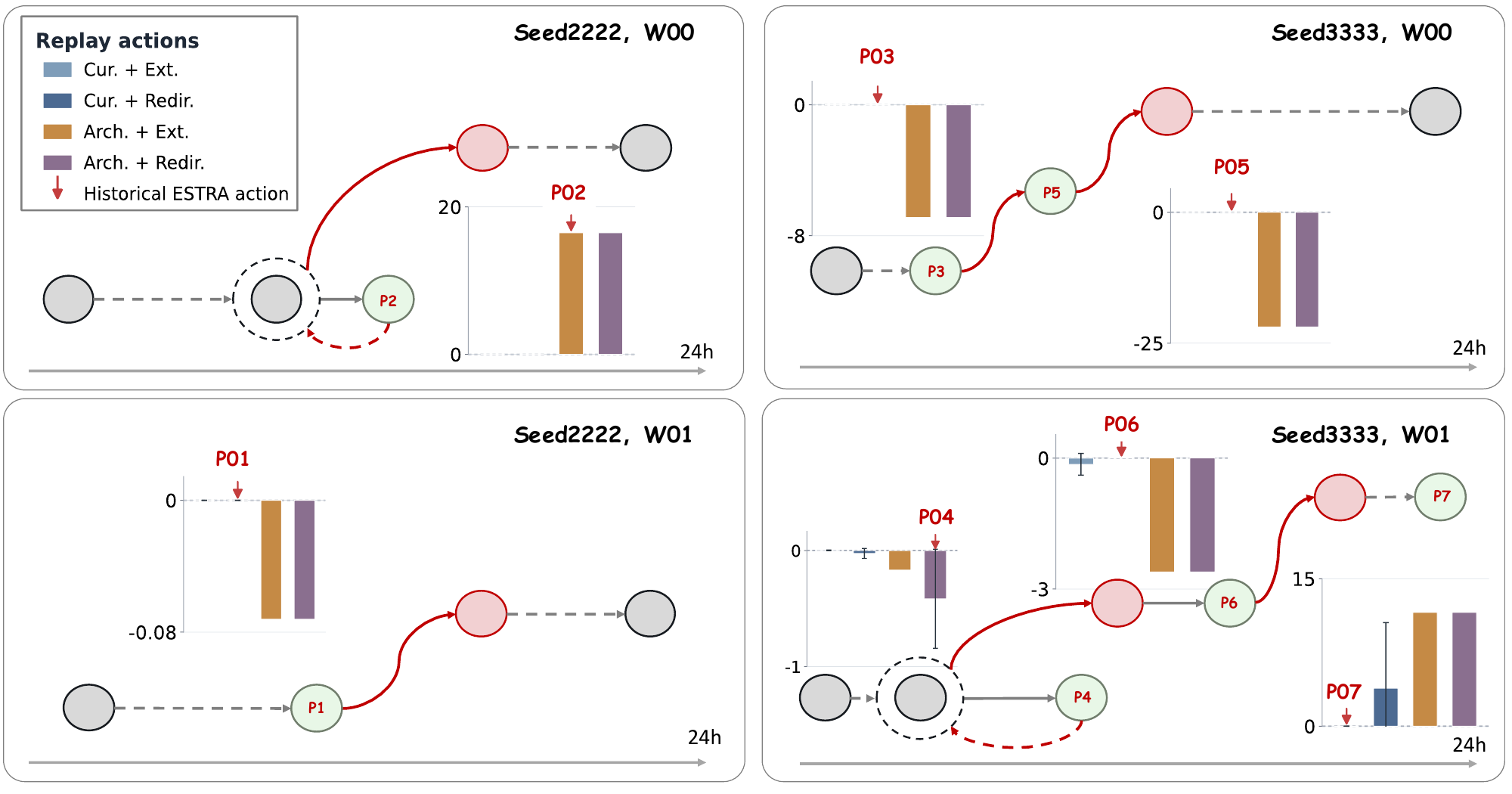}
    \caption{
    \textbf{State-matched counterfactual replay on the \emph{Tabular Playground Series May 2022} task.} Seven historical ESTRA decision points are replayed from their captured executable states under matched budgets. Each point compares four alternatives formed by current or archived anchors and extend or redirect directions. 
    Bars show the mean change in retrospective private-test AUROC relative to the source state across three replay seeds, with error bars denoting sample standard deviation. 
    Red arrows mark the actions selected in the historical run.
    }
    \label{fig:state-matched-re-anchoring}
\end{figure}
\paragraph{State-matched re-anchoring study}
To isolate the effect of re-anchoring from differences in accumulated context and workspace artifacts, we replay seven historical ESTRA decisions from their captured executable states (Figure~\ref{fig:state-matched-re-anchoring}).
At each decision point, four alternatives are evaluated by crossing the execution anchor (current or archived state) with the research direction (extend or redirect), using three replay seeds and matched four-hour resource budgets, for a total of 84 replay branches. 
Because all alternatives are launched from the same recorded decision context and resource budget, their differences enable a controlled comparison of the execution-anchor and research-direction choices. 
The historical ESTRA action achieves the best mean replay score, including ties, at five of the seven decision points. 
Relative to an oracle that selects the best of the four replayed actions at each point, the historical policy has zero median regret and a mean regret of $1.725\times10^{-3}$ AUROC, with the largest regret of $11.572\times10^{-3}$ occurring at \texttt{P07}.

The two deviations provide complementary evidence about how action value depends on execution reliability and the remaining search horizon. 
At \texttt{P04}, the historical archived-redirect action underperforms the source state in all three four-hour replays, including one run that exhausts the replay budget and degrades substantially. 
The corresponding original trajectory later reaches a stronger downstream result after 5.26 hours, indicating that the observed regret reflects both short-horizon execution instability and delayed payoff. 
At \texttt{P07}, the historical decision extends the current trajectory near the end of the 24-hour budget, while the counterfactual replay favors restoration from a strong archived state. 
These cases illustrate that effective re-anchoring should account jointly for transition reliability, remaining budget, and archived-state quality.

\begin{figure}[t]
    \centering
    \includegraphics[width=0.75\textwidth]{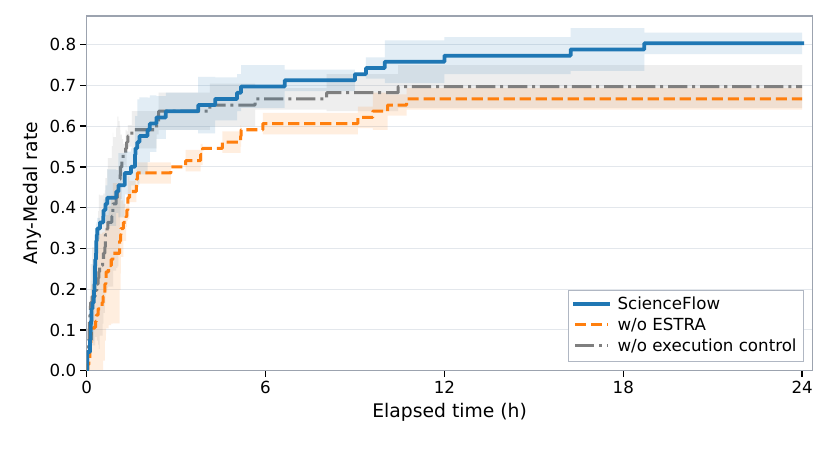}
    \caption{MLE-bench Lite mechanism ablation over a 24-hour window. 
    Curves show the cumulative Any-Medal rate averaged over three seeds, and shaded bands show the sample standard deviation.
    }
    \label{fig:mlebench-lite-ablation}
\end{figure}
\paragraph{Mechanism ablation}
We ablate ESTRA and Evidence-Aware Execution Control on the official 22-task MLE-bench Lite split under a 24-hour evaluation window. Figure~\ref{fig:mlebench-lite-ablation} shows the cumulative first-medal rate over time, where a task is counted once it first reaches Bronze, Silver, or Gold. Curves show the mean across three seeds, with shaded bands denoting the sample standard deviation. At 24 hours, the full system reaches an Any-Medal rate of $80.30 \pm 2.62\%$, compared with $66.67 \pm 2.62\%$ without ESTRA and $69.70 \pm 5.25\%$ without execution control. The difference emerges early: by 12 hours, ScienceFlow reaches $77.27 \pm 4.55\%$, while the two ablations remain at $66.67 \pm 2.62\%$ and $69.70 \pm 5.25\%$, respectively. The full system therefore improves both final Any-Medal rate and the speed of medal acquisition.

Task-level trajectories further separate the two effects. On the \emph{Jigsaw Toxic Comment Classification Challenge}, the median first-medal time increases from 2.12 hours with ScienceFlow to 5.17 hours without ESTRA, while the variant without execution control remains close to the full system at 2.41 hours. This comparison more directly associates the delay with the removal of ESTRA.

Execution control becomes more consequential on longer-running tasks. 
On the \emph{APTOS 2019 Blindness Detection} competition, removing Evidence-Aware Execution Control increases the median first-medal time from 5.18 to 8.05 hours, a 2.87-hour delay, while the slowest seed increases from 6.64 to 10.45 hours. 
The corresponding median without ESTRA is 5.91 hours, indicating that the larger delay is specifically associated with execution control. 
On the \emph{Leaf Classification} competition, ScienceFlow succeeds in all three seeds, whereas both ablations fail in all three. Therefore, this task can be treated as evidence of joint mechanism dependence rather than attributing it to either mechanism alone.

\paragraph{System-wide mechanism telemetry} 
To complement the controlled ablations, Table~\ref{tab:operational-telemetry} summarizes mechanism-level telemetry for 54 of the 75 MLE-bench tasks (72.0\%) with both ESTRA activity and workspace-state records. We focus on post-ESTRA completions, checkpoint reuse, and snapshot reuse as operational indicators of recovery and state reuse, since explicit restoration events are only sparsely logged. These statistics characterize the behavior of ScienceFlow during execution and are not used for benchmark scoring.

\begin{table}[t]
    \centering
    \small
    \caption{
    Operational telemetry for ScienceFlow mechanisms on 54 of 75 MLE-bench tasks (72.0\% coverage), including ESTRA activity, workspace-state reuse, storage efficiency, and resource control.
    }
    \label{tab:operational-telemetry}
    \setlength{\tabcolsep}{3pt}
    \begin{tabularx}{\textwidth}{@{}
        >{\RaggedRight\arraybackslash}p{0.17\textwidth}
        >{\RaggedRight\arraybackslash}p{0.27\textwidth}
        >{\RaggedRight\arraybackslash}p{0.27\textwidth}
        Y@{}}
        \toprule
        \textbf{Mechanism} & \textbf{Sub-metric} & \textbf{Value} & \textbf{Interpretation} \\
        \midrule
        \multirow{4}{*}{\makecell[l]{\textbf{ESTRA}}}
        & ESTRA decisions & 561 total & 469 continue; 92 switch. \\
        & Post-ESTRA completions & 71 completed points & Post-ESTRA outcomes. \\
        & Comparable improvements & 19 / 70 (27.1\%) & Selective gains. \\
        \midrule
        \multirow{4}{*}{\makecell[l]{\textbf{Workspace state}}}
        & Checkpoint reuse & 436 / 437 (99.8\%) & Near-complete reuse. \\
        & Snapshot object reuse & 177,234 / 221,925 (79.9\%) & Snapshot dedup. \\
        & Storage footprint & 2204.6 GiB $\to$ 189.2 GiB & Storage-covered sources. \\
        & Storage saving & 91.4\% & Delta footprint. \\
        \midrule
        \multirow{5}{*}{\makecell[l]{\textbf{Resource control}}}
        & Device-utilization samples & 20,075 samples & Device samples. \\
        & Mean utilization & 27.5\% & Observed mean. \\
        & Active device buckets & 42.7\% & $>10\%$ util. buckets. \\
        & Guard terminations & 202 stops & Guarded stops. \\
        & Low-value early stops & 157 stops & Low-value stops. \\
        \bottomrule
\end{tabularx}
\end{table}

\subsubsection{Resource Sensitivity and Efficiency} \label{sec:scaling_efficiency}

\paragraph{Resource-Constrained Performance} 
We examine ScienceFlow under tighter storage, compute, and time budgets.
We simulate a 32\,GiB per-task storage limit and compare full workspace snapshots with delta-based states.
Among the 33 tasks with available footprint measurements, all delta states remain below the limit, whereas nine medal-producing tasks exceed it when stored as full snapshots. Applying these observed violations to the three evaluation seeds reduces the Any-Medal rate from $70.22\pm1.18\%$ with delta states to $59.11\pm0.44\%$ with full snapshots. Across the measured tasks, delta states reduce the aggregate footprint from 2204.6\,GiB to 189.2\,GiB, a 91.4\% reduction.
Using the recorded task-level GPU configurations, we estimate each task to at most one GPU. A tighter compute budget produces a similar reduction in retained performance. Limiting each task to at most one GPU yields medals on 48, 48, and 47 tasks across the three runs, corresponding to $63.56\pm0.44\%$ Any-Medal, compared with $70.22\pm1.18\%$ under the reported two-GPU ceiling. 

\begin{figure}[t]
    \centering
    \begin{subfigure}[t]{0.49\textwidth}
        \centering
        \includegraphics[width=\linewidth]{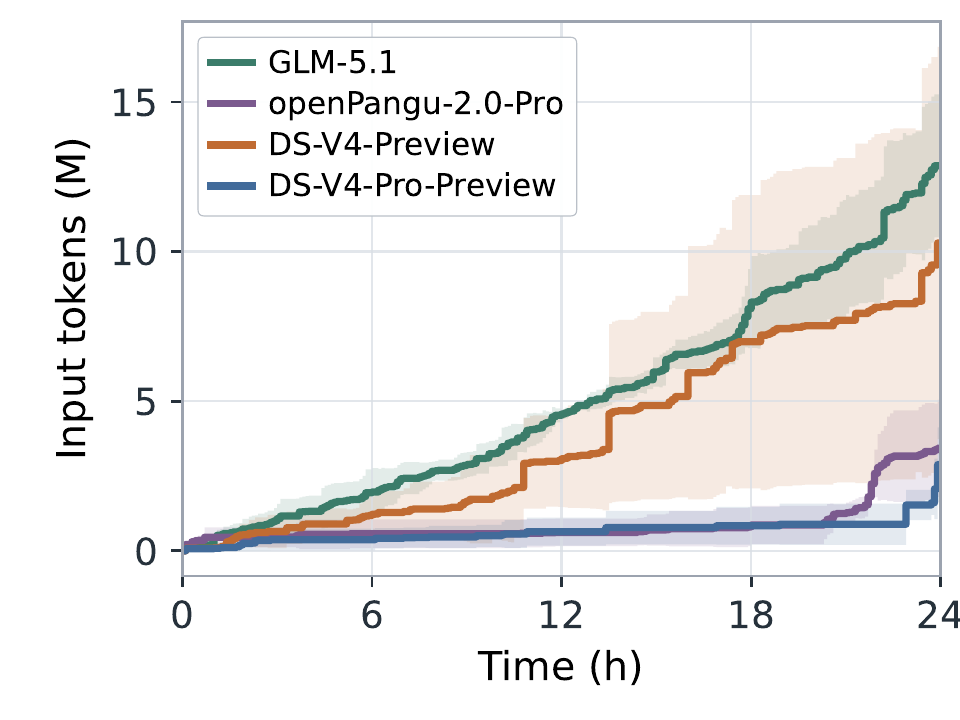}
        \caption{Cumulative input tokens.}
        \label{fig:tabular-multimodel-tokens}
    \end{subfigure}\hfill
    \begin{subfigure}[t]{0.49\textwidth}
        \centering
        \includegraphics[width=\linewidth]{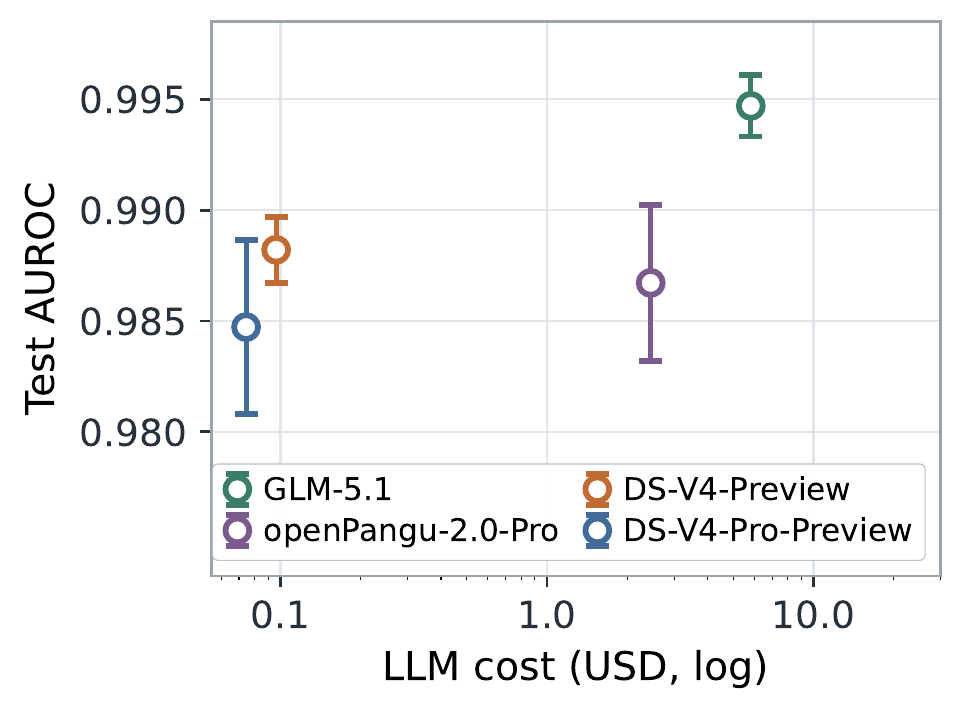}
        \caption{Mean best test and cost.}
        \label{fig:tabular-multimodel-frontier}
    \end{subfigure}
    \caption{Backbone sensitivity and efficiency on the \emph{Tabular Playground Series May 2022} task under the same two-worker, 24-hour protocol, showing cumulative input-token consumption and endpoint performance-cost trade-offs over three independent runs. ``DS'' is an abbreviation for ``DeepSeek'' in the figure legends.}
    \label{fig:tabular-multimodel}
\end{figure}

\paragraph{Backbone Sensitivity and Efficiency} 
We test ScienceFlow using four backbones: GLM-5.1, openPangu-2.0-Pro, DeepSeek-V4-Flash-Preview, and DeepSeek-V4-Pro-Preview, under the same two-worker, 24-hour protocol on the \emph{Tabular Playground Series May 2022} task.
In addition to token consumption, model-priced LLM cost is reported to reflect differences in inference pricing across backbones.
Figure~\ref{fig:tabular-multimodel} compares their 24-hour token trajectories and endpoint performance-cost trade-offs over three independent runs.
openPangu-2.0-Pro is served from our local deployment without prefix/KV-cache optimization, resulting in a 0\% cache-hit rate. 
Its cost is estimated using Huawei Cloud's official uncached Pro pricing at the corresponding request-length tier.\footnote{\href{https://www.huaweicloud.com/product/modelarts/studio.html}{Huawei Cloud ModelArts Studio pricing}, accessed August 3, 2026.} 
We use the same model-specific pricing procedure for the other backbones so that reported costs reflect their actual inference tariffs rather than token counts alone.
The results reveal distinct quality, throughput, and cost trade-offs. 
GLM-5.1 processes the most input context and achieves the highest mean test AUROC of $0.9947$, but its \$5.81 LLM cost is approximately $60\times$ that of DeepSeek-V4-Flash-Preview. DeepSeek-V4-Flash-Preview reaches $0.9882$ for \$0.096, providing the strongest quality--cost trade-off among the lower-cost backbones. openPangu-2.0-Pro consumes fewer input tokens than DeepSeek-V4-Flash-Preview but reaches a lower mean AUROC of $0.9867$ at a higher cost of \$2.45, and is therefore dominated by DeepSeek-V4-Flash-Preview on this task. DeepSeek-V4-Pro-Preview has the lowest cost at \$0.074, but also the lowest mean AUROC at $0.9847$. The wider seed variation of openPangu-2.0-Pro and DeepSeek-V4-Pro-Preview further indicates that backbone choice affects reliability as well as throughput, cost, and endpoint quality. Given only three runs on one task, these comparisons are descriptive rather than statistically conclusive.

\subsection{Mathematical and Engineering Optimization}
\label{sec:exp-opt}

We evaluate ScienceFlow in two complementary optimization settings: mathematical optimization problems and tournament scheduling optimization. 
The former includes three auditable mathematical problems, while the latter considers the easy, medium, and hard tracks of the SpOC4 KTTSP challenge. 
Both require iterative construction and executable validation of candidate solutions, but differ in problem structure, resource budget, and evaluation protocol. 
Their setups and results are presented separately below.

\subsubsection{Continuous Mathematical Optimization}
\label{subsec:exp-math}

\paragraph{Tasks}
The mathematical optimization suite contains three problems.
\emph{Circle Packing} maximizes the sum of the radii of 26 disjoint circles inside a unit square under boundary and non-overlap constraints. 
\emph{Ratio Minimization} minimizes $d_{\max}/d_{\min}$ for a configuration of 16 planar points, where $d_{\min}$ and $d_{\max}$ are the minimum and maximum pairwise distances. 
\emph{Uncertainty Inequality} searches over Hermite--Gaussian constructions to tighten a valid upper bound on the Fourier sign-uncertainty constant $C_4$; Appendix~\ref{app:math-task-definitions} gives the formal definition.

\paragraph{Protocol}
For each mathematical problem, ScienceFlow runs for a 12-hour wall-clock budget using two research workers and a single backbone per run. We evaluate DeepSeek-V4-Flash-Preview and openPangu-2.0-Pro. We rerun OpenEvolve~\citep{sharma2025openevolve} locally using its 500-iteration configuration with DeepSeek-V4-Flash-Preview, while published baselines retain their reported setups. Because budgets and worker topologies differ across systems, the comparison is descriptive rather than compute-normalized. All ScienceFlow scores are re-evaluated from saved solution artifacts using task-specific feasibility checks; Appendix~\ref{app:math-audit} provides the complete audit procedure.

\paragraph{Results}
Table~\ref{tab:math-optimization} reports the best audited score from each ScienceFlow configuration. On \emph{Circle Packing}, ScienceFlow with DeepSeek-V4-Flash-Preview obtains 2.6359830849, approximately $7.5\times10^{-9}$ above ThetaEvolve; at this scale, the two results are best interpreted as a numerical near tie. On \emph{Ratio Minimization}, ScienceFlow obtains 3.590157365310609, matching MLEvolve at the table's 12-decimal precision. On \emph{Uncertainty Inequality}, ScienceFlow with openPangu-2.0-Pro reaches 0.343293122432, reducing the strongest published Hermite-based upper bound by 2.5\%. DeepSeek-V4-Flash-Preview gives the better ScienceFlow result on the first two tasks, whereas openPangu-2.0-Pro performs better on the uncertainty task.

\begin{table}[t]
\centering
\footnotesize 
\setlength{\tabcolsep}{2pt}
\caption{Best-score comparison on three mathematical optimization tasks. Arrows indicate the optimization direction. Bold entries denote the best compared results, including ties at the reported precision; underlined entries denote the second-best results. Published baselines retain their reported protocols, while OpenEvolve is rerun locally under the configuration described in the text. Uncertainty-inequality results compare only Hermite--Gaussian constructions.}
\label{tab:math-optimization}
\begin{tabularx}{\linewidth}{@{}p{0.20\linewidth}lYr@{}}
\toprule
Task & Method & Model & Performance \\
\midrule
\multirow{7}{=}{\raggedright Circle Packing\\$\sum_i r_i$ ($\uparrow$)}
& AlphaEvolve~\citep{novikov2025alphaevolve} & Gemini 2.0 Pro + Flash & 2.6358627564 \\
& ShinkaEvolve~\citep{lange2025shinkaevolve} & Claude Sonnet 4 + GPT-4.1 family & 2.6359828390 \\
& \underline{ThetaEvolve}~\citep{wang2025thetaevolve} &
\underline{DeepSeek-R1-0528-Qwen3-8B} & \underline{2.6359830774} \\
& MLEvolve~\citep{du2026mlevolve} & Gemini-3.1-Pro-Preview & 2.6359830395 \\
\cmidrule(l){2-4}
& OpenEvolve~\citep{sharma2025openevolve} & DeepSeek-V4-Flash-Preview & 2.6344194866 \\
& \textbf{ScienceFlow (Ours)} & \textbf{DeepSeek-V4-Flash-Preview} & \textbf{2.6359830849} \\
& ScienceFlow (Ours) & openPangu-2.0-Pro & 2.6359824748 \\
\midrule
\multirow{6}{=}{\raggedright Ratio Minimization\\$d_{\max}/d_{\min}$ ($\downarrow$)}
& AlphaEvolve~\citep{novikov2025alphaevolve} & Gemini 2.0 Pro + Flash & 3.590162407473 \\
& FM Agent~\citep{li2025fmagent} & Gemini-2.5-Pro & 3.590157406159 \\
& \textbf{MLEvolve}~\citep{du2026mlevolve} &
\textbf{Gemini-3.1-Pro-Preview} & \textbf{3.590157365311} \\
\cmidrule(l){2-4}
& OpenEvolve~\citep{sharma2025openevolve} & DeepSeek-V4-Flash-Preview & 3.658667107456 \\
& \textbf{ScienceFlow (Ours)} & \textbf{DeepSeek-V4-Flash-Preview} & \textbf{3.590157365311} \\
& ScienceFlow (Ours) & openPangu-2.0-Pro & 3.590157365325 \\
\midrule
\multirow{6}{=}{\raggedright Uncertainty Inequality\\$C_4$ bound ($\downarrow$)}
& AlphaEvolve~\citep{novikov2025alphaevolve} & Gemini Pro + Flash & 0.352099104423 \\
& FM Agent~\citep{li2025fmagent} & Gemini-2.5-Pro & 0.352099104416 \\
& MLEvolve~\citep{du2026mlevolve} & Gemini-3.1-Pro-Preview & 0.352099104416 \\
\cmidrule(l){2-4}
& OpenEvolve~\citep{sharma2025openevolve} & DeepSeek-V4-Flash-Preview & 0.352581132500 \\
& \underline{ScienceFlow (Ours)} & \underline{DeepSeek-V4-Flash-Preview} &
\underline{0.348200107555} \\
& \textbf{ScienceFlow (Ours)} & \textbf{openPangu-2.0-Pro} &
\textbf{0.343293122432} \\
\bottomrule
\end{tabularx}
\end{table}

\subsubsection{Combinatorial Scheduling Optimization}
\label{subsec:exp-kttsp}

\paragraph{Task}
KTTSP is a main challenge in the fourth ESA Space Optimisation Competition (SpOC4), organized with GECCO 2026 \citep{esa_spoc4_2026}. It models a lunar-orbit collection mission in which a spacecraft must visit all targets in the shortest possible time. A solution jointly determines the target order, departure epochs, and flight durations. 
Transfers follow Lambert dynamics and must satisfy a $\Delta V$ limit, with at most $E$ higher-budget exceptions.
We evaluate the easy, medium, and hard instances through the official ESA Optimise evaluator.

\paragraph{Protocol}
ScienceFlow evaluates all three KTTSP tracks through the official SpOC4 evaluator, which accepts candidate solutions serialized in the prescribed JSON format. KTTSP-hard is assigned a separate ten-day campaign using DeepSeek-V4-Pro-Preview, DeepSeek-V4-Flash-Preview, and GLM-5.1, and provides the persisted worker trace used in the subsequent analysis. We report the public leaderboard outcomes for all three tracks. Because competing teams do not disclose uniform compute budgets or model configurations, the leaderboard provides an outcome comparison rather than a compute-normalized evaluation.

\paragraph{Overall results}
ScienceFlow obtains scores of 116.911, 234.929, and 393.229 on KTTSP-easy, KTTSP-medium, and KTTSP-hard, ranking 10th, 8th, and 3rd, respectively. Because the three tracks use different instances, their raw mission times are not directly comparable across difficulty levels. On KTTSP-hard, ScienceFlow finishes 42.786 mission days ahead of the fourth-place entry, fcmaes (436.015), and 55.583 days behind the winning entry, TGMA (337.646). The campaign-level leaderboard score is 393.229; the worker-level trace analyzed below ends at 393.229.

\begin{figure}[t]
    \centering
    \includegraphics[width=0.8\textwidth]{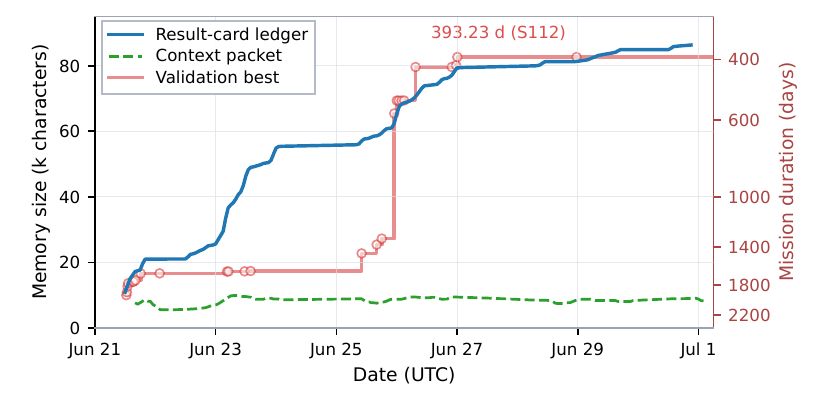}
    \caption{Persistent-memory growth and validation progress for worker W01 during the ten-day KTTSP-hard campaign. 
    The right axis uses $\log_{10}(\mathrm{score}/2000)$ coordinates with raw mission-day tick labels; lower is better. Invalid evaluation outputs are excluded.
    }
    \label{fig:kttsp-context-performance}
\end{figure}

\paragraph{Long-horizon memory folding}
Figure~\ref{fig:kttsp-context-performance} traces worker W01 during the ten-day KTTSP-hard campaign. Across 188 persisted snapshots, the result-card ledger in \texttt{.run\_results.md}, which serializes the persistent memory $m_v$, grows from 0.65k to 87.9k characters. In contrast, the 114 recorded context packets assembled at research-segment boundaries remain between 4.4k and 13.4k characters, with a median size of 8.8k and a median paired packet-to-ledger ratio of 13.7\%. This separation reflects the \textsc{Add}, \textsc{Fold}, and \textsc{Assemble} operations in Section~\ref{sec:memory}: accumulated stage records remain persistent, while each new research segment receives a bounded, anchor-specific memory view. The reported packet sizes characterize this dynamic memory component rather than the complete context $P_{n+1}$ of the next research segment.

The trace contains 185 valid evaluation stages and 26 incumbent updates. W01 reduces the objective from 1928.39 mission days at \texttt{S01} to 393.229 at \texttt{S112}, a 79.6\% reduction. Improvement continues late in the campaign: after the incumbent remains at 1641.42 days on June~25, phase-aware route construction coincides with the \texttt{S80} improvement to 573.03 days, and subsequent route and timing refinement reaches 420.27 at \texttt{S97}, 415.56 at \texttt{S109}, and 393.229 at \texttt{S112}. At this final incumbent, the most recent paired context packet contains 6.37k characters, compared with an 80.17k-character result-card ledger (7.9\%). Fourteen subsequent valid trials do not improve the incumbent, indicating late-stage saturation rather than termination at the first strong solution. This single-run trace does not establish that memory folding causes the score improvements; it instead shows that late-stage improvement and continued validation can coexist with a persistent result-card history and bounded anchor-specific context packets.
\begin{table}[t]
\centering
\footnotesize
\setlength{\tabcolsep}{5pt}
\renewcommand{\arraystretch}{1.08}
\caption{Public leaderboard for the KTTSP-hard track. Scores report total mission elapsed time in days (lower is better), and timestamps identify each team's best submission in UTC+8. Boldface marks the winning score and the ScienceFlow entry. In the model column, DS-V4-Preview denotes DeepSeek-V4-Flash-Preview, and DS-V4-Pro-Preview denotes DeepSeek-V4-Pro-Preview. Team names link to their public ESA Optimise profiles. Leaderboard snapshot accessed on July 6, 2026.}
\label{tab:kttsp_results}
\begin{tabular}{@{}cllrl@{}}
\toprule
\textbf{Rank} & \textbf{Team} & \textbf{Model} &
\textbf{Score $\downarrow$} & \textbf{Submitted (UTC+8)} \\
\midrule
\textbf{1} &
\href{https://optimise.esa.int/user/5af55cab6f2345b689bce6f1a816dcf6}{\textbf{TGMA}} &
-- & \textbf{337.646} & 2026-07-01 13:50 \\
\textbf{2} &
\href{https://optimise.esa.int/user/f5a760fd68d140bda580b85198f58b02}{AC\_TUWien} &
-- & 337.700 & 2026-07-01 01:05 \\
\textbf{3} & \textbf{ScienceFlow (Ours)} &
\makecell[l]{\textbf{DS-V4-Pro-Preview /}\\\textbf{DS-V4-Preview + GLM-5.1}} & \textbf{393.229} &
\textbf{2026-07-01 09:33} \\
4 & \href{https://optimise.esa.int/user/867404f708354718b547739c784da79c}{fcmaes} &
-- & 436.015 & 2026-06-30 14:01 \\
5 & \href{https://optimise.esa.int/user/db54c27458cb496ea71ee0cc29d9ef0a}{Team HRI} &
-- & 526.078 & 2026-06-30 21:17 \\
6 & \href{https://optimise.esa.int/user/49d02031c9c54a37a869fbb3aa6fa0a0}{SINTEF} &
-- & 587.050 & 2026-06-30 20:27 \\
7 & \href{https://optimise.esa.int/user/6b07996318ac4e1790bc6ad7ae71cf36}{\$tellaris} &
-- & 613.794 & 2026-06-30 04:57 \\
8 & \href{https://optimise.esa.int/user/48138eb8ccef4aaa9fbccfde209c5fdd}{J\&C SolExp} &
-- & 879.511 & 2026-07-05 22:52 \\
9 & \href{https://optimise.esa.int/user/dd46ba62e88c45b7958a97d529e3a220}{ScholORs\_HFUU+Sunway} &
-- & 1965.276 & 2026-05-02 18:10 \\
\bottomrule
\end{tabular}
\end{table}

\paragraph{Peer-guided search}
The same trace illustrates cross-worker coordination at research-segment boundaries. At W01's \texttt{S25} research-segment boundary, W00 had reached 521.9054 mission days while W01 remained at 1746.8935; W00's validated score prompted W01 to redirect from timing-only refinement toward route ordering. After W01 improved to 1321.8824 days, the \texttt{S79} research-segment boundary supplied a compact method summary of W00's search, prompting an operator-level redirect toward route swaps and continuous retiming. W01 subsequently reached a validated score of 420.2742 within 12 hours and 17 minutes of the \texttt{S79} decision and later improved to 393.229. The workers exchanged no executable artifacts or workspace state; all subsequent candidates were generated and validated within W01's isolated workspace. The trace therefore illustrates how peer evidence can guide route selection without direct solution reuse, rather than providing a controlled estimate of the coordination effect. Appendix~\ref{app:kttsp-peer-trace} provides the underlying records and checkpoint timeline.

\begin{figure}[H]
    \centering
    \begin{subfigure}[t]{0.49\textwidth}
        \centering
        \includegraphics[width=\linewidth]{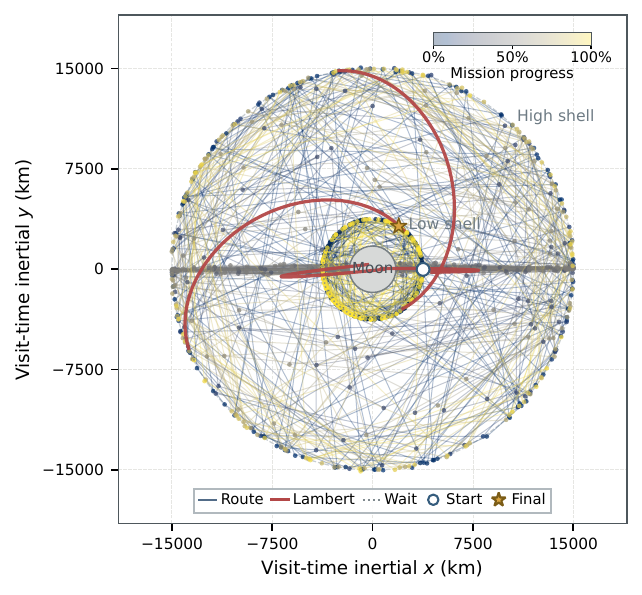}
        \caption{Initial strategy at \texttt{S78}: inclination-binned phase ordering
        (1321.8824 days).}
        \label{fig:kttsp-route-s78}
    \end{subfigure}\hfill
    \begin{subfigure}[t]{0.49\textwidth}
        \centering
        \includegraphics[width=\linewidth]{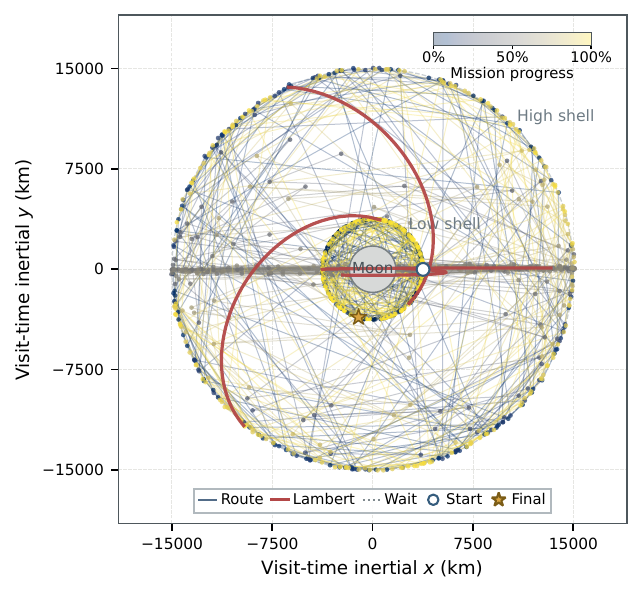}
        \caption{Refined strategy at \texttt{S112}: phase-proximal construction and
        retiming (393.229 days).}
        \label{fig:kttsp-route-final}
    \end{subfigure}
    \caption{Route and timing refinement within the W01 KTTSP-hard trajectory. Figures (a) and (b) visualize two evaluator-valid candidates that visit the same target set. 
    Target color progresses from blue to yellow with scheduled visit order, and thin links connect consecutive targets. 
    }
    \label{fig:kttsp-route-refinement}
\end{figure}

\paragraph{Route and timing refinement}
A KTTSP candidate contains two coupled decision components: a discrete visit permutation $\pi=(v_1,\ldots,v_N)$ and a continuous schedule $\tau=\{(t_i,\mathrm{tof}_i)\}_{i=1}^{N-1}$ of departure epochs and flight durations. Figure~\ref{fig:kttsp-route-refinement} compares two validated candidates generated during W01's search: an inclination-binned phase ordering and a phase-proximal construction followed by multiresolution retiming. The procedures below summarize their decision structures while omitting implementation-specific optimization loops. These strategies are task-specific artifacts produced by the research worker rather than fixed components of ScienceFlow.

\begin{tcolorbox}[
    colback=black!2!white,
    colframe=black!35!white,
    boxrule=0.45pt,
    arc=2pt,
    left=6pt,right=6pt,top=5pt,bottom=5pt,
    before skip=4pt,after skip=5pt]
\small

\textbf{Initial strategy: Inclination-binned phase ordering.}\par
{\ttfamily\scriptsize
BIN-INCLINATION $\rightarrow$ SORT-PHASE $\rightarrow$ CONCATENATE
$\rightarrow$ FIXED-ROUTE RETIME
}\par
The worker groups targets by inclination, orders each group by orbital phase, concatenates the groups, and then retimes the fixed route. This construction provides a feasible starting route but can retain unfavorable adjacencies at group boundaries.

\smallskip
{\color{black!25}\hrule}
\smallskip

\textbf{Refined strategy: Phase-proximal construction and retiming.}\par
{\ttfamily\scriptsize
PHASE-NEIGHBORS $\rightarrow$ LAMBERT-FILTER $\rightarrow$ APPEND-ROUTE
$\rightarrow$ MULTIRESOLUTION-RETIME $\rightarrow$ FORWARD-PASS
}\par
At each construction step, the worker considers phase-proximal successors, filters infeasible candidates with the Lambert evaluator, and appends a transfer-aware feasible choice. Coarse-to-fine retiming and a forward feasibility pass then update the schedule under the temporal and transfer constraints.
\end{tcolorbox}

Between the evaluator-valid \texttt{S78} and \texttt{S112} candidates, fewer than 7\% of directed consecutive target pairs are shared, while mission duration decreases from 1321.8824 to 393.229 days, a 70.25\% reduction. This descriptive comparison associates the improvement with joint route reordering and retiming rather than isolating individual search operators.

\Needspace{12\baselineskip}
\subsection{Scientific Modeling}
\label{sec:exp-sci}

\subsubsection{Setup}
\label{subsec:exp-scimodelingbench}

\paragraph{Benchmark and tasks}
SciModelingBench tests whether agents can use offline observations to find
high-value designs for a hidden scientific or engineering objective. Its 12
tasks cover seven settings across DNA binding, RNA and protein design,
superconducting materials, preclinical toxicology, and embodied control. Six
DrugMatrix tasks use the same study collection but predict different clinical
pathology endpoints.

The suite begins with scientific settings from
Design-Bench~\citep{trabucco2022designbench}, but returns to their original data
and rebuilds the tasks. Learned evaluators make validation approximate and
risky under extrapolation~\citep{beckham2024validationmetrics}; oracle
architecture and training seed can also change method
rankings~\citep{surana2024overconfident}. We instead
standardize candidate identity and repeated measurements, remove split
leakage, regenerate simulator labels with repeated rollouts, and keep only
tasks with reproducible evaluators.  
The full provenance and reconstruction details are provided in Appendix~\ref{app:scimodeling-metadata}.

\paragraph{Protocol}
Each task provides an objective, a typed manifest, and offline observations;
evaluation candidates are separated by lower-score truncation or a structured
holdout. The agent submits a ranked list of candidates. When trusted scores
cover the full domain, it may propose any valid design for black-box
optimization. Otherwise, it ranks a disclosed pool whose labels remain hidden.
Both settings use measurements, exact lookups, or simulator outcomes rather
than a fitted proxy. A fixed query budget allows iterative improvement, but
each query returns only the score of the submitted batch, not individual
candidate outcomes.

\paragraph{Performance metrics}
We use \textit{best-$K$ mean} to reward a few strong designs,
\textit{normalized enrichment} for unordered batches, and \textit{global
NDCG} for ranked submissions. For a batch $B$ of size $N$ and reference pool
$P$, normalized enrichment (NE) is defined as
$\mathrm{NE}(B)=\frac{\bar{s}_{B}-\bar{s}_{P}}{\bar{s}_{\operatorname{Top}_{N}(P)}-\bar{s}_{P}}$;
random selection has expected score zero and the ideal batch scores one. We
set split thresholds and batch sizes from data audits and preliminary
baselines before the agent runs. For the aggregate score, Random maps to zero
and the attainable oracle optimum to 100. The six DrugMatrix endpoints are
averaged as one group, weighted equally with each of the six other tasks.

\begin{table}[htbp]
    \caption{
    Task-level results under a two-hour budget with
    DeepSeek-V4-Flash-Preview. BKM, NE, and NDCG denote best-$K$ mean,
    normalized enrichment, and global NDCG. Best and second-best agent scores
    are \textbf{bolded} and \underline{underlined}.
    }
    \label{tab:scimodeling_task_results}
    \centering
    \scriptsize
    \setlength{\tabcolsep}{1.2pt}
    \renewcommand{\arraystretch}{1.0}
        \begin{tabularx}{\textwidth}{
            @{\hspace{5pt}}
            >{\raggedright\arraybackslash}p{0.16\textwidth}
            >{\centering\arraybackslash}p{0.065\textwidth}
            *{5}{>{\centering\arraybackslash}X}
            !{\vrule width 0.4pt}
            >{\centering\arraybackslash}X
            @{\hspace{5pt}}
        }
            \toprule
            \textbf{Task} & \textbf{Metric} & \textbf{Random} &
            \textbf{OpenCode} & \textbf{Pi} & \textbf{Codex} &
            \mbox{\textbf{Claude Code}} &
            \makecell{\textbf{ScienceFlow}\\\textbf{(Ours)}} \\
            \midrule
            \multicolumn{8}{l}{\textit{Standalone scientific design tasks}} \\
            \cmidrule{1-8}
            TFBind8 & BKM & 0.7533 & \underline{0.9764} & 0.9730 &
            0.9718 & 0.9648 & \textbf{0.9769} \\
            Superconductor & NDCG & 0.2800 & 0.4929 & \textbf{0.8671} &
            0.8188 & 0.8132 & \underline{0.8583} \\
            UTR MRL & NE & -0.0006 & 0.6523 & \underline{0.7533} &
            0.7424 & 0.7367 & \textbf{0.7605} \\
            TFBind10 Pho4 & NE & 0.0000 & 0.2467 & 0.2473 &
            0.2024 & \textbf{0.3197} & \underline{0.2543} \\
            GFP & NE & 0.0001 & 0.2600 & \underline{0.2692} &
            0.2586 & 0.1977 & \textbf{0.2732} \\
            Hopper Controller & NDCG & 0.1698 & \textbf{0.3917} & 0.3248 &
            0.3755 & 0.3718 & \underline{0.3756} \\
            \cmidrule{1-8}
            \addlinespace[2pt]
            \multicolumn{8}{l}{\textit{Rat clinical pathology tasks (DrugMatrix)}} \\
            \cmidrule{1-8}
            MCHC & NDCG & 0.1733 & \textbf{0.7344} & 0.3720 &
            0.6178 & 0.6503 & \underline{0.7220} \\
            MCH & NDCG & 0.2669 & 0.4147 & 0.3673 &
            \underline{0.4242} & 0.3623 & \textbf{0.4793} \\
            Creatinine & NDCG & 0.1955 & 0.6081 & \underline{0.7638} &
            0.5681 & \textbf{0.8038} & 0.6995 \\
            Sodium & NDCG & 0.1519 & 0.7619 & 0.8283 &
            0.7696 & \textbf{0.8498} & \underline{0.8488} \\
            Chloride & NDCG & 0.1689 & 0.5106 & 0.4551 &
            \textbf{0.5589} & 0.5397 & \underline{0.5488} \\
            Phosphorus & NDCG & 0.2040 & 0.4130 & \underline{0.4917} &
            0.3749 & 0.3999 & \textbf{0.5985} \\
            \bottomrule
        \end{tabularx}
\end{table}

\begin{table}[htbp]
    \caption{
    Group-balanced results. Scores map Random to 0 and the attainable optimum
    to 100. The DrugMatrix endpoints form one group, weighted equally with each
    standalone task. Average rank uses the same seven groups. Best and
    second-best values are \textbf{bolded} and \underline{underlined}; lower
    rank is better.
    }
    \label{tab:scimodeling_aggregate_results}
    \centering
    \scriptsize
    \renewcommand{\arraystretch}{0.9}
    \begin{tabular*}{\textwidth}{@{\extracolsep{\fill}}lccccc@{}}
        \toprule
        \textbf{Agent} &
        \makecell{\textbf{Standalone}\\\textbf{Design (6)}} &
        \makecell{\textbf{Rat Clinical}\\\textbf{Pathology (6)}} &
        \makecell{\textbf{Group-balanced}\\\textbf{Score} $\uparrow$} &
        \makecell{\textbf{7-group Avg.}\\\textbf{Rank} $\downarrow$} &
        \textbf{Task Wins} \\
        \midrule
        Pi & \underline{52.77} & 43.12 &
        \underline{51.39} & \underline{2.79} & 1/12 \\

        \mbox{Claude Code} & 51.65 & \underline{49.64} &
        51.37 & 3.69 & \underline{3/12} \\

        Codex & 51.49 & 43.79 &
        50.39 & 3.64 & 1/12 \\

        OpenCode & 43.84 & 46.44 &
        44.21 & 3.33 & 2/12 \\
        \midrule
        \mbox{\textbf{ScienceFlow (Ours)}} &
        \textbf{54.16} & \textbf{55.91} &
        \textbf{54.41} & \textbf{1.55} & \textbf{5/12} \\
        \bottomrule
    \end{tabular*}
\end{table}

\paragraph{Model, compute and evaluation budget}
All systems use DeepSeek-V4-Flash-Preview with the same two-hour limit and the
same task-level quota of 8--20 batch submissions, fixed before the runs. We
run each system--task pair once and report its best valid submission within
these limits. ScienceFlow splits the quota between two homogeneous workers,
which share only short summaries of their best score and method. Each
system--task run uses eight logical CPU cores, no accelerator, and less than
32\,GiB peak RSS.

\paragraph{Tools and baselines}
Agents receive an isolated workspace with shell, Python, and common data and
machine-learning packages. Hidden labels and evaluator state remain outside
the workspace, and network access is disabled. The model may still use
knowledge learned during pretraining. We compare ScienceFlow with OpenCode,
Pi, Codex, and Claude Code~\citep{opencode2026,zechner2026pi,
openai2026codex,anthropic2026claudecode}.

\subsubsection{Main Results}
\label{subsec:results-sci}

We report results at both the task level and after group-balanced aggregation.
As a reference, Random is the mean score of 5,000 uniformly sampled,
task-sized batches. Each batch contains distinct candidates and, for ranking
tasks, uses a random order.


\begin{table}[t]
    \caption{Candidate coverage in the Phosphorus case study. Repeated
    conditions across batches are counted once; coverage is relative to the
    390-condition candidate pool.}
    \label{tab:scimodeling-phosphorus-coverage}
    \centering
    \footnotesize
    \setlength{\tabcolsep}{8pt}
    \renewcommand{\arraystretch}{1.}
    \begin{tabular}{lcc}
        \toprule
        \textbf{System} & \textbf{Distinct candidates} & \textbf{Pool coverage} \\
        \midrule
        ScienceFlow & \textbf{66} & \textbf{16.9\%} \\
        Pi & 38 & 9.7\% \\
        OpenCode & 32 & 8.2\% \\
        Codex & 24 & 6.2\% \\
        Claude Code & 32 & 8.2\% \\
        \bottomrule
    \end{tabular}
\end{table}

\begin{figure}[t]
    \centering
    \begin{subfigure}[t]{0.8\textwidth}
        \centering
        \includegraphics[width=\linewidth]{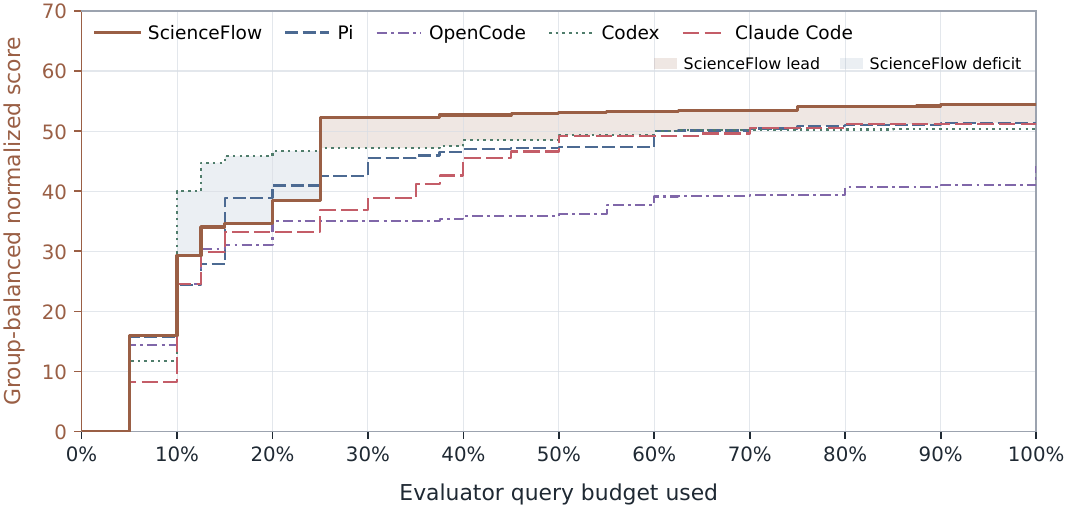}
        \caption{Group-balanced best-so-far score.}
        \label{fig:scimodeling-query-budget-score}
    \end{subfigure}
    \vspace{0.5em}
    \begin{subfigure}[t]{0.8\textwidth}
        \centering
        \includegraphics[width=\linewidth]{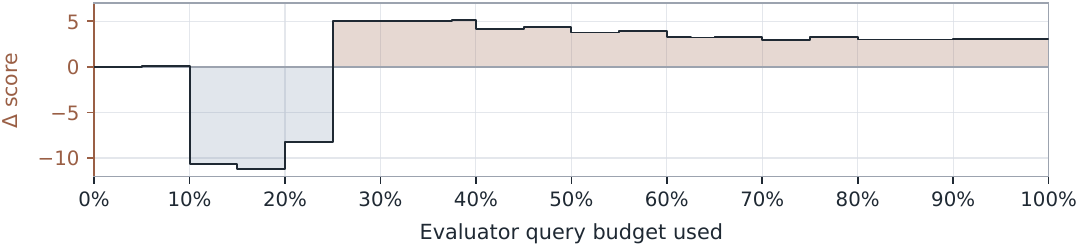}
        \caption{ScienceFlow's signed margin over the strongest baseline at
        each budget fraction.}
        \label{fig:scimodeling-query-budget-advantage}
    \end{subfigure}
    \caption{Cross-task progress under matched query budgets. The DrugMatrix
    endpoints form one group, and ScienceFlow's worker submissions are merged
    in chronological order.}
    \label{fig:scimodeling-query-budget}
\end{figure}

Table~\ref{tab:scimodeling_task_results} reports the task-level results. All
five systems beat Random on every task. ScienceFlow ranks first on five tasks
and second on six, placing in the top two on 11 of 12. It leads on TFBind8, UTR
MRL, GFP, MCH, and Phosphorus. Its largest margins over the runner-up are
0.0551 on MCH and 0.1068 on Phosphorus, while TFBind8 is nearly tied at a
margin of 0.0005. Creatinine is its only result outside the top two.

Table~\ref{tab:scimodeling_aggregate_results} summarizes the group-balanced
comparison. ScienceFlow scores highest on both groups, with 54.16 on the
standalone tasks and 55.91 on rat clinical pathology. Its overall score is
54.41, 3.02 points above Pi, and its average rank is 1.55 versus Pi's 2.79.
Each of the other agents leads at least one task, showing that their strengths
differ by domain.

\subsubsection{Analysis}
\label{subsec:analysis-scimodelingbench}

\paragraph{Case study: independent hypothesis coverage}
We use DrugMatrix Phosphorus as a case study of exploration under sparse
feedback. Each system has eight batch evaluations and ranks 16 of 390
label-hidden five-day treatment conditions by their change in phosphorus
relative to matched controls. The evaluator reports only a batch score, with
no candidate-level labels. Table~\ref{tab:scimodeling-phosphorus-coverage}
summarizes coverage across the eight batches.


ScienceFlow evaluates 66 distinct conditions, compared with 24--38 for the
baselines. Its two workers each cover 43 conditions, with only 20 shared. One
uses a general molecular and experimental-context model; the other models
repeated measurements and matches treatment and control across dose and
duration. Their partly non-overlapping coverage reflects complementary,
protocol-aware hypotheses rather than duplicate searches, under the same eight
submissions and batch-level feedback as the baselines.

\paragraph{Query-budget saturation across tasks}
We next track the aggregate best-so-far score across all 12 tasks to see how
quickly each system uses its query quota.


Figure~\ref{fig:scimodeling-query-budget} shows that Codex leads from 10\%
through 20\% of the quota, but ScienceFlow jumps from 38.43 to 52.24 at 25\%
and remains ahead thereafter. Its advantage over the strongest baseline is
about five points at that point and ends at 3.01. By 50\%, ScienceFlow, Codex,
and Claude Code are within 2.2 points of their final scores, while Pi has
reached 92\% of its final score. By 80\%, those four are within 0.36 points;
OpenCode improves later. Most systems therefore approach their final result
before exhausting the quota, leaving room for several rounds of refinement
rather than making the last query decisive.

\Needspace{10\baselineskip}
\section{Related Work}
\label{sec:related_work}

\subsection{Scientific Agents}
\label{subsec:scidiscovery}


Several recent surveys systematize this rapidly fragmenting landscape along complementary axes: a six-stage methodological pipeline and three-phase historical evolution~\citep{tie2026aiscientists}; a five-task taxonomy spanning scientific comprehension, academic survey, discovery, writing, and peer review~\citep{chen2025ai4research}; a decomposition into five foundational agent capabilities (reasoning and planning, tool integration, memory, multi-agent collaboration, and optimization) measured along an autonomy scale from ``computational oracle'' to ``generative architect''~\citep{wei2025agenticscience}; and a task-and-evaluation lifecycle view that emphasizes benchmarking, trustworthiness, and the risks of generative misuse~\citep{eger2026transforming}.

Representative end-to-end systems instantiate this paradigm across disciplines.
In machine learning, \emph{The AI Scientist}~\citep{aiscientistv1} unified idea generation, code execution, and paper writing into the first fully autonomous research loop, and its successor employs agentic tree search to explore parallel research directions and self-review the resulting manuscripts~\citep{aiscientistv2}.
In the natural sciences, agents now couple LLM reasoning with wet-lab hardware to autonomously plan, execute, and interpret experiments, from chemical synthesis~\citep{coscientist} to the de-novo design of experimentally validated nanobodies by multi-agent virtual research teams~\citep{virtuallab}, while generate--debate--evolve multi-agent systems~\citep{wei2025agenticscience} and self-evolving pipelines that expand their own toolkits~\citep{novikov2025alphaevolve} push toward long-horizon, cross-domain discovery.
The same agentic primitives---iterative refinement over literature~\citep{researchagent}, case-based reasoning~\citep{guo2024ds}, reflection~\citep{reflexion}, and executable skill libraries for open-ended embodied learning~\citep{wang2023voyager}---recur across these systems, and analogous techniques now extend to software engineering~\citep{swesearch}, algorithm and reward design~\citep{fun_search,discopop,omniepic,revolve}, and open-ended scientific discovery~\citep{o2025sparks}.

Despite this breadth, the field shares persistent limitations that the surveys converge on: brittle reproducibility and provenance, uncalibrated confidence and weak novelty validation, monolithic domain-specific architectures that fail to transfer, and long-horizon memories that cannot sustain causally-linked experiment histories~\citep{tie2026aiscientists,wei2025agenticscience,eger2026transforming}.
\emph{Machine-learning engineering} is the most mature and most heavily benchmarked slice of this paradigm: a fully simulated discovery loop that exercises experimental preparation, execution, and iterative optimization without a wet lab, with protocols such as MLE-bench~\citep{chan2025mle} providing a standardized, compute-bounded testbed.
ScienceFlow is positioned squarely within this slice, advancing the execution, optimization, and memory capabilities that the broader AI-Scientist literature identifies as the binding bottlenecks to trustworthy autonomy.

\subsection{MLE Agents}
\label{subsec:mlebench}
MLE-bench~\citep{chan2025mle} has recently emerged as the de facto evaluation protocol for autonomous machine-learning-engineering (MLE) agents. It curates 75 real-world Kaggle competitions spanning tabular, vision, language, audio, and time-series modalities, requiring agents to deliver end-to-end solutions within a fixed 24-hour budget on a single GPU.
Submissions are evaluated directly against original Kaggle medal thresholds and screened for plagiarism via automated detectors. The reference baseline---the tree-search agent AIDE~\citep{jiang2025aide} powered by \texttt{o1-preview}---attains a mere 16.9\% medal rate on the full benchmark. This pronounced gap between AI and human performance has catalyzed a rapid proliferation of novel agent architectures.

We survey this expanding landscape along three complementary dimensions: search, multi-agent collaboration, and memory.

\paragraph{Search- and evolution-driven agents}
A prominent line of work formulates solution generation as search over a structured code space. 
\citet{toledo2025ai} provide a unifying formalization of these agents; critically, they demonstrate that under AIDE's operator set, advanced search policies (e.g., Monte Carlo tree search, evolutionary algorithms) yield marginal benefits. 
Their analysis points to the \emph{operator set} as a key bottleneck and further reveals a validation-to-test generalization gap that can misdirect search trajectories. 
ML-Master~\citep{liu2025ml} expands this search process with MCTS-inspired exploration and a steerable reasoning engine conditioned on adaptive memory from parent and sibling trajectories. 
MARS~\citep{chen2026mars} combines budget-aware MCTS with an efficiency-guided reward, a modular \emph{design-decompose-implement} pipeline, and comparative reflective memory for credit assignment, achieving a 56\%-62.7\% medal rate. 
MLEvolve~\citep{du2026mlevolve} further replaces the rigid search tree with Monte Carlo \emph{graph} search, allowing non-adjacent solutions to recombine through cross-branch references. Together with an entropy-driven exploration schedule, it reaches a 65.3\% medal rate under half of the standard time budget.
Iris~\citep{fu2026iris} shifts from solution-centric search to an inquiry, which builds revision loop over a revisable information state, using epistemic actions to resolve decision-critical unknowns and reaching a 64.9\% Any-Medal rate under a 12-hour budget.

\paragraph{Hierarchical multi-agent systems}
Rather than refining a centralized search policy, an alternative paradigm decomposes the engineering workflow into collaborating specialists.
For instance, R\&D Agent~\citep{yang2025rdagent} implements a dynamic interaction between "Researcher" and "Developer" roles to periodically synthesize superior outcomes, while InternAgent~\citep{internagentteam2025internagent} proposes a closed-loop framework that tightly integrates idea generation with experimental execution.
To scale such coordination, the FM Agent~\citep{li2025fmagent} employs a multi-population island evolutionary model, augmented by expert-guided cold starts and diversity-driven sampling on a distributed asynchronous infrastructure.
Addressing cognitive biases, MLE-STAR~\citep{nam2025mle} counteracts the tendency of agents to over-rely on familiar, outdated models through web-search-based initialization and ablation-guided code refinement, integrating explicit data-leakage checkers for robustness.
More recently, AIBuildAI~\citep{zhang2026aibuildai} introduces a hierarchical Manager--Designer--Coder--Tuner topology that adaptively orchestrates seven parallel solution repositories, boosting the full-benchmark medal rate to 63.1\%.

\paragraph{Memory, knowledge, and long-horizon control}
Because long-horizon MLE tasks generate extensive execution histories that quickly saturate context windows, contemporary architectures prioritize restructuring memory over simply expanding it.
ML-Master 2.0~\citep{zhu2026mlmaster2} introduces a three-tier hierarchical cognitive cache that distills transient execution traces into cross-task insights, lifting the full-benchmark success rate to 56.4\%.
KAPSO~\citep{nadafian2026kapso} grounds code optimization in a Git-native experimentation engine and a typed knowledge graph extracted from thousands of repositories, achieving a 50.7\% medal rate.
Other works target environment reliability and state persistence: a file-as-bus framework~\citep{chen2026longhorizon} externalizes decision-relevant states for long-horizon control; Arbor~\citep{jin2026arbor} maintains a persistent hypothesis tree regulated by a held-out merge gate; and EurekAgent~\citep{xin2026eurekagent} posits that engineering the operational environment, including permissions, artifacts, and budgets, is more critical than prescribing rigid workflows.
Complementary efforts explore reinforcement learning for strategic ideation~\citep{zhang2026ideate}, lightweight ReAct-style memory tiers~\citep{chopde2025piml}, and alignment risks, demonstrating that agents can be steered to sandbag or backdoor solutions while evading language-model monitors~\citep{ward2025sabotage}.

\paragraph{Closest mechanisms and distinction}
Several adjacent systems expose individual mechanisms used by ScienceFlow,
but assign them different roles. AutoSci stores typed project artifacts and
lifecycle states in an active research memory~\citep{qian2026autosci}, while
MAGE organizes action--observation histories as an execution-state tree and
revises erroneous segments from a restored boundary~\citep{chen2026mage}.
PIVOT instead refines planned trajectories through repeated execution and
verification~\citep{zhang2026pivot}. At the systems layer, recent agentic
schedulers combine admission control with live CPU--GPU telemetry
~\citep{wang2026marscoscheduling} or use an LLM and runtime monitor to select
immediate GPU execution, queued GPU execution, or CPU offload
~\citep{lu2026agenticscheduling}. ScienceFlow does not claim that memory,
checkpointing, trajectory revision, or resource scheduling is individually
new. Its distinction is to make a recoverable executable research state---not
a dialogue trace, candidate script, workflow variable, or inference
session---the common object of persistence, trajectory adaptation through
re-anchoring, and evidence-aware execution control. Consequently, scientific route
selection remains with the research agent, whereas a separate controller owns
resource admission and termination authority using both physical constraints
and validated research progress.

\subsection{Optimization Agents}
\label{subsec:opt}
Parallel to the MLE-bench paradigm, a distinct class of autonomous agents has emerged to tackle numerical and combinatorial optimization. Rather than relying on fixed heuristics, these agents iteratively propose, evaluate, and refine solutions, and are benchmarked on standard black-box suites (e.g., BBOB~\citep{hansen2021coco}) as well as high-stakes engineering tasks. The core challenge is that an agent must operate within a limited interaction budget, extract actionable knowledge from a growing trial history, and balance exploration with exploitation. We survey this agent-driven optimization landscape along three complementary dimensions: prompt-level search, agent-guided algorithm design, and hierarchical multi-agent end-to-end pipelines.

\paragraph{Agents That Optimize at the Prompt Level}
The most direct agent architecture converts optimization into a natural-language sequential decision process. Instead of explicitly coding search operators, the agent relies entirely on prompt-based reasoning over previous trials~\citep{cheng2024black}. The OPRO framework exemplifies this approach: it encapsulates the entire optimization trajectory within a meta-prompt, enabling the agent to propose improved candidates solely by analyzing score sequences~\citep{yang2023large}. This paradigm has achieved state-of-the-art results in prompt tuning, code generation, and even mathematical reasoning. Nevertheless, pure prompt-level agents often exhibit strong warm-start behavior but stagnate in later iterations, lacking the rigorous exploration-exploitation trade-off mechanisms inherent to classical solvers. Consequently, hybrid agents are being developed that couple prompt-based candidate generation with Bayesian optimization surrogates, combining semantic insight with principled uncertainty management.

\paragraph{Agents That Design Optimization Algorithms}
A second dimension elevates the role of the agent from solution proposer to solver architect. Here, agents are embedded within classical optimization pipelines as intelligent components that design, configure, and evolve the algorithms themselves. Rather than merely executing a fixed solver, these agents actively shape the optimization process: they recommend promising candidate solutions informed by domain priors, strategically narrow or restructure the search space to focus on high-potential regions, and filter or rank evaluated candidates to guide subsequent iterations~\citep{pandit2025llmbox, yang2025reasoning}. Beyond solution-level guidance, agents also autonomously synthesize novel crossover and mutation operators by analyzing historical population statistics~\citep{suwandi2025adaptive}, dynamically adjust algorithm hyperparameters in response to convergence trends, and serve as lightweight surrogate evaluators that predict solution quality to reduce costly full simulations~\citep{yuan2026agenticgeo}. Works such as EvoLLM~\citep{lange2024large} demonstrate that an agent can iteratively refine a differential evolution algorithm across generations, effectively endowing the solver with self-evolution capabilities. This shifts the optimization problem from finding a single solution to continuously improving the algorithm that finds solutions.

\paragraph{Hierarchical Multi-Agent End-to-End Optimization Pipelines}
A hierarchical multi-agent system can decompose complex optimization workflows into specialized roles when a single agent cannot effectively manage problem formulation, constraint analysis, code generation, and evaluation.
Typical architectures decompose the workflow into specialized roles such as algorithm selection, constraint analysis, code generation, and execution evaluation, which collectively form a fully automated pipeline that covers the entire optimization life cycle~\citep{guo2026designx, baumann2025anllmbased}. These collectives have demonstrated competitive performance on standard benchmarks and have been extended to adversarial settings where multiple agents cooperatively generate adversarial examples to probe and improve model robustness. Furthermore, experience-driven multi-agent frameworks allow agents to accumulate and transfer knowledge across tasks, dramatically improving sample efficiency in repeated black-box attack scenarios. By distributing cognitive load and enabling inter-agent critique, these hierarchical architectures deliver more robust and high-quality solutions than their single-agent counterparts.


\section{Conclusion}
\label{sec:conclusion}
We have introduced \textbf{ScienceFlow}, a workspace-grounded autonomous research system for long-horizon executable work. 
ScienceFlow combines recoverable workspace states, ESTRA-governed transitions between research segments, and evidence-aware
execution control, while supporting configurable homogeneous research workers
with isolated workspaces and synchronization at research-segment boundaries. 
We evaluated ScienceFlow across machine learning engineering, mathematical and engineering optimization, and scientific modeling and design. 
ScienceFlow achieves $70.22 \pm 1.18\%$ Any-Medal on the full 75-task MLE-bench.
Across mathematical and engineering optimization, ScienceFlow matches the strongest results on circle packing and ratio minimization, improves the best published Hermite-based uncertainty bound by 2.5\%, and ranks third on KTTSP-hard.
On SciModelingBench, ScienceFlow achieves the best group-balanced score of 54.41 among the evaluated agents.
Overall, ScienceFlow delivers strong and consistent performance across a diverse range of long-horizon autonomous research tasks.


\clearpage
\phantomsection
\section*{\textcolor{huaweired}{Contributions and Acknowledgments}}
\label{sec:contributions-acknowledgments}
\addcontentsline{toc}{section}{Contributions and Acknowledgments}

Mingming Zhao, Jiqian Dong, Kangping Xu{\renewcommand{\thefootnote}{\textdagger}\footnote{Work completed during an internship.}}, Zadid Hasan, Chengrui Fan,
Shan Jiang, Shuai Mao, Yating Ling, Linyi Zou, Tailin Zhou, Yun Hin Chan,
Wenkai Zhang, Zhanhong Zhou, Guowei Huang, Hongliang Li, Wenjing Cun,
Zhitang Chen{\renewcommand{\thefootnote}{*}\footnote{Team leaders.}}, Mingxuan Yuan\textsuperscript{*}, and Yanhui Geng\textsuperscript{*}.

\clearpage

\bibliography{references}

@STRING{NeurIPS = "Advances in Neural Information Processing Systems"}

@STRING{ICML = "International Conference on Machine Learning"}

@STRING{ICLR = "International Conference on Learning Representations"}

@misc{aiscientistv2,
      title={{The AI Scientist-v2: Workshop-Level Automated Scientific Discovery via Agentic Tree Search}}, 
      author={Yutaro Yamada and Robert Tjarko Lange and Cong Lu and Shengran Hu and Chris Lu and Jakob Foerster and Jeff Clune and David Ha},
      year={2025},
      eprint={2504.08066},
      archivePrefix={arXiv},
      primaryClass={cs.AI},
      note={arXiv:2504.08066},
}

@misc{aiscientistv1,
      title={{The AI Scientist: Towards Fully Automated Open-Ended Scientific Discovery}},
      author={Chris Lu and Cong Lu and Robert Tjarko Lange and Jakob Foerster and Jeff Clune and David Ha},
      year={2024},
      eprint={2408.06292},
      archivePrefix={arXiv},
      primaryClass={cs.AI},
      note={arXiv:2408.06292},
}

@misc{tie2026aiscientists,
      title={{A Survey of AI Scientists}},
      author={Guiyao Tie and Pan Zhou and Lichao Sun},
      year={2026},
      eprint={2510.23045},
      archivePrefix={arXiv},
      primaryClass={cs.AI},
      note={arXiv:2510.23045},
}

@misc{chen2025ai4research,
      title={{AI4Research: A Survey of Artificial Intelligence for Scientific Research}},
      author={Qiguang Chen and Mingda Yang and Libo Qin and Jinhao Liu and Zheng Yan and Jiannan Guan and Dengyun Peng and Yiyan Ji and Hanjing Li and Mengkang Hu and Yimeng Zhang and Yihao Liang and Yuhang Zhou and Jiaqi Wang and Zhi Chen and Wanxiang Che},
      year={2025},
      eprint={2507.01903},
      archivePrefix={arXiv},
      primaryClass={cs.CL},
      note={arXiv:2507.01903},
}

@misc{wei2025agenticscience,
      title={{From AI for Science to Agentic Science: A Survey on Autonomous Scientific Discovery}},
      author={Jiaqi Wei and Yuejin Yang and Xiang Zhang and Yuhan Chen and Xiang Zhuang and Zhangyang Gao and Dongzhan Zhou and Guangshuai Wang and Zhiqiang Gao and Juntai Cao and Zijie Qiu and Ming Hu and Chenglong Ma and Shixiang Tang and Junjun He and Chunfeng Song and Xuming He and Qiang Zhang and Chenyu You and Shuangjia Zheng and Ning Ding and Wanli Ouyang and Nanqing Dong and Yu Cheng and Siqi Sun and Lei Bai and Bowen Zhou},
      year={2025},
      eprint={2508.14111},
      archivePrefix={arXiv},
      primaryClass={cs.LG},
      note={arXiv:2508.14111},
}

@misc{eger2026transforming,
      title={{Transforming Science with Large Language Models: A Survey on AI-assisted Scientific Discovery, Experimentation, Content Generation, and Evaluation}},
      author={Steffen Eger and Yong Cao and Jennifer D'Souza and Andreas Geiger and Christian Greisinger and Stephanie Gross and Yufang Hou and Brigitte Krenn and Anne Lauscher and Yizhi Li and Chenghua Lin and Nafise Sadat Moosavi and Wei Zhao and Tristan Miller},
      year={2026},
      eprint={2502.05151},
      archivePrefix={arXiv},
      primaryClass={cs.CL},
      note={arXiv:2502.05151},
}

@inproceedings{researchagent,
  title={Researchagent: Iterative research idea generation over scientific literature with large language models},
  author={Baek, Jinheon and Jauhar, Sujay Kumar and Cucerzan, Silviu and Hwang, Sung Ju},
  booktitle={Proceedings of the 2025 conference of the nations of the Americas chapter of the association for computational linguistics: human language technologies (volume 1: long papers)},
  pages={6709--6738},
  year={2025}
}

@article{coscientist,
  author    = {Daniil A. Boiko and Robert MacKnight and Ben Kline and Gabe Gomes},
  title     = {Autonomous chemical research with large language models},
  journal   = {Nature},
  year      = {2023},
  volume    = {624},
  number    = {7992},
  pages     = {570--578},
  issn      = {1476-4687}
}

@article{virtuallab,
  author  = {Swanson, Kyle and Wu, Wesley and Bulaong, Nash L. and Pak, John E. and Zou, James},
  title   = {The Virtual Lab of {AI} Agents Designs New {SARS-CoV-2} Nanobodies},
  journal = {Nature},
  volume  = {646},
  pages   = {716--723},
  year    = {2025},
}

@article{o2025sparks,
  title={Sparks of Science: Hypothesis Generation Using Structured Paper Data},
  author={O'Neill, Charles and Ghosal, Tirthankar and R{\u{a}}ileanu, Roberta and Walmsley, Mike and Bui, Thang and Schawinski, Kevin and Ciuc{\u{a}}, Ioana},
  journal={arXiv preprint arXiv:2504.12976},
  year={2025}
}

@inproceedings{revolve,
    title={{RE}volve: Reward Evolution with Large Language Models using Human Feedback},
    author={Rishi Hazra and Alkis Sygkounas and Andreas Persson and Amy Loutfi and Pedro Zuidberg Dos Martires},
    booktitle=ICLR,
    year={2025}
}

@article{fun_search,
  author    = {Bernardino Romera-Paredes and Mohammadamin Barekatain and Alexander Novikov and Matej Balog and M. Pawan Kumar and Emilien Dupont and Francisco J. R. Ruiz and Jordan S. Ellenberg and Pengming Wang and Omar Fawzi and Pushmeet Kohli and Alhussein Fawzi},
  title     = {Mathematical discoveries from program search with large language models},
  journal   = {Nature},
  volume    = {625},
  number    = {7995},
  pages     = {468--475},
  year      = {2024},
  issn      = {1476-4687}
}

@inproceedings{reflexion,
	title        = {Reflexion: language agents with verbal reinforcement learning},
	author       = {Shinn, Noah and Cassano, Federico and Gopinath, Ashwin and Narasimhan, Karthik and Yao, Shunyu},
	year         = 2023,
	booktitle    = NeurIPS,
	volume       = 36,
	pages        = {8634--8652}
}

@inproceedings{swesearch,
        title={{SWE}-Search: Enhancing Software Agents with Monte Carlo Tree Search and Iterative Refinement},
        author={Antonis Antoniades and Albert {\"O}rwall and Kexun Zhang and Yuxi Xie and Anirudh Goyal and William Yang Wang},
        booktitle=ICLR,
        year={2025}
}

@inproceedings{discopop,
    title={Discovering Preference Optimization Algorithms with and for Large Language Models},
    author={Chris Lu and Samuel Holt and Claudio Fanconi and Alex James Chan and Jakob Nicolaus Foerster and Mihaela van der Schaar and Robert Tjarko Lange},
    booktitle=NeurIPS ,
    year={2024}
}

@inproceedings{omniepic,
    title={{OMNI}-{EPIC}: Open-endedness via Models of human Notions of Interestingness with Environments Programmed in Code},
    author={Maxence Faldor and Jenny Zhang and Antoine Cully and Jeff Clune},
    booktitle=ICLR,
    year={2025}
}

@inproceedings{guo2024ds,
  title={{DS}-agent: Automated data science by empowering large language models with case-based reasoning},
  author={Guo, Siyuan and Deng, Cheng and Wen, Ying and Chen, Hechang and Chang, Yi and Wang, Jun},
  booktitle={Proceedings of the 41st International Conference on Machine Learning},
  volume={235},
  series={Proceedings of Machine Learning Research},
  pages={16813--16848},
  publisher={PMLR},
  year={2024},
  url={https://proceedings.mlr.press/v235/guo24b.html}
}

@article{jiang2025aide,
  title={AIDE: AI-Driven Exploration in the Space of Code},
  author={Jiang, Zhengyao and Schmidt, Dominik and Srikanth, Dhruv and Xu, Dixing and Kaplan, Ian and Jacenko, Deniss and Wu, Yuxiang},
  journal={arXiv preprint arXiv:2502.13138},
  year={2025}
}

@article{chan2025mle,
  title={Mle-bench: Evaluating machine learning agents on machine learning engineering},
  author={Chan, Jun Shern and Chowdhury, Neil and Jaffe, Oliver and Aung, James and Sherburn, Dane and Mays, Evan and Starace, Giulio and Liu, Kevin and Maksin, Leon and Patwardhan, Tejal and others},
  journal=ICLR,
  year={2025}
}

@article{wang2023voyager,
  title   = {Voyager: An Open-Ended Embodied Agent with Large Language Models},
  author  = {Guanzhi Wang and Yuqi Xie and Yunfan Jiang and Ajay Mandlekar and Chaowei Xiao and Yuke Zhu and Linxi Fan and Anima Anandkumar},
  year    = {2023},
  journal = {arXiv preprint arXiv:2305.16291}
}

@article{wang2024openhands,
  title={Openhands: An open platform for ai software developers as generalist agents},
  author={Wang, Xingyao and Li, Boxuan and Song, Yufan and Xu, Frank F and Tang, Xiangru and Zhuge, Mingchen and Pan, Jiayi and Song, Yueqi and Li, Bowen and Singh, Jaskirat and others},
  journal=ICLR,
  year={2025}
}

@article{internagentteam2025internagent,
  title={InternAgent: When Agent Becomes the Scientist -- Building Closed-Loop System from Hypothesis to Verification},
  author={{InternAgent Team}},
  journal={arXiv preprint arXiv:2505.16938 [cs.AI]},
  year={2025}
}

@article{yang2025rdagent,
  title={{R\&D-Agent}: An {LLM}-Agent Framework Towards Autonomous Data Science},
  author={Yang, Xu and Yang, Xiao and Fang, Shikai and Zhang, Yifei and Wang, Jian and Xian, Bowen and Li, Qizheng and Li, Jingyuan and Xu, Minrui and Li, Yuante and Pan, Haoran and Zhang, Yuge and Liu, Weiqing and Shen, Yelong and Chen, Weizhu and Bian, Jiang},
  journal={arXiv preprint arXiv:2505.14738},
  year={2025}
}

@article{liu2025ml,
  title={ML-Master: Towards AI-for-AI via Integration of Exploration and Reasoning},
  author={Liu, Zexi and Cai, Yuzhu and Zhu, Xinyu and Zheng, Yujie and Chen, Runkun and Wen, Ying and Wang, Yanfeng and Chen, Siheng and others},
  journal={arXiv preprint arXiv:2506.16499},
  year={2025}
}

@article{toledo2025ai,
  title={AI Research Agents for Machine Learning: Search, Exploration, and Generalization in MLE-bench},
  author={Toledo, Edan and Hambardzumyan, Karen and Josifoski, Martin and Hazra, Rishi and Baldwin, Nicolas and Audran-Reiss, Alexis and Kuchnik, Michael and Magka, Despoina and Jiang, Minqi and Lupidi, Alisia Maria and others},
  journal={arXiv preprint arXiv:2507.02554},
  year={2025}
}

@article{nam2025mle,
  title={MLE-STAR: Machine Learning Engineering Agent via Search and Targeted Refinement},
  author={Nam, Jaehyun and Yoon, Jinsung and Chen, Jiefeng and Shin, Jinwoo and Ar{\i}k, Sercan {\"O} and Pfister, Tomas},
  journal={arXiv preprint arXiv:2506.15692},
  year={2025}
}

@article{zhu2026mlmaster2,
  title={Toward Ultra-Long-Horizon Agentic Science: Cognitive Accumulation for Machine Learning Engineering},
  author={Zhu, Xinyu and Cai, Yuzhu and Liu, Zexi and Zheng, Bingyang and Wang, Cheng and Ye, Rui and Chen, Jiaao and Wang, Hanrui and Wang, Wei-Chen and Zhang, Yuzhi and Zhang, Linfeng and E, Weinan and Jin, Di and Chen, Siheng},
  journal={arXiv preprint arXiv:2601.10402},
  year={2026}
}

@article{chen2026mars,
  title={MARS: Modular Agent with Reflective Search for Automated AI Research},
  author={Chen, Jiefeng and Mishra, Bhavana Dalvi and Nam, Jaehyun and Meng, Rui and Pfister, Tomas and Yoon, Jinsung},
  journal={arXiv preprint arXiv:2602.02660},
  year={2026}
}

@article{du2026mlevolve,
  title={MLEvolve: A Self-Evolving Framework for Automated Machine Learning Algorithm Discovery},
  author={Du, Shangheng and Yan, Xiangchao and Shi, Jinxin and Cao, Zongsheng and Feng, Shiyang and Liang, Zichen and Sun, Boyuan and Peng, Tianshuo and Zhou, Yifan and Li, Xin and Zhou, Jie and He, Liang and Zhang, Bo and Bai, Lei},
  journal={arXiv preprint arXiv:2606.06473},
  year={2026}
}

@article{fu2026iris,
  title={Beyond Solution-Centric Search: Adaptive Inquiry and Knowledge Revision for Autonomous {ML} Engineering},
  author={Fu, Shaokang and Tao, Yulong and Jin, Linbo and Zhao, Jiarong and Shi, Qiming and Pan, Tianjun and Li, Haonan and Wang, Chengyu and Wu, Jia and Huo, Chengfu},
  journal={arXiv preprint arXiv:2608.02143},
  year={2026}
}

@article{qian2026autosci,
  title={AutoSci: A Memory-Centric Agentic System for the Full Scientific Research Lifecycle},
  author={Qian, Weitong and Xu, Beicheng and Xie, Zhongao and Fan, Bowen and others},
  journal={arXiv preprint arXiv:2605.31468},
  year={2026}
}

@article{chen2026mage,
  title={Beyond Semantic Organization: Memory as Execution State Management for Long-Horizon Agents},
  author={Chen, Yaoqi and Lai, Haibin and Feng, Yuru and Han, Chuyu and others},
  journal={arXiv preprint arXiv:2606.06090},
  year={2026}
}

@article{zhang2026pivot,
  title={PIVOT: Bridging Planning and Execution in LLM Agents via Trajectory Refinement},
  author={Zhang, Tuo and Popa, Alin-Ionut and Xu, Yan and Song, Rui and Dimitriadis, Dimitrios},
  journal={arXiv preprint arXiv:2605.11225},
  year={2026}
}

@article{wang2026marscoscheduling,
  title={MARS: Efficient, Adaptive Co-Scheduling for Heterogeneous Agentic Systems},
  author={Wang, Yifei and Ye, Hancheng and Xu, Yechen and Guo, Cong and Wei, Chiyue and Wang, Qinsi and Li, Dongting and Chen, Tingjun and Li, Hai and Zhuo, Danyang and Chen, Yiran},
  journal={arXiv preprint arXiv:2604.26963},
  year={2026}
}

@article{lu2026agenticscheduling,
  title={Agentic CPU-GPU Scheduling for Heterogeneous AI Workloads},
  author={Lu, Tianxi and Reda, Sherief},
  journal={arXiv preprint arXiv:2607.22242},
  year={2026}
}

@article{zhang2026aibuildai,
  title={AIBuildAI: An AI Agent for Automatically Building AI Models},
  author={Zhang, Ruiyi and Qin, Peijia and Cao, Qi and Zhang, Li and Xie, Pengtao},
  journal={arXiv preprint arXiv:2604.14455},
  year={2026}
}

@article{li2025fmagent,
  author       = {Annan Li and
                  Chufan Wu and
                  Zengle Ge and
                  Yee Hin Chong and
                  Zhinan Hou and
                  Lizhe Cao and
                  Cheng Ju and
                  Jianmin Wu and
                  Huaiming Li and
                  Haobo Zhang and
                  Shenghao Feng and
                  Mo Zhao and
                  Fengzhi Qiu and
                  Rui Yang and
                  Mengmeng Zhang and
                  Wenyi Zhu and
                  Yingying Sun and
                  Quan Sun and
                  Shunhao Yan and
                  Danyu Liu and
                  Dawei Yin and
                  Dou Shen},
  title        = {The {FM} Agent},
  journal      = {CoRR},
  volume       = {abs/2510.26144},
  year         = {2025}
}

@misc{sharma2025openevolve,
  author       = {Asankhaya Sharma},
  title        = {{OpenEvolve}: An Open-Source Evolutionary Coding Agent},
  year         = {2025},
  howpublished = {GitHub software},
  url          = {https://github.com/codelion/openevolve},
  note         = {Accessed August 12, 2026}
}

@misc{botla2025pievolve,
  author       = {Botla, Sai Kiran and Sankar, Kirubanath and
                  Chopde, Abhishek and Pettiwala, Fardeen},
  title        = {{Pi-Evolve}: Long-Horizon Evolutionary Optimization for
                  Autonomous Scientific Discovery},
  year         = {2025},
  howpublished = {GitHub software},
  url          = {https://github.com/FractalAIResearchLabs/PiEvolve},
  note         = {Accessed August 12, 2026}
}

@article{nadafian2026kapso,
  title={KAPSO: A Knowledge-grounded framework for Autonomous Program Synthesis and Optimization},
  author={Nadafian, Alireza and Mohammadshahi, Alireza and Yazdani, Majid},
  journal={arXiv preprint arXiv:2601.21526},
  year={2026}
}

@article{chen2026longhorizon,
  title={Toward Autonomous Long-Horizon Engineering for ML Research},
  author={Chen, Guoxin and Chen, Jie and Chen, Lei and Zhao, Jiale and Meng, Fanzhe and Zhao, Wayne Xin and Song, Ruihua and Chen, Cheng and Wen, Ji-Rong and Jia, Kai},
  journal={arXiv preprint arXiv:2604.13018},
  year={2026}
}

@article{jin2026arbor,
  title={Toward Generalist Autonomous Research via Hypothesis-Tree Refinement},
  author={Jin, Jiajie and Hu, Yuyang and Qiu, Kai and Dai, Qi and Luo, Chong and Dong, Guanting and Li, Xiaoxi and Zhao, Tong and Ma, Xiaolong and Zhang, Gongrui and others},
  journal={arXiv preprint arXiv:2606.11926},
  year={2026}
}

@article{xin2026eurekagent,
  title={EurekAgent: Agent Environment Engineering is All You Need for Autonomous Scientific Discovery},
  author={Xin, Amy and Siow, Jiening and Wang, Junjie and Yao, Zijun and Song, Jian and Hou, Lei and Li, Juanzi and Zhang, Fanjin},
  journal={arXiv preprint arXiv:2606.13662},
  year={2026}
}

@inproceedings{chopde2025piml,
  title={{PiML}: Automated Machine Learning Workflow Optimization using {LLM} Agents},
  author={Chopde, Abhishek and Pettiwala, Fardeen and Kirubananth, Sankar and Botla, Sai Kiran and Kethan, Pachipulusu Ayyappa},
  booktitle={Proceedings of the Fourth International Conference on Automated Machine Learning},
  volume={293},
  series={Proceedings of Machine Learning Research},
  pages={1/1--42},
  publisher={PMLR},
  year={2025},
  url={https://proceedings.mlr.press/v293/chopde25a.html}
}

@article{zhang2026ideate,
  title={Learning to Ideate for Machine Learning Engineering Agents},
  author={Zhang, Yunxiang and Zhou, Kang and Xu, Zhichao and Ramnath, Kiran and Zhou, Yun and Woo, Sangmin and Ding, Haibo and Cheong, Lin Lee},
  journal={arXiv preprint arXiv:2601.17596},
  year={2026}
}

@article{ward2025sabotage,
  title={{CTRL-ALT-DECEIT}: Sabotage Evaluations for Automated {AI} {R\&D}},
  author={Ward, Francis Rhys and van der Weij, Teun and G{\'a}bor, Hanna and Martin, Sam and Mehta Moreno, Raja and Lidar, Harel and Makower, Louis and Jodrell, Thomas and Robson, Lauren},
  journal=NeurIPS,
  year={2025}
}

@misc{neo2025,
  title        = {Neo: Next-Generation AI Agents},
  author       = {{HeyNeo Team}},
  howpublished = {\url{https://heyneo.so/blog}},
  year = {2025}
}

@misc{novikov2025alphaevolve,
  title={AlphaEvolve: A Coding Agent for Scientific and Algorithmic Discovery},
  author={Novikov, Alexander and V{\~u}, Ng{\^a}n and Eisenberger, Marvin and Dupont, Emilien and Huang, Po-Sen and Wagner, Adam Zsolt and Shirobokov, Sergey and Kozlovskii, Borislav and Ruiz, Francisco J. R. and Mehrabian, Abbas and Kumar, M. Pawan and See, Abigail and Chaudhuri, Swarat and Holland, George and Davies, Alex and Nowozin, Sebastian and Kohli, Pushmeet and Balog, Matej},
  year={2025},
  eprint={2506.13131},
  archivePrefix={arXiv},
  primaryClass={cs.AI},
  url={https://arxiv.org/abs/2506.13131}
}

@inproceedings{yang2023large,
  title={Large Language Models as Optimizers},
  author={Yang, Chengrun and Wang, Xuezhi and Lu, Yifeng and Liu, Hanxiao and Le, Quoc V and Zhou, Denny and Chen, Xinyun},
  booktitle={The Twelfth International Conference on Learning Representations},
  year={2024}
}

@inproceedings{lange2024large,
  title={Large language models as evolution strategies},
  author={Lange, Robert and Tian, Yingtao and Tang, Yujin},
  booktitle={Proceedings of the Genetic and Evolutionary Computation Conference Companion},
  pages={579--582},
  year={2024}
}

@article{goncalves2017hermite,
  title={Hermite polynomials, linear flows on the torus, and an uncertainty principle for roots},
  author={Gon{\c{c}}alves, Felipe and e Silva, Diogo Oliveira and Steinerberger, Stefan},
  journal={Journal of Mathematical Analysis and Applications},
  volume={451},
  number={2},
  pages={678--711},
  year={2017},
  publisher={Elsevier},
  doi={10.1016/j.jmaa.2017.02.030},
  url={https://doi.org/10.1016/j.jmaa.2017.02.030}
}

@article{lange2025shinkaevolve,
  title={Shinkaevolve: Towards open-ended and sample-efficient program evolution},
  author={Lange, Robert Tjarko and Imajuku, Yuki and Cetin, Edoardo},
  journal={arXiv preprint arXiv:2509.19349},
  year={2025} 
}

@article{wang2025thetaevolve,
  title={ThetaEvolve: Test-time Learning on Open Problems},
  author={Wang, Yiping and Su, Shao-Rong and Zeng, Zhiyuan and Xu, Eva and Ren, Liliang and Yang, Xinyu and Huang, Zeyi and He, Xuehai and Ma, Luyao and Peng, Baolin and Cheng, Hao and He, Pengcheng and Chen, Weizhu and Wang, Shuohang and Du, Simon Shaolei and Shen, Yelong},
  journal={arXiv preprint arXiv:2511.23473},
  year={2025}
}

@article{hansen2021coco,
  title={COCO: A platform for comparing continuous optimizers in a black-box setting},
  author={Hansen, Nikolaus and Auger, Anne and Ros, Raymond and Mersmann, Olaf and Tu{\v{s}}ar, Tea and Brockhoff, Dimo},
  journal={Optimization Methods and Software},
  volume={36},
  number={1},
  pages={114--144},
  year={2021},
  publisher={Taylor \& Francis}
}

@article{guo2026designx,
  title={{DesignX}: Human-Competitive Algorithm Designer for Black-Box Optimization},
  author={Guo, Hongshu and Ma, Zeyuan and Ma, Yining and Zhang, Xinglin and Chen, Wei-Neng and Gong, Yue-Jiao},
  journal={Advances in Neural Information Processing Systems},
  volume={38},
  pages={6582--6615},
  year={2025}
}

@inproceedings{cheng2024black,
  title={Black-box prompt optimization: Aligning large language models without model training},
  author={Cheng, Jiale and Liu, Xiao and Zheng, Kehan and Ke, Pei and Wang, Hongning and Dong, Yuxiao and Tang, Jie and Huang, Minlie},
  booktitle={Proceedings of the 62nd Annual Meeting of the Association for Computational Linguistics (Volume 1: Long Papers)},
  pages={3201--3219},
  year={2024}
}

@inproceedings{baumann2025anllmbased,
author = {Baumann, Jill and Kramer, Oliver},
title = {An LLM-Based Multi-Agent Framework for Evolutionary Blackbox Optimization},
year = {2025},
isbn = {9798400714641},
publisher = {Association for Computing Machinery},
address = {New York, NY, USA},
doi = {10.1145/3712255.3726575},
booktitle = {Proceedings of the Genetic and Evolutionary Computation Conference Companion},
pages = {671–674},
numpages = {4},
location = {NH Malaga Hotel, Malaga, Spain},
series = {GECCO '25 Companion}
}

@inproceedings{
pandit2025llmbox,
title={{LLM}-Box : An Agentic Framework for Guided Black-Box Optimization in Mapping {LLM}s onto Specialized Hardware Accelerators},
author={Sujay Pandit and Akanksha Jain and Rami Cohen and Zhijie Deng and Sagar Karandikar and Sagi Perel and Anand Raghunathan and Parthasarathy Ranganathan},
booktitle={Machine Learning for Systems 2025},
year={2025}
}

@inproceedings{
suwandi2025adaptive,
title={Adaptive Kernel Design for Bayesian Optimization Is a Piece of {CAKE} with {LLM}s},
author={Richard Cornelius Suwandi and Feng Yin and Juntao Wang and Renjie Li and Tsung-Hui Chang and Sergios Theodoridis},
booktitle={The Thirty-ninth Annual Conference on Neural Information Processing Systems},
year={2025}
}

@article{yang2025reasoning,
  title={Reasoning BO: Enhancing Bayesian optimization with long-context reasoning power of LLMs},
  author={Yang, Zhuo and Wang, Daolang and Ge, Lingli and Wang, Beilun and Fu, Tianfan and Li, Yuqiang},
  journal={arXiv preprint arXiv:2505.12833},
  year={2025}
}

@article{yuan2026agenticgeo,
  title={AgenticGEO: A Self-Evolving Agentic System for Generative Engine Optimization},
  author={Yuan, Jiaqi and Wang, Jialu and Wang, Zihan and Sun, Qingyun and Wang, Ruijie and Li, Jianxin},
  journal={arXiv preprint arXiv:2603.20213},
  year={2026}
}

@inproceedings{trabucco2022designbench,
  title={{Design-Bench}: Benchmarks for Data-Driven Offline Model-Based Optimization},
  author={Trabucco, Brandon and Geng, Xinyang and Kumar, Aviral and Levine, Sergey},
  booktitle={Proceedings of the 39th International Conference on Machine Learning},
  series={Proceedings of Machine Learning Research},
  volume={162},
  pages={21658--21676},
  year={2022},
  publisher={PMLR},
  url={https://proceedings.mlr.press/v162/trabucco22a.html}
}

@article{barrera2016variation,
  title={Survey of Variation in Human Transcription Factors Reveals Prevalent {DNA} Binding Changes},
  author={Barrera, Luis A. and Vedenko, Anastasia and Kurland, Jesse V. and others},
  journal={Science},
  volume={351},
  number={6280},
  pages={1450--1454},
  year={2016},
  doi={10.1126/science.aad2257}
}

@article{hume2015uniprobe,
  title={{UniPROBE}, Update 2015: New Tools and Content for the Online Database of Protein-Binding Microarray Data on Protein--{DNA} Interactions},
  author={Hume, Maxwell A. and Barrera, Luis A. and Gisselbrecht, Stephen S. and Bulyk, Martha L.},
  journal={Nucleic Acids Research},
  volume={43},
  number={D1},
  pages={D117--D122},
  year={2015},
  doi={10.1093/nar/gku1045}
}

@article{le2018binding,
  title={Comprehensive, High-Resolution Binding Energy Landscapes Reveal Context Dependencies of Transcription Factor Binding},
  author={Le, Daniel D. and Shimko, Tyler C. and Aditham, Arjun K. and Keys, Allison M. and Longwell, Scott A. and Orenstein, Yaron and Fordyce, Polly M.},
  journal={Proceedings of the National Academy of Sciences},
  volume={115},
  number={16},
  pages={E3702--E3711},
  year={2018},
  doi={10.1073/pnas.1715888115}
}

@misc{fordycelab2018betseq,
  author       = {{Fordyce Lab}},
  title        = {{BET-seq Processed Data}},
  year         = {2018},
  howpublished = {figshare dataset},
  doi          = {10.6084/m9.figshare.5728467.v1},
  url          = {https://doi.org/10.6084/m9.figshare.5728467.v1},
  note         = {Dataset; CC BY 4.0}
}

@article{hamidieh2018superconductor,
  title={A Data-Driven Statistical Model for Predicting the Critical Temperature of a Superconductor},
  author={Hamidieh, Kam},
  journal={Computational Materials Science},
  volume={154},
  pages={346--354},
  year={2018},
  doi={10.1016/j.commatsci.2018.07.052}
}

@misc{hamidieh2018superconductordata,
  author       = {Hamidieh, Kam},
  title        = {Superconductivty Data},
  year         = {2018},
  howpublished = {{UCI Machine Learning Repository}},
  doi          = {10.24432/C53P47},
  url          = {https://doi.org/10.24432/C53P47},
  note         = {Dataset; CC BY 4.0}
}

@article{sample2019utr,
  title={Human 5' {UTR} Design and Variant Effect Prediction from a Massively Parallel Translation Assay},
  author={Sample, Paul J. and Wang, Ban and Reid, David W. and Presnyak, Vlad and McFadyen, Iain J. and Morris, David R. and Seelig, Georg},
  journal={Nature Biotechnology},
  volume={37},
  number={7},
  pages={803--809},
  year={2019},
  doi={10.1038/s41587-019-0164-5}
}

@misc{sample2018utrdata,
  author       = {Sample, Paul J. and Wang, Ban and Seelig, Georg},
  title        = {Human 5' {UTR} Design and Variant Effect Prediction from a Massively Parallel Translation Assay},
  year         = {2018},
  howpublished = {{NCBI Gene Expression Omnibus}},
  url          = {https://www.ncbi.nlm.nih.gov/geo/query/acc.cgi?acc=GSE114002},
  note         = {Dataset; accession GSE114002}
}

@article{sarkisyan2016gfp,
  title={Local Fitness Landscape of the Green Fluorescent Protein},
  author={Sarkisyan, Karen S. and Bolotin, Dmitry A. and Meer, Margarita V. and others},
  journal={Nature},
  volume={533},
  number={7603},
  pages={397--401},
  year={2016},
  doi={10.1038/nature17995}
}

@misc{bolotin2016gfpdata,
  author       = {Bolotin, Dmitry},
  title        = {Local Fitness Landscape of the Green Fluorescent Protein},
  year         = {2016},
  howpublished = {figshare dataset},
  doi          = {10.6084/m9.figshare.3102154.v1},
  url          = {https://doi.org/10.6084/m9.figshare.3102154.v1},
  note         = {Dataset; CC BY 4.0}
}

@inproceedings{todorov2012mujoco,
  title={{MuJoCo}: A Physics Engine for Model-Based Control},
  author={Todorov, Emanuel and Erez, Tom and Tassa, Yuval},
  booktitle={IEEE/RSJ International Conference on Intelligent Robots and Systems},
  pages={5026--5033},
  year={2012},
  publisher={IEEE},
  doi={10.1109/IROS.2012.6386109}
}

@misc{ntp2023drugmatrix,
  title={{DrugMatrix}},
  author={{National Toxicology Program (NTP)}},
  year={2023},
  howpublished={Chemical Effects in Biological Systems (CEBS)},
  publisher={National Institute of Environmental Health Sciences},
  doi={10.22427/NTP-DATA-107-022-001-000-3},
  url={https://doi.org/10.22427/NTP-DATA-107-022-001-000-3}
}

@article{beckham2024validationmetrics,
  title={Exploring Validation Metrics for Offline Model-Based Optimisation with Diffusion Models},
  author={Beckham, Christopher and Pich{\'e}, Alexandre and V{\'a}zquez, David and Pal, Christopher},
  journal={Transactions on Machine Learning Research},
  year={2024},
  url={https://openreview.net/forum?id=wC4ZID0H9a}
}

@inproceedings{surana2024overconfident,
  title={Overconfident Oracles: Limitations of In Silico Sequence Design Benchmarking},
  author={Surana, Shikha and Grinsztajn, Nathan and Atkinson, Timothy and Duckworth, Paul and Barrett, Thomas D.},
  booktitle={ICML 2024 Workshop on AI for Science},
  year={2024},
  url={https://openreview.net/forum?id=fPBCnJKXUb}
}

@misc{scimodelingbench2026,
  author       = {{SciModelingBench}},
  title        = {{SciModelingBench Design-Bench Data}},
  year         = {2026},
  howpublished = {Hugging Face dataset},
  note         = {Hugging Face dataset, version 0.10.0}
}

@misc{opencode2026,
  author       = {{OpenCode Contributors}},
  title        = {{OpenCode}: The Open Source {AI} Coding Agent},
  year         = {2026},
  howpublished = {npm software release},
  url          = {https://www.npmjs.com/package/opencode-ai/v/1.18.4},
  note         = {Version 1.18.4; accessed August 12, 2026}
}

@misc{zechner2026pi,
  author       = {Zechner, Mario},
  title        = {{Pi Coding Agent}},
  year         = {2026},
  howpublished = {npm software release},
  url          = {https://www.npmjs.com/package/@earendil-works/pi-coding-agent/v/0.81.1},
  note         = {Version 0.81.1; accessed August 12, 2026}
}

@misc{openai2026codex,
  author       = {{OpenAI}},
  title        = {{Codex CLI}},
  year         = {2026},
  howpublished = {npm software release},
  url          = {https://www.npmjs.com/package/@openai/codex/v/0.144.5},
  note         = {Version 0.144.5; accessed August 12, 2026}
}

@misc{anthropic2026claudecode,
  author       = {{Anthropic}},
  title        = {{Claude Code}},
  year         = {2026},
  howpublished = {npm software release},
  url          = {https://www.npmjs.com/package/@anthropic-ai/claude-code/v/2.1.72},
  note         = {Version 2.1.72; accessed August 12, 2026}
}

@misc{openai2026mlebenchleaderboard,
  author       = {{OpenAI}},
  title        = {{MLE-bench Leaderboard}},
  year         = {2026},
  howpublished = {GitHub repository},
  url          = {https://github.com/openai/mle-bench},
  note         = {Accessed August 12, 2026}
}

@misc{esa_spoc4_2026,
  author       = {{European Space Agency, Advanced Concepts Team}},
  title        = {{SpOC 4: Space Logistics}},
  year         = {2026},
  howpublished = {European Space Agency competition webpage},
  url          = {https://www.esa.int/gsp/ACT/news/spoc-2026/},
  note         = {Accessed August 12, 2026}
}
\bibliographystyle{iclr2026_conference}

\clearpage
\appendix
\addtocontents{toc}{\protect\setcounter{tocdepth}{1}}
\appendix
\setcounter{table}{0}
\setcounter{figure}{0}
\setcounter{equation}{0}

\renewcommand{\thetable}{A\arabic{table}}
\renewcommand{\thefigure}{A\arabic{figure}}
\renewcommand{\theequation}{A\arabic{equation}}
\section{Implementation Details}
\label{app:implementation-details}

\subsection{Context Construction and Interaction Pipeline}
\label{app:context-pipeline}

Within each research segment, ScienceFlow follows a standard reasoning--action--observation loop. At the start of the segment, it assembles the model-facing context in a fixed order from a stable prefix, the selected anchor state, and the ESTRA direction; segment-local interaction history is then accumulated during tool use. Table~\ref{tab:context-pipeline-components} details these components and their update scopes, complementing the context assembly defined in Section~\ref{sec:memory}.

\begin{table}[H]
    \centering
    \small
    \caption{Model-facing context components in assembly order. The first four
    components form the cache-friendly prefix; subsequent components vary with
    the recoverable state or current research segment.}
    \label{tab:context-pipeline-components}
    \begin{tabular}{@{}p{0.18\linewidth}p{0.55\linewidth}p{0.18\linewidth}@{}}
        \toprule
        \textbf{Component} & \textbf{Contents} & \textbf{Update scope} \\
        \midrule
        $P_{\mathrm{worker}}$ & Generic research-worker and workspace-interaction policy & Worker lifecycle \\
        $P_{\mathrm{runtime}}$ & Runtime behavior, safety, and tool-use policy & Worker lifecycle \\
        $P_{\mathrm{tools}}$ & Tool names, descriptions, argument schemas, and call policy & Worker lifecycle \\
        $P_{\mathrm{task}}$ & Objective, evaluator contract, budget envelope, task adapter, and initial analysis guidance & Task run \\
        $P_{\mathrm{anchor}}(a_n)$ & Workspace view, folded or unfolded memory, validation evidence, and resource status for the selected anchor & Research segment \\
        $P_{\mathrm{dir}}(d_n)$ & Extend-or-redirect instruction selected by ESTRA & Research segment \\
        $h_{v,j}$ & Reasoning outputs, tool calls, and compact observations & Tool-use round \\
        \bottomrule
    \end{tabular}
\end{table}

\paragraph{Initial segment.}
The initial segment combines the stable context $P_{\mathrm{stable}}$ with a compact view of the initial workspace $W_0$. The task component specifies the objective, evaluator contract, resource budget, and required artifact interface. For MLE-bench, the initial guidance prioritizes file and modality inspection, leakage-safe validation, and a bounded baseline before expensive training. For optimization tasks, it specifies the objective, constraints, evaluator, and candidate-solution contract. These instructions define the initial inspection priorities, while subsequent actions remain selected by the worker's local policy.

\begin{promptbox}{gray!65!black}{Schematic context views for initial and subsequent segments}
\begin{Verbatim}[fontsize=\scriptsize,breaklines]
[stable prefix: P_stable]
worker:  generic research-worker policy
runtime: execution and tool-use policy
tools:   read / search / edit / write / execute / inspect
task:    objective + evaluator + budget + artifact contract

[initial segment]
workspace_view: View(W_0)
instruction: inspect task -> analyze data -> build bounded baseline

[subsequent segment]
anchor:         current_state | archived_state
workspace_view: View(W(a_n))
memory_view:    Fold(m(a_n); B_mem) [+ optional Unfold(...)]
validation:     e(a_n)
resources:      l(a_n) + remaining budget B_t
direction:      extend | redirect
closed_branch:  Fold(completed exploration)

[segment-local history: h_n,j]
reasoning -> tool action -> observation -> workspace update -> ...
\end{Verbatim}
\end{promptbox}

\paragraph{Stage-gate and research-segment transitions.}
During forward research, a task-specific result signal invokes the stage gate. The gate records a result card $q_v$, snapshots the active workspace as $W_v$, stores validation evidence $e_v$ and resource records $\ell_v$, and inserts the resulting state $s_v$ into the archive. The worker then resumes the current research segment, with the stage-gate exchange excluded from its segment-local history.

A text-only response or context-capacity limit closes the current segment and invokes ESTRA to select the next anchor $a_n$ and direction $d_n$. A current-state anchor retains the live workspace, whereas an archived-state anchor restores the corresponding workspace exactly and folds the post-anchor branch into completed evidence. The underlying result cards remain indexed and can later be retrieved through \textsc{Unfold}.

\paragraph{Stable-prefix caching.}
The stable prefix$P_{\mathrm{stable}}=P_{\mathrm{worker}}\oplus P_{\mathrm{runtime}}\oplus P_{\mathrm{tools}}\oplus P_{\mathrm{task}}$ is placed before all anchor-specific and segment-local content. The worker policy, runtime instructions, and tool schemas remain fixed over the worker lifecycle, while the task contract remains fixed within a task run. Workspace views, memory, validation evidence, resource status, and ESTRA directions are appended afterward and updated at research-segment boundaries. This ordering preserves a common prefix across tool rounds and research segments. Backend-reported cache telemetry is recorded when available, while cache availability does not alter the research-state transitions.

\subsection{ESTRA Decision and Re-Anchoring Pipeline}
\label{app:estra-pipeline}

ESTRA, introduced in Section~\ref{subsubsec:reanchoring}, jointly selects an execution anchor $a_n$ and a research direction $d_n$ at each research-segment boundary. The anchor determines whether the next research segment retains the current workspace or restores an archived executable state, while the direction determines whether research extends the selected route or redirects from it. When an archived anchor is selected, exploration performed after that anchor is folded into completed evidence before the next research segment begins.

\begin{table}[H]
    \centering
    \small
    \caption{Evidence provided to ESTRA at a research-segment boundary.
    Optional components are included only when corresponding records are
    available.}
    \label{tab:estra-prompt-components}
    \begin{tabular}{@{}p{0.24\linewidth}p{0.7\linewidth}@{}}
        \toprule
        \textbf{Component} & \textbf{Contents} \\
        \midrule
        Current state &
        Active workspace summary, latest result card, and current validation evidence \\
        Archived anchors &
        Recoverable states with workspace summaries and associated validation evidence \\
        Memory view &
        Folded research history, recent result cards, and indexed evidence available through \textsc{Unfold} \\
        Peer evidence &
        Optional validated scores and compact method summaries from other workers \\
        Resource envelope $B_t$ &
        Remaining wall-clock budget and current execution constraints \\
        \bottomrule
    \end{tabular}
\end{table}

\paragraph{Decision prompt}
ESTRA evaluates candidate anchors using their validation evidence, recoverable workspace contents, recent progress, failure history, peer evidence, and the remaining resource budget. The validation metric is treated as one signal rather than the sole decision criterion, allowing ESTRA to preserve promising but not yet leading routes or redirect from saturated ones. Tool use is disabled during this deliberation, and the research worker returns one structured decision specifying the anchor \(a_n\), direction \(d_n\), supporting evidence, and intended search focus.

\begin{promptbox}{gray!65!black}{Schematic ESTRA decision contract}
\begin{Verbatim}[fontsize=\scriptsize,breaklines]
decision axes:
  anchor    = current_state | archived_state
  direction = extend | redirect

evidence:
  current_state
  archived_anchor_candidates
  memory_view
  validation_history
  optional_peer_evidence
  remaining_resource_budget

requirements:
  select exactly one anchor and one direction
  identify the current bottleneck
  justify the decision using recorded evidence
  specify the next search focus

structured output:
  anchor, target_state, direction,
  bottleneck, supporting_evidence,
  decision_reason, next_search_focus
\end{Verbatim}
\end{promptbox}

\paragraph{Folding and segment initialization.}
At a research-segment boundary, \textsc{Fold} summarizes the completed exploration as historical evidence for the next research segment. If ESTRA selects the current state, ScienceFlow retains the active workspace and folds the research segment that just ended. If it selects an archived state, ScienceFlow restores the corresponding workspace snapshot and folds the post-anchor branch that is no longer active. The original result cards and archived states remain preserved and addressable. ScienceFlow then applies \textsc{Assemble} to combine the stable prefix, selected-anchor context, ESTRA direction, and optional peer evidence into $P_{n+1}$.

\begin{promptbox}{gray!65!black}{Schematic ESTRA folding and segment initialization}
\begin{Verbatim}[fontsize=\scriptsize,breaklines]
[completed exploration]
terminal_state: Syy
selected_anchor: current_state | archived_state Sxx
direction: extend | redirect
folded_evidence:
  methods attempted
  validated outcomes
  observed failures
  avoid-repeat guidance

[next segment]
stable_prefix: P_stable
workspace: current workspace | Restore(W(Sxx))
anchor_context:
  memory view
  validation evidence
  resource records
  workspace view
direction_context: extend | redirect
optional_context: peer evidence
local_history: empty
\end{Verbatim}
\end{promptbox}

The anchor and direction axes yield four ESTRA outcomes: extending or redirecting from either the current state or an archived state. Exact workspace restoration is required only when an archived anchor is selected. Appendix~\ref{app:kttsp-estra-folding} presents a recorded archived-anchor transition and its folded post-anchor branch, while Appendix~\ref{app:kttsp-peer-trace} presents a KTTSP research-segment boundary at which peer evidence is included in the ESTRA context before a subsequent redirect.

\clearpage
\section{MLE-bench Supplementary Results}
\label{app:mlebench-supplement}

\subsection{Evaluation on the Full MLE-bench Set}
\label{app:full-mlebench}

We report per-task results on all 75 MLE-bench competitions, grouped by the official Lite, Medium, and High complexity splits. Each \textbf{Score} is the mean task-specific benchmark score over three independent runs. Because evaluation metrics differ across competitions, these raw scores should not be compared across tasks.

A checkmark in \textbf{Any-Medal} indicates that at least one of the three runs reached a bronze, silver, or gold threshold. For each run, \textbf{Cost(\$)} is the cumulative LLM/API expenditure up to the first medal-producing result; if no medal is obtained, it is the total expenditure over the complete run. The reported cost is averaged over the three runs and excludes accelerator infrastructure costs. \textbf{First Medal Time (h)} is averaged only over medal-producing runs and is reported as ``--'' when no run earns a medal. A value of 0 in \textbf{GPUs} denotes CPU-only execution. All runs use DeepSeek-V4-Flash-Preview as the research-worker backbone.

The per-task Any-Medal indicator reports the best observed outcome across runs and is therefore descriptive. It differs from Table~\ref{tab:main_results}, which computes the medal rate independently for each run and reports mean $\pm$ SEM across three runs. Collapsing the archive by competition yields 54 of 75 tasks (72.0\%) with at least one medal: 22 gold, 17 silver, and 15 bronze.

\paragraph{Four-hour archival cutoff.}
The Time dimension in Figure~\ref{fig:homepage-system-profile} uses archived
first-medal timestamps to characterize continued progress beyond a short
horizon. By the four-hour cutoff, 35 of 75 tasks (46.67\%) had a documented
medal-producing result; the final archive contains at least one medal for 54
tasks. The four-hour value is a descriptive best-observed cutoff reconstructed
from available task logs, rather than the mean of three independent four-hour
evaluations. It should therefore be interpreted as trajectory evidence that
additional tasks continue to reach medal quality after four hours, not as a
compute-normalized comparison or a causal estimate of any individual
ScienceFlow mechanism. The full-setting endpoint shown in the profile follows
the headline three-run 24-hour result of $70.22\pm1.18\%$.

Official-leaderboard entries below follow the public MLE-bench
leaderboard~\citep{openai2026mlebenchleaderboard}; available system papers are
cited alongside the corresponding agent names.

\begin{table}[p]
    \centering
    \scriptsize
    \renewcommand{\arraystretch}{1.04}
    \caption{Full-set MLE-bench Any-Medal results. Values are percentages
    reported as mean $\pm$ SEM over three runs. Baseline results retain their
    published reporting protocols. Systems are grouped by source, and the best
    result in each column is bolded.}
    \label{tab:mlebench-full-leaderboard}
    \setlength{\tabcolsep}{2.2pt}
    \begin{threeparttable}
        \begin{tabularx}{\textwidth}{
            @{}
            >{\raggedright\arraybackslash}p{0.205\textwidth}
            >{\raggedright\arraybackslash}p{0.215\textwidth}
            *{4}{>{\centering\arraybackslash}X}
            @{}
        }
            \toprule
            \textbf{Agent} & \textbf{LLM(s) used} &
            \makecell{\textbf{Lite}\\\textbf{(\%)}} &
            \makecell{\textbf{Medium}\\\textbf{(\%)}} &
            \makecell{\textbf{High}\\\textbf{(\%)}} &
            \makecell{\textbf{All}\\\textbf{(\%)}} \\
            \midrule
            \textbf{ScienceFlow (Ours)}\tnote{5} & DeepSeek-V4-Flash-Preview &
            \textbf{80.30 $\pm$ 1.52} & \textbf{74.56 $\pm$ 0.88} &
            44.44 $\pm$ 2.22 & \textbf{70.22 $\pm$ 1.18} \\
            \midrule
            \multicolumn{6}{l}{\textit{Public study}} \\
            \href{https://arxiv.org/abs/2608.02143}{Iris}~\citep{fu2026iris}\tnote{4} &
            Claude-Opus-4.6 & \textbf{80.30 $\pm$ 1.50} &
            64.00 $\pm$ 0.90 & 44.40 $\pm$ 2.20 & 64.90 $\pm$ 0.40 \\
            \href{https://github.com/InternScience/MLEvolve}{MLEvolve}~\citep{du2026mlevolve}\tnote{6} &
            Gemini-3.1-Pro-Preview & \textbf{80.30 $\pm$ 1.50} &
            64.00 $\pm$ 0.90 & \textbf{46.70 $\pm$ 0.00} & 65.30 $\pm$ 0.80 \\
            \addlinespace[2pt]
            \multicolumn{6}{l}{\textit{Official leaderboard}} \\
            \href{https://github.com/baidubce/FM-Agent}{Famou-Agent 2.0}~\citep{li2025fmagent} &
            Gemini-3-Pro-Preview & \textbf{80.30 $\pm$ 1.52} &
            64.04 $\pm$ 2.32 & 42.22 $\pm$ 2.22 & 64.44 $\pm$ 1.18 \\
            \href{https://github.com/aibuildai/AI-Build-AI}{AIBuildAI}~\citep{zhang2026aibuildai} &
            Claude-Opus-4.6 & 77.27 $\pm$ 0.00 & 61.40 $\pm$ 0.88 &
            46.67 $\pm$ 0.00 & 63.11 $\pm$ 0.44 \\
            \href{https://research.google/teams/cloud-ai-research/}{CAIR} MARS+~\citep{chen2026mars} &
            Gemini-3-Pro-Preview & 78.79 $\pm$ 1.52 & 60.53 $\pm$ 1.52 &
            44.44 $\pm$ 2.22 & 62.67 $\pm$ 0.77 \\
            \href{https://github.com/InternScience/MLEvolve}{MLEvolve}\tnote{4} &
            Gemini-3-Pro-Preview & \textbf{80.30 $\pm$ 1.52} &
            57.89 $\pm$ 1.52 & 42.22 $\pm$ 2.22 & 61.33 $\pm$ 1.33 \\
            \href{https://github.com/FractalAIResearchLabs/PiEvolve}{PiEvolve}~\citep{botla2025pievolve} &
            Gemini-3-Pro-Preview\tnote{3} & \textbf{80.30 $\pm$ 1.52}\tnote{2} &
            58.77 $\pm$ 0.88\tnote{2} & 40.00 $\pm$ 0.00\tnote{2} &
            61.33 $\pm$ 0.77\tnote{2} \\
            \href{https://github.com/baidubce/FM-Agent}{Famou-Agent 2.0} &
            Gemini-2.5-Pro & 75.76 $\pm$ 1.52 & 57.89 $\pm$ 1.52 &
            40.00 $\pm$ 0.00 & 59.56 $\pm$ 0.89 \\
            \href{https://github.com/sjtu-sai-agents/ML-Master}{ML-Master 2.0}~\citep{zhu2026mlmaster2} &
            DeepSeek-V3.2-Speciale & 75.76 $\pm$ 1.51 & 50.88 $\pm$ 3.51 &
            42.22 $\pm$ 2.22 & 56.44 $\pm$ 2.47 \\
            \href{https://research.google/teams/cloud-ai-research/}{CAIR} MARS &  Gemini-3-Pro-Preview & 74.24 $\pm$ 1.52 & 52.63 $\pm$ 3.04 & 37.78 $\pm$ 2.22 & 56.00 $\pm$ 1.54 \\
            \href{https://github.com/FractalAIResearchLabs/PiEvolve}{PiEvolve}\tnote{4} &
            Gemini-3-Pro-Preview\tnote{3} & 74.24 $\pm$ 3.03\tnote{2} & 45.61 $\pm$ 0.88\tnote{2} & 35.55 $\pm$ 2.22\tnote{2} &  52.00 $\pm$ 0.77\tnote{2} \\
            \href{https://github.com/Leeroo-AI/kapso}{Leeroo}~\citep{nadafian2026kapso} &
            Gemini-3-Pro-Preview\tnote{3} & 68.18 $\pm$ 2.62\tnote{2} & 44.74 $\pm$ 1.52\tnote{2} & 40.00 $\pm$ 0.00\tnote{2} & 50.67 $\pm$ 1.33\tnote{2} \\
            \href{https://thesislabs.ai}{Thesis} & gpt-5-codex &
            65.15 $\pm$ 1.52 & 45.61 $\pm$ 7.18 & 31.11 $\pm$ 2.22 &  48.44 $\pm$ 3.64 \\
            \href{https://research.google/teams/cloud-ai-research/}{CAIR} MLE-STAR-Pro-1.5~\citep{nam2025mle} &
            Gemini-2.5-Pro & 68.18 $\pm$ 2.62 & 34.21 $\pm$ 1.52 &
            33.33 $\pm$ 0.00 & 44.00 $\pm$ 1.33 \\
            \href{https://github.com/baidubce/FM-Agent}{Famou-Agent} &
            Gemini-2.5-Pro & 62.12 $\pm$ 1.52 & 36.84 $\pm$ 1.52 &
            33.33 $\pm$ 0.00 & 43.56 $\pm$ 0.89 \\
            \href{https://operand.com}{Operand} ensemble &
            gpt-5 (low verbosity/effort)\tnote{1} & 63.64 $\pm$ 0.00 &
            33.33 $\pm$ 0.88\tnote{2} & 20.00 $\pm$ 0.00\tnote{2} &
            39.56 $\pm$ 0.44\tnote{2} \\
            \href{https://research.google/teams/cloud-ai-research/}{CAIR} MLE-STAR-Pro-1.0\tnote{4} &
            Gemini-2.5-Pro & 66.67 $\pm$ 1.52 & 25.44 $\pm$ 0.88 &
            31.11 $\pm$ 2.22 & 38.67 $\pm$ 0.77 \\
            \href{https://github.com/Alpha-Innovator/InternAgent/}{InternAgent}~\citep{internagentteam2025internagent}\tnote{4} &
            DeepSeek-R1 & 62.12 $\pm$ 3.03 & 26.32 $\pm$ 2.63 &
            24.44 $\pm$ 2.22 & 36.44 $\pm$ 1.18 \\
            \href{https://github.com/microsoft/RD-Agent}{R\&D-Agent}~\citep{yang2025rdagent}\tnote{4} &
            gpt-5 & 68.18 $\pm$ 2.62 & 21.05 $\pm$ 1.52 &
            22.22 $\pm$ 2.22 & 35.11 $\pm$ 0.44 \\
            \href{https://heyneo.so/}{Neo}~\citep{neo2025} multi-agent\tnote{4} &
            undisclosed & 48.48 $\pm$ 1.52 & 29.82 $\pm$ 2.32 &
            24.44 $\pm$ 2.22 & 34.22 $\pm$ 0.89 \\
            \href{https://github.com/facebookresearch/aira-dojo/}{AIRA-dojo}~\citep{toledo2025ai} &
            o3 & 55.00 $\pm$ 1.47 & 21.97 $\pm$ 1.17 &
            21.67 $\pm$ 1.07 & 31.60 $\pm$ 0.82 \\
            \href{https://github.com/microsoft/RD-Agent}{R\&D-Agent} &
            o3 + GPT-4.1 & 51.52 $\pm$ 4.01 & 19.30 $\pm$ 3.16 &
            26.67 $\pm$ 0.00 & 30.22 $\pm$ 0.89 \\
            \href{https://github.com/zeroxleo/ML-Master}{ML-Master}~\citep{liu2025ml}\tnote{4} &
            DeepSeek-R1 & 48.48 $\pm$ 1.52 & 20.18 $\pm$ 2.32 &
            24.44 $\pm$ 2.22 & 29.33 $\pm$ 0.77 \\
            \href{https://github.com/microsoft/RD-Agent}{R\&D-Agent} &
            o1-preview & 48.18 $\pm$ 1.11 & 8.95 $\pm$ 1.05 &
            18.67 $\pm$ 1.33 & 22.40 $\pm$ 0.50 \\
            \href{https://github.com/wecoai/aideml}{AIDE}~\citep{jiang2025aide} & o1-preview &
            35.91 $\pm$ 1.86 & 8.45 $\pm$ 0.43 & 11.67 $\pm$ 1.27 &
            17.12 $\pm$ 0.61 \\
            \href{https://github.com/wecoai/aideml}{AIDE} & gpt-4o-2024-08-06 &
            18.55 $\pm$ 1.26 & 3.06 $\pm$ 0.33 & 8.15 $\pm$ 0.84 &
            8.63 $\pm$ 0.54 \\
            \href{https://github.com/wecoai/aideml}{AIDE} & claude-3-5-sonnet &
            19.70 $\pm$ 1.52 & 2.63 $\pm$ 1.52 & 2.22 $\pm$ 2.22 &
            7.56 $\pm$ 1.60 \\
            OpenHands~\citep{wang2024openhands} & gpt-4o-2024-08-06 & 12.12 $\pm$ 1.52 &
            1.75 $\pm$ 0.88 & 2.22 $\pm$ 2.22 & 4.89 $\pm$ 0.44 \\
            \href{https://github.com/wecoai/aideml}{AIDE} & llama-3.1-405b &
            10.23 $\pm$ 1.14 & 0.66 $\pm$ 0.66 & 0.00 $\pm$ 0.00 &
            3.33 $\pm$ 0.38 \\
            MLAB & gpt-4o-2024-08-06 & 4.55 $\pm$ 0.86 &
            0.00 $\pm$ 0.00 & 0.00 $\pm$ 0.00 & 1.60 $\pm$ 0.27 \\
            \bottomrule
        \end{tabularx}
            \begin{tablenotes}[flushleft]
            \scriptsize
            \item[1] Uses light assistance from Gemini-2.5-Pro, Grok-4, and
            Claude 4.1 Opus, distilled by Gemini-2.5-Pro.
            \item[2] For incomplete three-run evaluations, missing runs are
            counted as Any-Medal failures when computing the reported mean and
            SEM.
            \item[3] Uses Gemini-3-Pro-Preview primarily, with selected modules
            using GPT-5 and GPT-5-mini.
            \item[4] Uses a reported 12\,h or 36\,h per-task budget rather than
            the standard 24\,h budget; the exact setting follows the cited source.
            \item[5] ScienceFlow evaluates
            \texttt{tensorflow-speech-recognition-challenge} using the corrected
            setup from MLE-bench issue~\#63; baseline aggregates are retained as
            reported.
            \item[6] MLEvolve with Gemini-3.1-Pro-Preview uses a 12\,h per-task
            budget.
            \end{tablenotes}
    \end{threeparttable}
\end{table}

\begin{table}[htbp]
    \caption{Per-task results on the 22-task MLE-bench Lite split. Field
    definitions follow Section~\ref{app:full-mlebench}.}
    \label{tab:full-mlebench-low}
    \centering
    \setlength{\heavyrulewidth}{1.5pt}
    \setlength{\lightrulewidth}{0.5pt}
    \renewcommand{\arraystretch}{1.18}
    \resizebox{\linewidth}{!}{
        \begin{tabular}{@{}lcccccc@{}}
            \toprule  
            \textbf{Competition Name} &
            \textbf{Year} &  
            \textbf{Score} &
            \textbf{Cost(\$)} &
            \textbf{GPUs} &
            \textbf{Any-Medal} &
            \makecell{\textbf{First Medal}\\\textbf{Time (h)}}\\
            \midrule  
            detecting-insults-in-social-commentary & 2012 & 0.9588 & 0.0618 & 1 & $\checkmark$ & 2.54\\
            the-icml-2013-whale-challenge-right-whale-redux & 2013 & 0.9486 & 1.4995 & 1 & $\checkmark$ & 0.12\\
            mlsp-2013-birds & 2013 & 0.9099 & 0.1483 & 1 & $\checkmark$ & 1.95\\
            random-acts-of-pizza & 2015 & 0.7713 & 1.0641 & 1 & $\checkmark$ & 0.35\\
            denoising-dirty-documents & 2015 & 0.0116 & 0.0141 & 1 & $\checkmark$ & 1.08\\
            text-normalization-challenge-russian-language & 2017 & 0.9791 & 0.3177 & 1 & $\checkmark$ & 1.35\\
            dogs-vs-cats-redux-kernels-edition & 2017 & 0.0150 & 0.0150 & 1 & $\checkmark$ & 0.48\\
            spooky-author-identification & 2017 & 0.2800 & 0.2299 & 1 & $\checkmark$ & 3.66\\
            text-normalization-challenge-english-language & 2017 & 0.9965 & 0.1472 & 1 & $\checkmark$ & 1.59\\
            leaf-classification & 2017 & 0.0121 & 0.0603 & 0 & $\checkmark$ & 5.76\\
            jigsaw-toxic-comment-classification-challenge & 2018 & 0.9865 & 0.1521 & 1 & $\checkmark$ & 2.93\\
            new-york-city-taxi-fare-prediction & 2018 & 4.8527 & 0.2986 & 0 & - & -\\
            nomad2018-predict-transparent-conductors & 2018 & 0.0550 & 0.1805 & 0 & $\checkmark$ & 0.09\\
            dog-breed-identification & 2018 & 0.3525 & 0.5283 & 1 & - & -\\
            aerial-cactus-identification & 2019 & 1.0000 & 0.2885 & 1 & $\checkmark$ & 2.11\\
            aptos2019-blindness-detection & 2019 & 0.9204 & 0.3543 & 1 & $\checkmark$ & 4.84\\
            histopathologic-cancer-detection & 2019 & 0.9970 & 0.0214 & 1 & $\checkmark$ & 1.07\\
            siim-isic-melanoma-classification & 2020 & 0.9353 & 1.9233 & 2 & $\checkmark$ & 11.98\\
            plant-pathology-2020-fgvc7 & 2020 & 0.9928 & 0.0283 & 1 & $\checkmark$ & 0.46\\
            ranzcr-clip-catheter-line-classification & 2021 & 0.9124 & 0.2221 & 1 & - & -\\
            tabular-playground-series-dec-2021 & 2021 & 0.9622 & 0.0106 & 0 & $\checkmark$ & 0.42\\
            tabular-playground-series-may-2022 & 2022 & 0.9882 & 0.5283 & 0 & - & -\\
            \bottomrule  
        \end{tabular}
    }
\end{table}

\begin{table}[htbp]
    \caption{Per-task results on the 38-task MLE-bench Medium split. Field
    definitions follow Section~\ref{app:full-mlebench}.}
    \label{tab:full-mlebench-medium}
    \centering
    \setlength{\heavyrulewidth}{1.5pt}
    \setlength{\lightrulewidth}{0.5pt}
    \resizebox{\linewidth}{!}{
        \begin{tabular}{@{}lcccccc@{}}
            \toprule  
            \textbf{Competition Name} &
            \textbf{Year} &  
            \textbf{Score} &
            \textbf{Cost(\$)} &
            \textbf{GPUs} &
            \textbf{Any-Medal} &
            \makecell{\textbf{First Medal}\\\textbf{Time (h)}}\\
            \midrule  
            AI4Code & 2022 & 0.7911 & 1.6478 & 1 & - & -\\
            alaska2-image-steganalysis & 2020 & 0.8992 & 1.6478 & 2 & $\checkmark$ & 22.27\\
            billion-word-imputation & 2014 & 4.4668 & 0.1794 & 2 & $\checkmark$ & 7.87\\
            cassava-leaf-disease-classification & 2021 & 0.8995 & 0.3477 & 2 & $\checkmark$ & 11.77\\
            cdiscount-image-classification-challenge & 2017 & 0.7248 & 0.4307 & 2 & $\checkmark$ & 23.04\\
            chaii-hindi-and-tamil-question-answering & 2021 & 0.7458 & 0.0056 & 1 & $\checkmark$ & 6.11\\
            champs-scalar-coupling & 2019 & 0.6758 & 0.4361 & 1 & - & -\\
            facebook-recruiting-iii-keyword-extraction & 2013 & 0.5037 & 1.6478 & 0 & - & -\\
            freesound-audio-tagging-2019 & 2019 & 0.7346 & 0.2054 & 1 & $\checkmark$ & 2.04\\
            google-quest-challenge & 2020 & 0.3806 & 0.1189 & 1 & $\checkmark$ & 1.05\\
            h-and-m-personalized-fashion-recommendations & 2022 & 0.0248 & 0.0309 & 0 & $\checkmark$ & 10.20\\
            herbarium-2020-fgvc7 & 2020 & 0.4146 & 0.1594 & 1 & $\checkmark$ & 2.97\\
            herbarium-2021-fgvc8 & 2021 & 0.1656 & 0.4524 & 1 & $\checkmark$ & 4.06\\
            herbarium-2022-fgvc9 & 2022 & 0.6994 & 0.2894 & 1 & $\checkmark$ & 6.88\\
            hotel-id-2021-fgvc8 & 2021 & 0.1116 & 0.1905 & 1 & $\checkmark$ & 2.57\\
            hubmap-kidney-segmentation & 2021 & 0.9681 & 0.0891 & 1 & $\checkmark$ & 0.45\\
            icecube-neutrinos-in-deep-ice & 2023 & 1.3585 & 0.2270 & 1 & - & -\\
            imet-2020-fgvc7 & 2020 & 0.6544 & 0.6925 & 1 & $\checkmark$ & 22.33\\
            inaturalist-2019-fgvc6 & 2019 & 0.2972 & 0.2171 & 1 & $\checkmark$ & 3.49\\
            iwildcam-2020-fgvc7 & 2020 & 0.7329 & 0.1496 & 1 & $\checkmark$ & 2.63\\
            jigsaw-unintended-bias-in-toxicity-classification & 2019 & 0.8494 & 0.3405 & 1 & - & -\\
            kuzushiji-recognition & 2019 & 0.9527 & 0.0462 & 1 & $\checkmark$ & 3.01\\
            learning-agency-lab-automated-essay-scoring-2 & 2024 & 0.8355 & 0.6734 & 1 & $\checkmark$ & 1.27\\
            lmsys-chatbot-arena & 2024 & 1.0010 & 0.4159 & 1 & $\checkmark$ & 9.55\\
            multi-modal-gesture-recognition & 2013 & 0.2177 & 0.1462 & 1 & $\checkmark$ & 7.96\\
            osic-pulmonary-fibrosis-progression & 2020 & -6.7620 & 1.5486 & 2 & $\checkmark$ & 11.87\\
            petfinder-pawpularity-score & 2022 & 16.9877 & 0.9553 & 1 & $\checkmark$ & 12.03\\
            plant-pathology-2021-fgvc8 & 2021 & 0.9072 & 0.1213 & 1 & $\checkmark$ & 1.42\\
            seti-breakthrough-listen & 2021 & 0.8002 & 2.1684 & 1 & $\checkmark$ & 6.25\\
            statoil-iceberg-classifier-challenge & 2018 & 0.1349 & 3.0067 & 1 & $\checkmark$ & 23.42\\
            tensorflow-speech-recognition-challenge & 2018 & 0.9841 & 1.2587 & 1 & $\checkmark$ & 1.52\\
            tensorflow2-question-answering & 2020 & 0.5691 & 0.1996 & 1 & - & -\\
            tgs-salt-identification-challenge & 2018 & 0.7704 & 1.1963 & 1 & - & -\\
            tweet-sentiment-extraction & 2020 & 0.7178 & 0.8099 & 1 & $\checkmark$ & 11.80\\
            us-patent-phrase-to-phrase-matching & 2022 & 0.8707 & 0.4126 & 1 & $\checkmark$ & 4.01\\
            uw-madison-gi-tract-image-segmentation & 2022 & 0.6902 & 2.4634 & 1 & - & -\\
            ventilator-pressure-prediction & 2021 & 0.3384 & 0.5535 & 1 & - & -\\
            whale-categorization-playground & 2018 & 0.4783 & 0.8077 & 1 & $\checkmark$ & 3.70\\
            \bottomrule  
        \end{tabular}
    }
\end{table}

\begin{table}[htbp]
    \caption{Per-task results on the 15-task MLE-bench High split. Field
    definitions follow Section~\ref{app:full-mlebench}.}
    \label{tab:full-mlebench-high}
    \centering
    \setlength{\heavyrulewidth}{1.5pt}
    \setlength{\lightrulewidth}{0.5pt}
    \resizebox{\linewidth}{!}{
        \begin{tabular}{@{}lcccccc@{}}
            \toprule  
            \textbf{Competition Name} &
            \textbf{Year} &  
            \textbf{Score} &
            \textbf{Cost(\$)} &
            \textbf{GPUs} &
            \textbf{Any-Medal} &
            \makecell{\textbf{First Medal}\\\textbf{Time (h)}}\\
            \midrule  
            3d-object-detection-for-autonomous-vehicles & 2019 & 0.0597 & 0.1874 & 1 & $\checkmark$ & 1.73\\
            bms-molecular-translation & 2021 & 4.8469 & 1.0800 & 1 & - & -\\
            google-research-identify-contrails-reduce-global-warming & 2023 & 0.4474 & 4.1043 & 1 & - & -\\
            hms-harmful-brain-activity-classification & 2024 & 0.8100 & 4.1043 & 1 & - & -\\
            iwildcam-2019-fgvc6 & 2019 & 0.2107 & 0.1442 & 1 & $\checkmark$ & 0.68\\
            nfl-player-contact-detection & 2023 & 0.6585 & 4.1043 & 1 & $\checkmark$ & 1.63\\
            predict-volcanic-eruptions-ingv-oe & 2021 & 3,068,390.3330 & 2.8259 & 1 & $\checkmark$ & 4.57\\
            rsna-2022-cervical-spine-fracture-detection & 2022 & 0.5613 & 4.1043 & 1 & - & -\\
            rsna-breast-cancer-detection & 2023 & 0.1225 & 4.1043 & 1 & - & -\\
            rsna-miccai-brain-tumor-radiogenomic-classification & 2021 & 0.6163 & 4.4492 & 1 & $\checkmark$ & 0.59\\
            siim-covid19-detection & 2021 & 0.4527 & 4.1043 & 1 & - & -\\
            smartphone-decimeter-2022 & 2022 & 4.6753 & 4.1043 & 0 & - & -\\
            stanford-covid-vaccine & 2020 & 0.2272 & 0.0580 & 1 & $\checkmark$ & 0.07\\
            vesuvius-challenge-ink-detection & 2023 & 0.1415 & 4.1043 & 1 & - & -\\
            vinbigdata-chest-xray-abnormalities-detection & 2021 & 0.3344 & 0.4418 & 1 & $\checkmark$ & 1.43\\
            \bottomrule  
        \end{tabular}
    }
\end{table}

\subsection{Operational Telemetry Coverage}
\label{app:operational-telemetry-coverage}

All 75 tasks are included in the performance evaluation under the same ScienceFlow protocol. The operational analysis is retrospective and requires both ESTRA-event records and checkpoint/snapshot telemetry. Both streams are available for 54 tasks; the remaining 21 tasks retain benchmark outcomes but lack complete mechanism-level records. The smaller denominator therefore reflects telemetry availability rather than task selection or a different system configuration. These records support the mechanism analysis in Table~\ref{tab:operational-telemetry}.

\clearpage

\section{Mathematical Optimization Details}
\label{app:math-optimization}

\subsection{Formal Task Definitions}
\label{app:math-task-definitions}

\paragraph{Circle packing.}
For $n=26$, a candidate consists of circle centers
$p_i=(x_i,y_i)\in[0,1]^2$ and radii $r_i\geq 0$. The task is
\begin{equation}
    \max_{\{p_i,r_i\}_{i=1}^{26}} \sum_{i=1}^{26} r_i
    \quad\text{subject to}\quad
    r_i \leq x_i,y_i \leq 1-r_i,
    \qquad
    \lVert p_i-p_j\rVert_2 \geq r_i+r_j
    \quad(i\neq j).
    \label{eq:circle-packing-task}
\end{equation}
The first constraint keeps every circle inside the unit square, and the second
prevents pairwise overlap. The reported score is the sum of radii of a feasible
configuration, so larger values are better~\citep{novikov2025alphaevolve}.

\paragraph{Ratio minimization.}
A candidate is a set of $n=16$ distinct points
$P=\{p_1,\ldots,p_{16}\}\subset\mathbb{R}^2$. Define
\begin{equation}
    d_{\min}(P)=\min_{i<j}\lVert p_i-p_j\rVert_2,
    \qquad
    d_{\max}(P)=\max_{i<j}\lVert p_i-p_j\rVert_2.
\end{equation}
The objective is
\begin{equation}
    \min_{P:\,d_{\min}(P)>0}
    \rho(P),
    \qquad
    \rho(P)=\frac{d_{\max}(P)}{d_{\min}(P)}.
    \label{eq:ratio-minimization-task}
\end{equation}
The objective is invariant to translation, rotation, and uniform scaling.
AlphaEvolve defines the objective as $\rho(P)$ but follows the source packing
tables in reporting $\rho(P)^2$~\citep{novikov2025alphaevolve}. To match
Table~\ref{tab:math-optimization}, we take the positive square root of such
published values and report $\rho(P)$ for every method; lower values are better.

\paragraph{Uncertainty inequality.}
For an integrable real-valued function $f:\mathbb{R}\rightarrow\mathbb{R}$,
let
\begin{equation}
    \widehat f(\xi)=\int_{\mathbb{R}}f(x)e^{-2\pi i x\xi}\,\mathrm{d}x,
    \qquad
    A(f)=\inf\{r>0:f(x)\geq 0\text{ for all }|x|\geq r\}.
\end{equation}
For nonzero even functions satisfying
$\max\{f(0),\widehat f(0)\}<0$ and having finite $A(f)$ and
$A(\widehat f)$, the Fourier sign-uncertainty constant is
\begin{equation}
    C_4=\inf_f A(f)A(\widehat f).
    \label{eq:uncertainty-c4}
\end{equation}
Our benchmark follows the first, Hermite-polynomial formulation used by
\citet{novikov2025alphaevolve}, which refines the construction of
\citet{goncalves2017hermite}. It searches over
\begin{equation}
    Q_{\boldsymbol c}(z)=\sum_{k=0}^{m}c_k H_{4k}(z),
    \qquad
    f_{\boldsymbol c}(x)=
    Q_{\boldsymbol c}(\sqrt{2\pi}\,x)e^{-\pi x^2},
    \label{eq:uncertainty-hermite-family}
\end{equation}
where $H_j$ is the physicists' Hermite polynomial. The degrees $4k$ make
$f_{\boldsymbol c}$ invariant under the Fourier transform. As in the official
AlphaEvolve verifier, the final coefficient is chosen so that
$Q_{\boldsymbol c}(0)=0$, and the polynomial is oriented to be positive at
infinity. After removing the resulting factor $z^2$, let $r_{\max}$ be the
largest positive real root across which
$Q_{\boldsymbol c}(z)/z^2$ changes sign. The evaluator returns
\begin{equation}
    U(\boldsymbol c)=\frac{r_{\max}^2}{2\pi},
    \qquad C_4\leq U(\boldsymbol c).
    \label{eq:uncertainty-upper-bound}
\end{equation}
Thus the reported score is the verified Hermite-construction upper bound on
$C_4$, and lower values are better. This benchmark does not include the
separate Laguerre-polynomial refinement also discussed by AlphaEvolve.

\subsection{Evaluation and Feasibility Audit}
\label{app:math-audit}

ScienceFlow candidates are re-evaluated from the JSON artifacts stored in
immutable workspace snapshots rather than from rounded stage-ledger metrics.
The audit parses the saved decimal literals, verifies the task-specific
structure and feasibility conditions, and recomputes the displayed objective.
Unreadable artifacts, malformed candidates, and candidates that fail the
corresponding feasibility test are excluded from comparison.

\paragraph{Circle packing.}
The strict audit treats every saved coordinate and radius as an exact decimal
rational and applies the inequalities in Equation~\ref{eq:circle-packing-task}
with zero numerical tolerance. It records the minimum boundary slack and the
minimum squared pairwise slack. For a candidate with a small negative slack, we
also compute the smallest common radius reduction $\delta\geq0$ that makes its
fixed centers feasible; the repaired score is
$\sum_i r_i-26\delta$. This repair is used only to produce a conservative
feasible score and never to improve a candidate. The artifact underlying the
reported value $2.6359830849$ requires
$\delta=9.26\times10^{-15}$ and yields the strictly feasible score
$2.63598308491745569$, which is unchanged at the precision shown in
Table~\ref{tab:math-optimization}.

\paragraph{Ratio minimization.}
The audit parses coordinates with 60-digit decimal precision, requires exactly
16 two-dimensional points, recomputes all 120 pairwise Euclidean distances, and
rejects a candidate when $d_{\min}=0$. It then evaluates
$d_{\max}/d_{\min}$ directly, avoiding the six-decimal inverse-squared metric
stored in the stage ledger. The best audited DeepSeek-V4-Flash-Preview candidate gives
$3.590157365310609$, while the best openPangu-2.0-Pro candidate gives
$3.590157365325370478$. Both values are reported as the unsquared ratio
$d_{\max}/d_{\min}$.

\paragraph{Uncertainty inequality.}
We reconstruct the public AlphaEvolve B.4 Hermite verifier exactly. The audit
enforces $Q_{\boldsymbol c}(0)=0$ and a positive limit at infinity, divides out
the factor $z^2$, computes the real roots symbolically, and selects the largest
positive sign-changing root using 200-digit root approximations. The resulting
bound is evaluated by Equation~\ref{eq:uncertainty-upper-bound}. The reported
ScienceFlow bounds $0.348200107555$ and $0.343293122432$ both pass this verifier.

\paragraph{Baseline provenance.}
Published entries in Table~\ref{tab:math-optimization} are transcribed from
their cited papers or released repositories and retain the construction and
precision reported by those sources; they are not presented as independent
reruns. The AlphaEvolve uncertainty entry is independently reproduced with its
public Hermite verifier. OpenEvolve is run locally as described in
Section~\ref{subsec:exp-math}; its retained scores come from the local task
evaluator. In particular, its circle-packing output does not preserve the final
evaluated coordinates for a separate exact-decimal audit. The local paper
archive contains the derived audit summaries used here; some corresponding raw
run artifacts remain in the original experiment storage and are not duplicated
in the paper repository.

\section{SpOC4-KTTSP Supplementary Details}
\label{app:kttsp-traces}

\subsection{Competition, Tracks, and Evaluation Protocol}
\label{app:kttsp-protocol}

SpOC4 is the fourth Space Optimisation Competition organized by ESA's Advanced
Concepts Team. Its Keplerian Tomato Traveling Salesperson Problem asks a
spacecraft to collect all targets in lunar orbit in minimum mission time while
respecting orbital-transfer and maneuverability constraints
\citep{esa_spoc4_2026}. We evaluate the three official instances. The easy,
medium, and hard tracks contain 50, 182, and 1,052 targets, respectively; their
minimum transfer times are 0.001, 0.01, and $1/1440$ days, and their maximum
mission durations are 200, 500, and 3,000 days.

For an $N$-target instance, a submission contains a permutation of all targets,
$N-1$ departure epochs, and $N-1$ transfer durations. The evaluator checks the
permutation, time bounds, and chronological consistency before solving each
transfer with a Lambert solver in both directions and with at most 20
revolutions. A nominal leg may use at most $100\,\mathrm{m\,s^{-1}}$ of
$\Delta V$; at most five exception legs may exceed this threshold, and no leg
may exceed $600\,\mathrm{m\,s^{-1}}$. The score is the arrival time at the
final target in days, so lower is better. Candidates are serialized as an
official-format JSON decision vector, and invalid candidates receive no score.

Trajectory evaluation is CPU-only. The easy and medium results were obtained through staged continuation runs, whereas KTTSP-hard used the separate ten-day campaign described in Section~\ref{subsec:exp-kttsp}, with DeepSeek-V4-Pro-Preview, DeepSeek-V4-Flash-Preview, and GLM-5.1. Because resumed stages and worker counts differed across tracks, we report the public leaderboard outcomes without treating them as a compute-normalized comparison.

\subsection{Context-Folding Trace on KTTSP-hard}
\label{app:kttsp-memory-folding}

This section instantiates the memory operations in Section~\ref{sec:memory}
with a recorded W01 checkpoint from the KTTSP-hard run. At \texttt{S115}, the
append-only stage ledger $m_v$ contained 115 result cards and occupied 73,964
characters. When the memory view exceeded $B_{\mathrm{mem}}$, \textsc{Fold}
compressed cards \texttt{S01--S112} into the addressable summary
\texttt{L\_S01\_S112}. The resulting view $\widetilde m_v$ occupied 4,661
characters, a 93.7\% reduction relative to the raw ledger. At the same
research-segment boundary, the runtime emitted a 6,773-character memory-bearing state packet for
the next research segment. This telemetry covers the folded view and associated state
metadata, rather than the complete context $P_{n+1}$; the stable prefix,
resource provenance, workspace view, and direction instruction are composed
during final context assembly.

\paragraph{Before context folding: persistent ledger.}
The memory $m_v$ keeps each stage as an independently addressable result card in
\texttt{.run\_results.md}. The excerpts below reproduce selected fields from
three decision-critical cards; line wrapping is added only for page layout.
The \texttt{validity} field denotes ScienceFlow's internal evidence tier, not
the KTTSP track difficulty or the official feasibility decision. The initial
feasible result, an unverified intermediate attempt, and the best validated
result are all retained rather than keeping improvements alone.

\begin{promptbox}{gray!65!black}{Selected fields from stage cards before context folding}
\begin{Verbatim}[fontsize=\scriptsize,breaklines]
S01  metric=1928.39350319; validity=medium
BRIEF: Bridge strategy achieves 0 infeasible transfers with
       4 exceptions and 1928.39d duration.
WHY:  First fully feasible candidate.

S18  metric=1761.93330934; validity=low
NOTE:  invalid_submission: no submission artifact found.
BRIEF: Finer local-window timing refinement reached 1761.93d.
WHY:  Timing resolution helped, but the result was not eligible
      as a verified comparable entry.

S112 metric=393.229; validity=high
BRIEF: Forward-pass timing refinement improved 696 legs and
       saved 7.70d locally, but yielded only 0.001d net gain.
WHY:  Five frozen exception legs absorbed the upstream savings;
      further progress requires re-timing or re-routing them.
FILES: code=refine_timing_fast.py
\end{Verbatim}
\end{promptbox}

\paragraph{After context folding: budgeted view.}
In $\widetilde m_v$, the historical portion is replaced by a compact summary
with an addressable identifier, while cards needed for verification and
immediate continuation remain raw. The generated summary records the folded range, its
best stage, representative early evidence, the number of omitted intermediate
cards, and the last folded stage:

\begin{promptbox}{gray!65!black}{Folded stage-memory view}
\begin{Verbatim}[fontsize=\scriptsize,breaklines]
Historical summary: L_S01_S112  [S01--S112]
best_stage=S112
- S01:  metric=1928.39350319; validity=medium; first feasible
- S02:  metric=1928.39350319; validity=medium; safe baseline
- S03:  metric=1928.39350319; validity=high; validated baseline
- S04:  metric=1891.24195011; validity=medium; timing re-opt
... 107 intermediate stages summarized ...
- S112: metric=393.229; validity=high;
        frozen exception legs block the timing cascade

key_stage_index: S112=best_valid
available_expand_id: L_S01_S112
verification_raw_cards: S01, S18, S112
current_raw_segment: S113, S114, S115
\end{Verbatim}
\end{promptbox}

\paragraph{Active review through unfolding.}
The runtime field \texttt{available\_expand\_id} exposes the identifier used by
the formal \textsc{Unfold} operation. The following view illustrates an active
comparison of the unverified timing attempt and the best validated route. This
is an observed \textsc{Unfold} event from the archived W01 runtime trace, shown
using the corresponding indexed cards:

\begin{promptbox}{gray!65!black}{On-demand unfolded memory view}
\begin{Verbatim}[fontsize=\scriptsize,breaklines]
request: Unfold(L_S01_S112, stages={S18, S112})
index:   L_S01_S112 -> .run_results.md [S01--S112]

returned_raw_cards:
- S18:  metric=1761.93330934; validity=low
  NOTE:  invalid_submission: no submission artifact found
  BRIEF: Finer local-window timing refinement reached 1761.93d.
  WHY:  Timing resolution helped, but the result was not eligible
        as a verified comparable entry.
- S112: metric=393.229; validity=high
  BRIEF: Forward-pass timing refinement improved 696 legs and
         saved 7.70d locally, but yielded only 0.001d net gain.
  WHY:  Five frozen exception legs absorbed the upstream savings;
        further progress requires re-timing or re-routing them.
  FILES: code=refine_timing_fast.py

context_effect: temporary augmentation of folded view
ledger_effect:  none (m_v remains append-only)
\end{Verbatim}
\end{promptbox}

The index resolves the request to the original cards in
\texttt{.run\_results.md}. In the implementation, this runtime-executed
\textsc{Unfold} is realized through indexed workspace reads rather than a
separate memory service.

\FloatBarrier
\subsection{ESTRA-Triggered Folding Trace on KTTSP-hard}
\label{app:kttsp-estra-folding}

This section instantiates the ESTRA-triggered fold in
Sections~\ref{subsubsec:reanchoring} and~\ref{sec:memory} with a recorded W01
transition. At 17:17 UTC on June~21, at a research-segment boundary caused by the context
limit, the worker selected the runtime action \texttt{switch\_stage}, moving
from terminal stage \texttt{S27} to the archived anchor at \texttt{S23}. The
transition folded the post-anchor branch \texttt{S24--S27} into an
845-character summary and assembled a 3,583-character state packet for the
restored trajectory. Unlike capacity-based folding of the persistent memory
view, this fold scope was determined by the selected archived anchor: it closes
the abandoned branch and carries its diagnosis into the next research segment.

\paragraph{Before ESTRA-triggered folding: abandoned tail.}
The four cards refer to the same 1746.8935-day artifact, and no new route is
produced. Their internal evidence tiers vary because the repeated evaluations
provide different amounts of verification evidence; this variation does not
indicate different outcomes under the official KTTSP feasibility checks:

\begin{promptbox}{gray!65!black}{Selected fields from abandoned-tail cards}
\begin{Verbatim}[fontsize=\scriptsize,breaklines]
S24 metric=1746.89350787; validity=medium
BRIEF: Existing artifact re-scored end-to-end with pykep;
       1046 feasible, 4 exceptions, 0 infeasible.

S25 metric=1746.89350787; validity=high
BRIEF: Direct pykep recomputation validates the artifact before
       route-ordering experiments.

S26 metric=1746.89350787; validity=medium
BRIEF: Baseline survives the transition but remains far behind peer W00.
WHY:  Timing-only refinement is exhausted on the sorted route.

S27 metric=1746.89350787; validity=medium
WHY:  Route ordering is the dominant bottleneck; the next route
      should use proxy-prefiltered construction and bounded checks.
\end{Verbatim}
\end{promptbox}

\paragraph{After ESTRA-triggered folding: restored-state evidence.}
The generated summary preserves the branch-level conclusion rather than
carrying all four cards verbatim into the next segment's agent-facing state
packet:

\begin{promptbox}{gray!65!black}{ESTRA-folded abandoned-tail view}
\begin{Verbatim}[fontsize=\scriptsize,breaklines]
runtime_action: switch_stage
restore: terminal S27 -> target S23
fold_scope: abandoned tail S24--S27

tail_summary:
- Repeated pykep rescoring only revalidated the existing 1746.89d
  artifact; no new candidate was produced.
- Validity fluctuated between medium and high on the same artifact;
  treat the repeated rescoring as non-informative.
- Timing-only refinement on the sorted route is saturated.
- The 1746.89d baseline remains far behind W00 at 521.91d;
  route-ordering or structural changes are required.
- Avoid redundant validation passes and further timing polish
  without a new ordering or construction strategy.

summary_chars: 845
restored_state_packet_chars: 3583
preserved: raw cards, snapshot, logs, terminal archive
\end{Verbatim}
\end{promptbox}

After restoration, the worker resumes from the executable workspace snapshot
at \texttt{S23}, while the folded \texttt{S24--S27} branch is retained as
completed negative evidence. The original cards remain addressable in the
archived memory through \textsc{Unfold}. Thus, the ESTRA-triggered fold changes
the active research route and agent-facing context without altering the
persistent research record.

\FloatBarrier
\subsection{Peer-Guided Search Trace on KTTSP-hard}
\label{app:kttsp-peer-trace}

The W01 trajectory in Figure~\ref{fig:kttsp-context-performance} contains a
concrete cross-worker coordination episode. W00 established an early validated
solution of 521.9054 mission days at \texttt{S60}, while W01 remained at
1746.8935 days. Each worker maintained an isolated executable workspace, and no
artifact or workspace state was transferred between them. Instead, compact peer
evidence comprising W00's validated score and method summary entered W01's
ESTRA context. The worker used this evidence to redirect its search from
timing-only refinement toward route-ordering strategies.

The archived runtime records preserve both the peer-evidence payload and the
resulting direction instruction. We identify each archived ESTRA decision by
its worker and stage as \texttt{ESTRA-Wxx-Syy}.

At W01's \texttt{S25} research-segment boundary, compact peer evidence comprising W00's
validated score and a brief windowed-2-opt summary entered the worker's ESTRA
decision (event \texttt{ESTRA-W01-S25}). The worker used the performance gap
to diagnose route ordering as the bottleneck, but redirected toward its own
proxy-prefiltered nearest-neighbor construction rather than copying W00's
procedure. PyKEP supplies the Lambert-transfer evaluation.

\begin{promptbox}{gray!65!black}{Selected fields from ESTRA-W01-S25}
\begin{Verbatim}[fontsize=\scriptsize,breaklines]
Route order is the bottleneck not timing; need faster NN using
orbital-element proxy pre-filter then pykep verify only top-3,
not full grid on 30 candidates.

Bottleneck: Route ordering is the dominant factor: 1746d vs peer
521d, but greedy NN with pykep evaluation is too slow to complete
within resource limits.

Exploration summary: Timing refinement on original sorted route
converged at 1746.89d; greedy NN route rewrite attempted but too
slow at 47s/10 legs. Peer at 521.91d via windowed 2-opt.

Redirect: Proxy-pre-filtered NN: rank by orbital distance,
pykep-verify only top 3-5 with coarse grid.
\end{Verbatim}
\end{promptbox}

After several route-ordering attempts, W01 reached 1321.8824 days but still
trailed W00 by approximately 802 days. At \texttt{S79}, the peer summary
provided more explicit method-level evidence, producing a second redirect at
the operator level (event \texttt{ESTRA-W01-S79}).

\begin{promptbox}{gray!65!black}{Selected fields from ESTRA-W01-S79}
\begin{Verbatim}[fontsize=\scriptsize,breaklines]
Worker decision: Adapt the peer-validated 2-opt+SLSQP operators;
inclination-bin sorting appears saturated.

Bottleneck: Static inclination-bin sorting cannot match W00's
route-reordering and timing-optimization strategy (519 days).

Evidence: An approximately 802-day gap remains; finer bins improve
the objective by only 54 days, while greedy variants remain infeasible.

Next action: Apply 2-opt route swaps and SLSQP timing refinement
to W01's 1321.8824-day incumbent.
\end{Verbatim}
\end{promptbox}

\begin{table}[H]
    \centering
    \small
    \caption{Validated checkpoints and ESTRA decisions in the KTTSP-hard
    peer-guided overtaking case. Mission duration is measured in days; lower
    is better.}
    \label{tab:kttsp-peer-overtaking}
    \begin{tabular}{@{}lllrl@{}}
        \toprule
        \textbf{Time (UTC)} & \textbf{Worker} & \textbf{Stage} &
        \textbf{Duration} & \textbf{Search event} \\
        \midrule
        Jun 21 16:12 & W00 & S60  & 521.9054 & Early peer best \\
        Jun 21 16:20 & W01 & S25  & 1746.8935 & Structural ESTRA redirect \\
        Jun 25 18:04 & W01 & S78  & 1321.8824 & Fine inclination bins \\
        Jun 25 19:12 & W01 & S79  & 1321.8824 & Operator-level ESTRA redirect \\
        Jun 25 23:01 & W01 & S80  & 573.0305 & Phase-greedy narrow bins \\
        Jun 26 00:00 & W01 & S85  & 525.9948 & Near peer best \\
        Jun 26 07:29 & W01 & S97  & \textbf{420.2742} & First validated lead over W00 \\
        Jun 27 00:06 & W01 & S110 & 393.2352 & Ultrafine timing grid \\
        Jun 28 23:29 & W01 & S112 & 393.229 & Final refinement \\
        \bottomrule
    \end{tabular}
\end{table}

Between the first peer-aware ESTRA decision at \texttt{S25} and its first
validated lead over W00 at \texttt{S97}, W01 reduced its incumbent from
1746.8935 to 420.2742 days, a 75.9\% improvement over 4 days and
15 hours. More directly, W01 produced the 420.2742-day result within
12 hours and 17 minutes of the operator-level redirect at \texttt{S79}.
It ultimately reached 393.229 days, outperforming W00's final validated
519.6255-day result by 24.32\%.

The trace supports peer-guided adaptation rather than direct solution reuse.
Applying windowed SLSQP to W01's existing route at \texttt{S94} improved
the objective by only 0.02 days. W01 established the subsequent lead after
combining a 10-degree inclination-bin route with phase-proximal construction
and finer timing grids. Peer evidence therefore served as search guidance:
it revealed the structural limitations of the current route family and
identified promising operators, while W01 independently instantiated and
validated a distinct route and schedule.

\FloatBarrier

\clearpage

\section{SciModelingBench Provenance and Evaluation Metadata}
\label{app:scimodeling-metadata}

\begingroup
\small

This appendix records the data sources and hidden evaluation boundary for the
12 SciModelingBench tasks in Section~\ref{sec:exp-sci}. We use the public
\href{https://github.com/xukp20/sci-modeling-bench/releases/tag/v0.10.0}{SciModelingBench
0.10.0 release} and its
\href{https://huggingface.co/datasets/sci-modeling-bench/design-bench}{released
dataset}. In Table~\ref{tab:scimodeling-metadata}, $N$ is the number of
candidates in each submission, $K$ is the summary size used by the shared
diagnostics, and $Q$ is the total query limit for one system--task pair.
ScienceFlow's two workers split this limit. The external baselines are OpenCode
1.18.4~\citep{opencode2026}, Pi 0.81.1~\citep{zechner2026pi}, Codex
0.144.5~\citep{openai2026codex}, and Claude Code
2.1.72~\citep{anthropic2026claudecode}. All four use
DeepSeek-V4-Flash-Preview through non-interactive clients.

\begin{table}[H]
    \centering
    \scriptsize
    \setlength{\tabcolsep}{1.7pt}
    \renewcommand{\arraystretch}{1.13}
    \caption{SciModelingBench task metadata. Visible scale describes the data
    given to the agent; evaluation scale describes the exact reference domain
    or label-hidden pool. Values in parentheses are canonical candidates
    represented by row-level observations. DrugMatrix contains six endpoint
    tasks from one study collection. BKM, NE, and NDCG denote \textit{best-$K$
    mean}, \textit{normalized enrichment}, and \textit{global NDCG}. Data terms
    come from the original source, while the Hopper row separately identifies
    simulator software licenses.}
    \label{tab:scimodeling-metadata}
    \begin{tabularx}{\textwidth}{
        @{}
        >{\raggedright\arraybackslash}p{0.115\textwidth}
        >{\raggedright\arraybackslash}p{0.14\textwidth}
        >{\raggedright\arraybackslash}p{0.12\textwidth}
        >{\centering\arraybackslash}p{0.075\textwidth}
        >{\raggedright\arraybackslash}p{0.225\textwidth}
        >{\raggedright\arraybackslash}X
        @{}
    }
        \toprule
        \textbf{Task group} &
        \textbf{Visible} &
        \textbf{Evaluation} &
        \textbf{$N/K/Q$} &
        \textbf{Trusted evaluator} &
        \textbf{Source / terms} \\
        \midrule
        \makecell[l]{TFBind8\\\textit{BBO}} &
        32,768 sequences &
        65,536 sequences &
        32/5/20 &
        Normalized PBM E-score lookup; BKM &
        PBM/UniPROBE~\citep{barrera2016variation,hume2015uniprobe}; source-specific academic-use terms \\
        \addlinespace[2pt]
        \makecell[l]{TFBind10 Pho4\\\textit{BBO}} &
        2,087,323 rows (524,300 seq.) &
        $4^{10}$ sequences &
        128/16/20 &
        Four-replicate affinity posterior; NE &
        BET-seq~\citep{le2018binding,fordycelab2018betseq}; CC BY 4.0 \\
        \addlinespace[2pt]
        \makecell[l]{Superconductor\\\textit{ranking}} &
        16,795 records (12,179 groups) &
        2,985 groups &
        32/5/20 &
        Median critical temperature by composition; NDCG &
        UCI Superconductivity Data~\citep{hamidieh2018superconductordata,hamidieh2018superconductor}; CC BY 4.0 \\
        \addlinespace[2pt]
        \makecell[l]{UTR MRL\\\textit{ranking}} &
        76,877 50-mers &
        5,043 50-mers &
        128/16/10 &
        Two-replicate mean ribosome load; NE &
        GEO GSE114002/mRNABench~\citep{sample2018utrdata,sample2019utr}; terms unknown \\
        \addlinespace[2pt]
        \makecell[l]{GFP\\\textit{ranking}} &
        41,372 proteins &
        10,343 proteins &
        128/16/20 &
        Protein-level median log brightness; NE &
        Sarkisyan Figshare~\citep{bolotin2016gfpdata,sarkisyan2016gfp}; CC BY 4.0 \\
        \addlinespace[2pt]
        \makecell[l]{Hopper\\Controller\\\textit{ranking}} &
        1,920 policies &
        1,280 policies &
        32/5/10 &
        Mean of 500 frozen Hopper-v5 rollouts; NDCG &
        Design-Bench policy assets (terms not separately stated); Gymnasium (MIT), MuJoCo (Apache-2.0)~\citep{trabucco2022designbench,todorov2012mujoco} \\
        \addlinespace[2pt]
        \makecell[l]{DrugMatrix (6)\\\textit{ranking}} &
        9,442 animal rows &
        390 conditions &
        16/5/8 &
        Absolute log treatment--control deviation; NDCG &
        NIEHS CEBS DrugMatrix~\citep{ntp2023drugmatrix}; terms unspecified \\
        \bottomrule
    \end{tabularx}
\end{table}

\paragraph{Reconstruction principle}
SciModelingBench takes its scientific settings from
Design-Bench~\citep{trabucco2022designbench}, but does not automatically treat
the packaged arrays as ground truth. For every retained setting, we return to
the original experiment or simulator assets. We then define the candidate
identity, visible observations, hidden reference data, and trusted objective
separately. Agents may train predictive models to rank candidates, but those
predictions are not used as ground truth outside the training distribution. We
omit settings that do not support reproducible candidates and evaluation,
rather than keep them only to match a historical learned oracle.

\paragraph{Transcription-factor binding}
For TFBind8, the original PBM table stores an 8-mer and its reverse complement
in separate columns that share one E-score. The historical preprocessing joins
these columns, producing 65,792 rows and duplicating the 256
reverse-complement palindromes. We remove only these exact duplicates, keep
non-palindromic reverse complements as distinct valid sequences, and verify
that the result contains all $4^8$ 8-mers exactly once. We also retain both the
published and normalized E-score scales. The resulting complete measured
landscape supports direct lookup without a learned oracle.

For TFBind10 Pho4, the legacy table contains replicate-level binding estimates.
The same 10-mer can therefore have conflicting values, and using row order as a
lookup rule silently selects one replicate. We rebuild the task from four
BET-seq bound/input count replicates. The lower-half observations expose the raw
replicate measurements, while a deterministic affinity posterior covers the
complete $4^{10}$ domain and is oriented so that higher values are better. This
keeps replicate disagreement visible instead of hiding it behind the final row.

\paragraph{Superconducting materials}
Design-Bench uses a fitted random forest to score arbitrary composition
vectors. Composition alone, however, omits crystal structure, phase, pressure,
defects, and processing. The source data also contain repeated compositions
with different measured critical temperatures. We first realign the two UCI
tables and normalize the amounts of the 86 elements. We then group
proportional formulas under one canonical composition while retaining every
measurement. The group median is the ranking target, and the measurement range
and dispersion remain visible to the agent. Evaluation is limited to the
measured high-temperature pool, so composition-only extrapolations are not
treated as physical ground truth.

\paragraph{UTR and GFP sequence measurements}
The original UTR setting uses a learned ResNet to score arbitrary sequences.
We instead use the unique 50-nucleotide variable regions measured in the eGFP
reporter assay. We select the unmodified-RNA condition, use the mean ribosome
load over two replicates, and hold out one group defined by
$(\text{uAUG presence},\text{Kozak quality})$ as a finite measured pool.

For GFP, translating the legacy nucleotide rows produces 56,086 rows but only
51,715 unique proteins. Among the 1,202 duplicated protein groups, 1,201 have
conflicting labels, and the wild type alone appears 534 times. We instead use
the authors' protein-level aggregate, verify the unique 237-residue sequences,
and split the data by protein identity before adding nucleotide and barcode
observations. This prevents synonymous encodings of the same protein from
appearing on both sides of the split. Both tasks therefore evaluate measured
candidates rather than predictions from historical neural-network oracles.

\paragraph{Hopper controller relabeling}
We keep Design-Bench's 3,200 policy vectors as candidates but replace the
historical labels, each of which came from a single return. Every policy is
evaluated over 500 stochastic episodes in Hopper-v5. Policies share the same
reset-seed schedule but use separate action-noise seeds. The released data
include raw returns, episode lengths, termination flags, and uncertainty
summaries; mean return is the frozen target. From the offline observations, the
agent ranks a held-out pool of higher-performing measured policies. It cannot
access the simulator, train PPO, or request online rollouts during the task.

\paragraph{DrugMatrix reconstruction}
The legacy ChEMBL arrays represent each condition only by a molecular token,
dropping its dose, duration, route, vehicle, study, sex, and animal-group
context. Nearly every molecule then maps to several endpoint labels, and a
row-level split can place the same molecular identity on both sides. We
therefore rebuild the task from the NIEHS CEBS individual-animal
clinical-pathology release instead of reusing these arrays. We preserve the
treatment context and controls, map chemical identity by strict name and CASRN
matching, and use ChEMBL only to link structures rather than supply endpoint
labels.

Candidates are five-day treatments at the highest observed dose, with complete
endpoints and controls matched by study, duration, route, vehicle, and sex. We
hide the treatment-animal rows for these candidates but keep their matched
controls and observations from other doses or times visible. All six tasks use
the same candidate conditions and rank them by the absolute log difference
between the endpoint mean and the matched-control mean. This score measures the
size of a biological change, not drug safety or quality.

\paragraph{Protocol and metric audit}
Before running any agents, we fixed the lower-score thresholds, structured
holdouts, submission sizes, and primary metrics using source-data analyses and
task-specific
\href{https://github.com/xukp20/sci-modeling-bench/tree/main/docs/suites/design-bench}{random,
simple, and cross-validated audits}. These audits helped us avoid saturated
metrics, distinguish strong submissions, and keep the queried fraction small
relative to the hidden domain. The choices were not based on ScienceFlow or
external-agent results. Here, an ``exact'' evaluator returns the same score
whenever it is applied to the same frozen data. The underlying experimental
measurements may still contain noise.

\paragraph{Runtime leakage controls and contamination scope}
The agent workspace contains only the exported observations, candidate view,
manifest, task contract, and submission directory. Hidden labels, trusted
objectives, evaluator caches, full dataset copies, and submission records stay
outside the workspace, and external network access is disabled. The evaluator
returns one score for the full batch rather than a score for each candidate.
The harness separately records query use and the best valid artifact. These
controls block runtime answer lookup through the benchmark infrastructure.

They do not show that public papers or upstream data were absent from model
pretraining. We therefore claim runtime leakage controls and audited
provenance, not a contamination-free evaluation. General scientific knowledge
is allowed, although in some cases it is difficult to distinguish inference
from memorized candidate labels.

\paragraph{Licensing and availability}
SciModelingBench and Design-Bench software use the MIT license, but this does
not relicense the underlying experimental data. Table~\ref{tab:scimodeling-metadata}
therefore lists the original source-specific, CC BY 4.0, or unknown data terms.
Because these terms differ, the dataset card lists the license as ``other''
rather than assigning one license to the full collection. Processing records
and task-specific details are available in the public
\href{https://github.com/xukp20/sci-modeling-bench/tree/main/docs/suites/design-bench}{suite
documentation} and
\href{https://huggingface.co/datasets/sci-modeling-bench/design-bench}{dataset
card}.
\endgroup

\end{document}